\documentclass[a4paper,fleqn]{cas-dc}

\usepackage[numbers]{natbib}
\usepackage{xcolor}
\usepackage{makecell}
\usepackage{placeins}
\usepackage{lipsum}
\usepackage{duckuments}
\usepackage{multirow}
\usepackage{tabularx}
\usepackage{graphicx}
\usepackage{subcaption}
\usepackage{siunitx}
\usepackage{comment}
\usepackage{amsmath}
\usepackage{amssymb}
\usepackage{graphicx}
\usepackage{tikz}
\usetikzlibrary{
    positioning,
    calc,
    arrows.meta,
    shapes,
    fit
}

\def\tsc#1{\csdef{#1}{\textsc{\lowercase{#1}}\xspace}}
\tsc{WGM}
\tsc{QE}
\tsc{EP}
\tsc{PMS}
\tsc{BEC}
\tsc{DE}

\begin{document}
\let\WriteBookmarks\relax
\def\floatpagepagefraction{1}
\def\textpagefraction{.001}

\shorttitle{A New Dataset and Framework for Robust Road Surface Classification via Camera–IMU Fusion}

\shortauthors{de Lima Costa~et~al.}

\title [mode = title]{A New Dataset and Framework for Robust Road Surface Classification via Camera–IMU Fusion}                      
\tnotemark[1]

\tnotetext[1]{© 2026. This manuscript version is made available under the CC-BY-NC-ND 4.0 license https://creativecommons.org/licenses/by-nc-nd/4.0/}

%

\author[1]{Willams de Lima Costa}
\cormark[1]
\ead{wlc2@cin.ufpe.br}
\credit{Conceptualization, Data curation, Formal analysis, Investigation, Methodology, Project administration, Supervision, Writing - original draft.}

\author[1]{Thifany Ketuli Silva de Souza}
\credit{Formal analysis, Investigation, Methodology, Validation, Writing - original draft}
\author[1]{Jonas Ferreira Silva}
\credit{Methodology, Software, Project administration}
\author[1]{Carlos Gabriel Bezerra Pereira}
\credit{Software, Visualization}
\author[1]{Bruno Reis Vila Nova}
\credit{Software, Methodology}
\author[1]{Leonardo Silvino Brito}
\credit{Software, Methodology}
\author[2]{Rafael Raider Leoni}
\credit{Data curation, Methodology, Supervision}
\author[2]{Juliano Silva}
\credit{Data curation, Methodology, Supervision}
\author[2]{Valter Ferreira}
\credit{Funding acquisition, Project administration, Supervision, Validation}
\author[3]{Sibele Miguel Soares Neto}
\credit{Data curation, Methodology, Supervision}
\author[4]{Samantha Uehara}
\credit{Funding acquisition, Project administration, Supervision, Validation}
\author[5]{Daniel Giacometti Amaral}
\credit{Validation}
\author[1]{João Marcelo Teixeira}
\credit{Funding acquisition, Project administration, Supervision, Validation, Writing - review \& editing}
\author[1]{Veronica Teichrieb}
\credit{Funding acquisition, Project administration, Supervision, Validation, Writing - review \& editing}
\author[1]{Cristiano Coelho de Araújo}
\credit{Funding acquisition, Project administration, Supervision, Validation, Writing - review \& editing}

\affiliation[1]{organization={Voxar Labs, Centro de Informática, Universidade Federal de Pernambuco},
    addressline={Av. Jornalista Aníbal Fernandes, s/n}, 
    city={Recife},
    postcode={50.740-560}, 
    country={Brazil}}

\affiliation[2]{organization={Volkswagen Truck and Bus},
    city={Resende},
    postcode={27537-803}, 
    state={Rio de Janeiro},
    country={Brazil}}

\affiliation[3]{organization={Stellantis Brasil},
    city={Porto Real},
    postcode={27570-000}, 
    state={Rio de Janeiro},
    country={Brazil}}

\affiliation[4]{organization={Volkswagen do Brasil},
    city={São Bernardo do Campo},
    postcode={09823-901}, 
    state={São Paulo},
    country={Brazil}}

\affiliation[5]{organization={Embeddo},
    city={Volta Redonda},
    postcode={27251-330}, 
    state={Rio de Janeiro},
    country={Brazil}}

\cortext[cor1]{Corresponding author}

\begin{abstract}
Road surface classification (RSC) is a key enabler for environment-aware predictive maintenance systems. However, existing RSC techniques often fail to generalize beyond narrow operational conditions due to limited sensing modalities and datasets that lack environmental diversity. This work addresses these limitations by introducing a multimodal framework that fuses images and inertial measurements using a lightweight bidirectional cross-attention module followed by an adaptive gating layer that adjusts modality contributions under domain shifts. 	Given the limitations of current benchmarks, especially regarding lack of variability, we introduce ROAD, a new dataset composed of three complementary subsets: (i) real-world multimodal recordings with RGB–IMU streams synchronized using a gold-standard industry datalogger, captured across diverse lighting, weather, and surface conditions; (ii) a large vision-only subset designed to assess robustness under adverse illumination and heterogeneous capture setups; and (iii) a synthetic subset generated to study out-of-distribution generalization in scenarios difficult to obtain in practice. Experiments show that our method achieves a +1.4 pp improvement over the previous state-of-the-art on the PVS benchmark and an +11.6 pp improvement on our multimodal ROAD subset, with consistently higher F1-scores on minority classes. The framework also demonstrates stable performance across challenging visual conditions, including nighttime, heavy rain, and mixed-surface transitions. These findings indicate that combining affordable camera and IMU sensors with multimodal attention mechanisms provides a scalable, robust foundation for road surface understanding, particularly relevant for regions where environmental variability and cost constraints limit the adoption of high-end sensing suites.
\end{abstract}


\maketitle

\section{Introduction}
\label{sec:introduction}
Unplanned fleet downtime caused by unexpected faults and generic maintenance schedules remains a significant source of operational inefficiency and cost in the transportation sector \citep{garner2021modeling, silva2023analysis, mittal2024iot, etukudoh2024theoretical}. Most maintenance plans are still based on Usage-Based Maintenance strategies, using simple indicators such as mileage or working hours, and therefore overlooking the variability of vehicle operating conditions. When a vehicle is subjected to harsher use than anticipated, delayed component replacement can increase safety risks and potentially lead to breakdowns \citep{hajizadeh2015anomaly, zhang2025multi}. On the other hand, when usage is milder than expected, premature servicing results in unnecessary maintenance costs and wasted resources \citep{etukudoh2024theoretical}.

Recent advances in Artificial Intelligence (AI) have opened new opportunities to make maintenance planning more adaptive to real operating conditions. Road Surface Classification (RSC) is an emerging AI task that uses sensors to classify the type of road a vehicle is currently driving on. This information can be used to improve maintenance plans, enabling manufacturers and service centers to obtain a data-driven understanding of the vehicle’s operating severity. In recent years, many techniques have been proposed to automatically map road surfaces using diverse sensors, such as Inertial Measurement Units (IMUs), i.e., accelerometer and gyroscope \citep{aslam2025transformer, hnoohom2023comprehensive, menegazzo2021road, van2025hybrid}, sound \citep{pothapragada2022acoustic}, and visual inputs \citep{sandeep2024road}. There are, however, two main limitations in the current state-of-the-art.

First, most existing techniques have been developed and validated on datasets captured exclusively under daylight and otherwise controlled conditions \citep{menegazzo2021road,pothapragada2022acoustic,van2025hybrid}. Although daylight itself is not inherently “controlled”, the lack of variation in illumination, weather, and capture context results in highly homogeneous scenarios that do not reflect the challenges of real-world operation. This limited diversity restricts scalability to higher Technology Readiness Levels and reduces applicability to deployment environments where nighttime, adverse weather, and mixed-surface conditions are common. Second, single-sensor approaches \citep{arce2024advancing, van2025hybrid} lack cross-modal redundancy, making them more vulnerable to performance degradation in out-of-distribution scenarios.

To address these limitations, we propose a multimodal framework that fuses camera and inertial data for RSC, designed and evaluated to operate reliably under diverse real-world conditions. Our proposal combines visual cues, which can be helpful in out-of-distribution settings (such as in diverse asphalt surfaces and conditions), with the vibration cues captured by inertial sensors, reducing ambiguity in fine-grained scenarios. Given the current limitations addressed above, we also introduce a new benchmark dataset that extends beyond conventional datasets by including challenging scenarios, such as heavy rain, nighttime, severe dust, and combined night-rain conditions. Our proposed \textbf{R}oad surface \textbf{O}bservation and \textbf{A}nalysis \textbf{D}ataset (ROAD) provides the first systematic basis for evaluating multimodal RSC on challenging and heterogeneous real-world scenarios. By fusing vision and inertial, it improves upon the state-of-the-art by 1.4\% on the Passive Vehicular Datasets benchmark \citep{menegazzo2021road} and by 11.6\% on our new proposed benchmark, ROAD.

In summary, the contributions of this work are threefold:
\begin{enumerate}
    \item We introduce the \textbf{R}oad surface \textbf{O}bservation and \textbf{A}nalysis \textbf{D}ataset (ROAD), a new benchmark designed to evaluate road surface classification under realistic and heterogeneous operating conditions. ROAD is the first dataset to jointly provide long, continuous driving sequences with synchronized camera–IMU data and substantial environmental diversity, including nighttime, heavy rain, dust, and mixed adverse scenarios. This dataset exposes significant performance degradation in existing methods and enables systematic analysis of robustness, temporal consistency, sensor degradation, and out-of-distribution generalization, capabilities that are not supported by prior RSC benchmarks.

    \item We propose a multimodal RSC framework explicitly designed to address four major sources of real-world variability (visual appearance, geometric and sensor placement, motion-induced effects, and environmental/domain shifts) through bidirectional cross-attention and adaptive gating. When evaluated on ROAD, this framework yields substantial performance gains over prior approaches, particularly on minority surface classes and under adverse conditions, while also maintaining competitive performance on the established PVS benchmark. Each of these variability sources is explicitly grounded in the methodological components described in \autoref{sec:methodology}.

    \item We provide a comprehensive experimental analysis across multimodal, vision-only, and sensor-degraded settings, combining quantitative results and qualitative diagnostics. This evaluation shows that IMU cues primarily act as a robustness enhancer rather than a primary accuracy driver, improving temporal stability during ambiguous surface transitions and revealing multimodal failure modes related to sensor misalignment and domain shifts.
\end{enumerate}


The remainder of this paper is organized as follows. Section 2 reviews related work on road surface classification and multimodal sensor fusion. Section 3 describes our proposed multimodal framework. Section 4 details our proposed dataset. Section 5 reviews the experimental setup and evaluation metrics. Section 6 discusses the experimental results and analyzes their implications for robust road surface perception, with high-level considerations for intelligent vehicle systems. Finally, Section 7 concludes the paper, highlights the major findings and limitations of our work, and outlines directions for future research.

\begin{figure*}
	\centering
		\includegraphics[width=\linewidth]{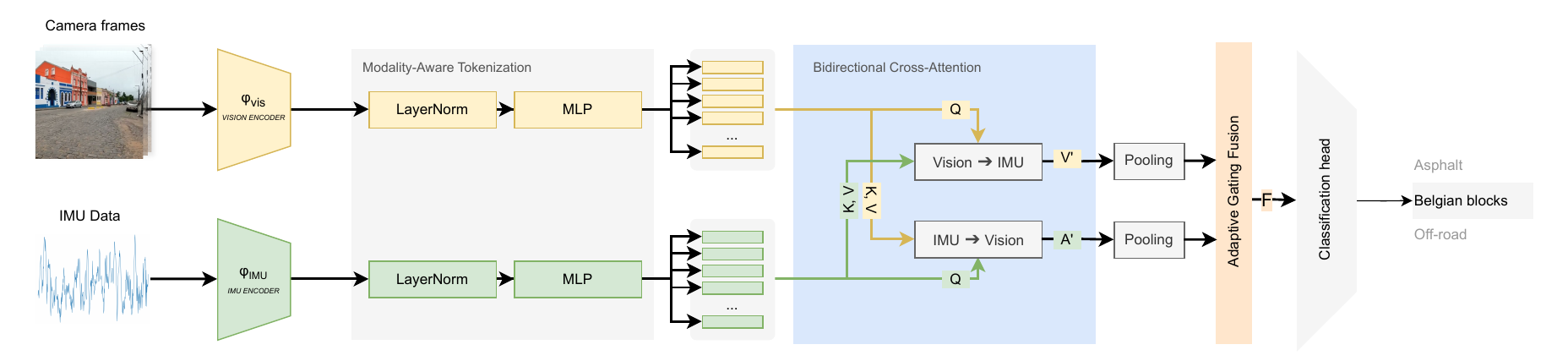}
	\caption{Overview of the proposed multimodal framework for road surface classification. An EfficientNet-B0 vision encoder processes raw RGB frames, while a CNN-BLSTM encoder processes inertial measurements (accelerometer and gyroscope). Each branch applies its own modality-specific LayerNorm+MLP tokenization module to generate visual and inertial tokens. These tokens are then aligned via a \textit{bidirectional cross-attention} mechanism, in which each modality queries the other to exchange contextual information and produce refined representations $V'$ and $A'$. The resulting embeddings are aggregated and combined through an \textit{adaptive gating fusion} module, which computes sample-dependent modality weights to produce a single fused representation $F$. Finally, $F$ is passed to a classification head to predict the road surface type.
}

	\label{fig:architecture}
\end{figure*}

\section{Related Works}
Modern RSC methods span multiple sensing modalities, problem formulations, and operational goals. We organize this section into four complementary parts. First, we review approaches focused on detecting localized road anomalies, such as potholes and speed bumps (\autoref{subsec:road-anomaly}). Second, we discuss methods aimed at classifying the global road surface type over longer temporal trajectories (\autoref{subsec:surface}). Third, we present and analyze publicly available datasets for road surface classification, highlighting their sensing modalities, environmental coverage, and limitations (\autoref{subsec:road-surface-datasets}). Finally, we provide a dedicated summary that synthesizes the main findings of the literature and explicitly identifies open gaps and challenges that motivate the contributions of this work (\autoref{subsec:summary-identified-gaps}).

\subsection{Road anomaly and surface defect detection}
\label{subsec:road-anomaly}
A substantial portion of the literature on road surface assessment focuses on local anomalies, which are short, spatially confined irregularities, such as potholes, speed bumps, patches, and uneven segments. These works aim to detect abrupt changes in the driving surface rather than to classify its global material type. One of the earliest efforts in this direction is reported in \citet{menegazzo2022speed}, which proposed a comprehensive inertial sensor pipeline for speed bump detection, comparing multiple Deep Learning architectures on a dataset collected using a smartphone for IMU data. Their experiments demonstrate that temporal Deep Learning models outperform purely convolutional ones, underscoring the importance of sequential dependencies in vibration-based anomaly detection.

Building on the vision-based line of research, \citet{qureshi2023deep} introduce a Deep Learning framework for pavement condition rating that primarily relies on image-based analysis of surface distress patterns. Their results highlight the strong discriminative capacity of CNNs in controlled environments, but their method, being purely vision-based, depends on pavement appearance, which may pose challenges in uncontrolled lighting or varying camera viewpoints.

\citet{ruggeri2024comparative} conducted a detailed comparative analysis of Deep Learning architectures for pothole identification, evaluating multiple vision-based representations. Their work highlights the feasibility of performing road surface analysis with inexpensive, non-specialized sensors, demonstrating that both architectures can detect anomalies reliably under real driving conditions. Importantly, they also evaluate these models across heterogeneous computational platforms, from cloud-based virtual machines to low-power edge devices such as the Raspberry Pi, showing how local processing can reduce data transmission overheads while maintaining adequate performance.

Similarly, \citet{raslan2024evaluation} investigate a broader taxonomy of anomalies that includes normal road surface, potholes, bad road surfaces, and speed bumps. The authors construct a multivariate time-series dataset from vehicular sensors and evaluate several data representation techniques to determine which formats are most effective for classifying these surface conditions. Their methodology involves transforming raw sensor streams into different feature representations before feeding them into deep learning models, enabling a systematic comparison of how various encodings influence predictive performance. The study provides an extensive benchmarking analysis across the defined anomaly classes, highlighting how representation choices impact model behavior within their proposed dataset.

Finally, \citet{arce2024advancing} advance anomaly detection through a richer categorization that includes even vs. uneven surfaces, patches, manholes, potholes, and speed bumps. Their Deep Learning methodology integrates visual cues with contextual priors to operate in dense urban environments. Although their taxonomy remains limited to localized defects, the work broadens the scope compared to pothole-only or speed-bump–only approaches.

Together, these studies demonstrate substantial progress in detecting localized defects using inertial signals, visual features, or hybrid pipelines. However, the scope of these works remains inherently limited: they focus on event-like abnormalities rather than on identifying the global surface material or modeling extended temporal context across continuous driving segments. This distinction is crucial for downstream applications such as predictive maintenance modeling, vehicle/terrain interaction modeling, and off-road navigation. These limitations motivate research into surface-type classification, discussed next.

\subsection{Surface type classification}
\label{subsec:surface}
In contrast to anomaly-focused approaches, a complementary line of research targets the global characterization of the road surface, aiming to distinguish between broader categories, such as asphalt, dirt, cobblestone, or other traversable materials. These methods generally rely on visual, acoustic, or inertial cues to infer the underlying surface type over longer temporal windows, supporting applications that require terrain awareness rather than localized defect detection. Below, we summarize representative works spanning this direction.

\citet{menegazzo2021road} present one of the earliest and most influential studies on road surface classification, proposing a framework based on inertial and GPS measurements to distinguish between asphalt, dirt, and cobblestone. Using a multivariate sensor dataset and evaluating several Machine Learning and Deep Learning models, the authors demonstrate the feasibility of surface type prediction without visual cues. Although their work remains a cornerstone contribution in this area and continues to serve as a baseline for subsequent research, its evaluation is inherently constrained by the characteristics of the dataset used. Despite being collected in real driving conditions, the recordings were captured under low-variability conditions, lacking challenging scenarios such as nighttime driving or substantial illumination and weather variability.

\citet{pothapragada2022acoustic} extend this research by combining acoustic and visual sensors captured by a rover platform. The authors evaluate their system using a simple K-Nearest Neighbors approach and explore how audio and image-based information can distinguish between traversable surfaces. This study exemplifies early efforts to integrate heterogeneous sensor streams for surface recognition, taking an important step toward multimodal understanding of road surfaces.

Following this pipeline, \citet{van2025hybrid} perform a deep analysis of how different Machine Learning models would perform for this task. Using a handcrafted set of features from vehicle-based sensor data, they propose a multi-stage feature selection strategy to identify the most relevant predictors before model training, aiming to improve computational efficiency. Their experimental results report how each ensemble method behaves under the selected feature subsets, providing a comparative analysis of accuracy and computational cost.

\subsection{Road surface datasets}
\label{subsec:road-surface-datasets}
RSC research has been supported by a diverse set of datasets that vary widely in sensing modalities, geographic context, and environmental coverage. Because the performance and generalization of both unimodal and multimodal approaches depend strongly on the underlying data, we summarize here the main publicly available datasets that target this task and discuss their relevance and limitations for multimodal fusion and robustness evaluation.

The Passive Vehicular Sensors (PVS) \cite{menegazzo2021road} dataset provides multimodal data comprising consumer-grade inertial sensors, GPS, and a monocular camera mounted on three passenger vehicles. PVS includes asphalt, cobblestone, and dirt surfaces and serves as a valuable benchmark for inertial-based surface classification, being one of the most widely used datasets for RSC to this day. Nonetheless, its recordings are constrained to daylight and low-variability weather, captured using non-synchronized consumer-grade sensors, and do not emphasize multimodal fusion or long-range temporal transitions under adverse conditions.

The Road Traversing Knowledge (RTK) \citep{rateke2021road} dataset contains approximately 62k labeled frames captured from a low-cost camera in two cities in southern Brazil, with annotations for road material, surface quality, and limited semantic segmentation. Although regionally relevant, RTK remains vision-only, covers a small geographic area, and lacks long sequences and challenging environmental diversity, reducing its suitability for studying multimodal fusion or robustness under degraded visibility.

RoadSaw \citep{cordes2022roadsaw} is one of the few datasets combining camera and IMU measurements for road-condition estimation, including dry, wet, and winter scenarios. However, its recordings are limited to short driving segments mostly in European environments, with a single inertial unit and without extended continuous trajectories or diverse adverse conditions such as heavy rain, dust, or mixed night–rain settings. RoadSaw does not target cross-modal synchronization studies or multi-IMU vibration analysis.

Lastly, the Road Surface Classification Dataset (RSCD) \citep{10101715, ZHAO2024111019} is the most recent dataset for this field. It provides nearly one million image patches labeled for surface material, friction level, and unevenness, collected from a monocular camera in urban environments. While RSCD offers large-scale visual data, it is restricted to a vision-only modality, focuses primarily on patch-level appearance rather than continuous sequences, and does not include adverse conditions such as nighttime, heavy rain, or dust.

Taken together, existing datasets each address specific aspects of road surface perception: large-scale vision, regional diversity, limited multimodality, or friction estimation, but none jointly provides: (i) synchronized camera–IMU streams with multiple high-frequency inertial units, (ii) long continuous recordings capturing natural transitions between surface types, and (iii) substantial environmental diversity encompassing nighttime, heavy rain, dust, and mixed adverse scenarios. These gaps motivate the development of the ROAD dataset proposed in this work and introduced in \autoref{sec:dataset}, designed to support research on multimodal fusion (i), robustness (ii), and out-of-distribution generalization under realistic operational conditions (i, ii, iii).

\subsection{Summary and identified gaps}
\label{subsec:summary-identified-gaps}

Taken together, the studies reviewed in \autoref{subsec:road-anomaly} and \autoref{subsec:surface} reveal two complementary directions in road surface understanding: approaches focused on detecting short-duration anomalies, such as potholes or speed bumps, and methods aimed at classifying the global surface type over extended trajectories. While both lines of research have advanced rapidly, their evaluation has been limited by the characteristics of the datasets on which they rely.

As discussed in \autoref{subsec:road-surface-datasets}, existing datasets differ substantially in modality, sequence length, and environmental variability. Vision-only collections such as RTK \citep{rateke2021road} and RSCD \citep{10101715, ZHAO2024111019} provide large-scale appearance data but lack inertial modalities, continuous trajectories, and adverse conditions. Multimodal datasets such as PVS \citep{menegazzo2021road} and RoadSaw \citep{cordes2022roadsaw} offer richer sensor streams, yet their recordings remain constrained to daylight or low-variability scenarios, feature short driving segments, or rely on non-synchronized consumer-grade hardware. Across all cases, the absence of long, continuous sequences with natural transitions between surface types limits the ability to analyze temporal consistency or to evaluate adaptive multimodal fusion strategies.

These constraints collectively make it challenging to assess model robustness under realistic driving variability, particularly when illumination degrades, weather introduces visual noise, or surface materials change gradually over time. Consequently, there is a pressing need for more comprehensive benchmarks that jointly provide: (i) synchronized multimodal data with high-frequency inertial measurements, (ii) long-duration recordings capturing naturally occurring transitions, and (iii) substantial environmental diversity, including nighttime, heavy rain, dust, and mixed adverse scenarios. These elements are essential for studying generalization, cross-modal redundancy, sensor degradation, and out-of-distribution behavior, key components for deploying RSC systems in real-world operational settings. In this work, we address these challenges by combining a multimodal benchmark with a fusion architecture based on cross-attention and adaptive gating, explicitly designed to analyze how visual and inertial cues interact under realistic and degraded sensing conditions.

\section{Multimodal Road Surface Classification Framework}
\label{sec:methodology}

We will detail our multimodal framework, also shown in \autoref{fig:architecture}, designed to explicitly mitigate the four sources of variability introduced in \autoref{sec:introduction}. To address visual appearance variability, we propose a vision stream (\autoref{subsec:vision-processing-stream}) that extracts features that, through training, remain informative across changes in illumination, weather, and surface material. An inertial stream (\autoref{subsec:inertial-processing-stream}) targets motion-induced variability by learning vibration patterns associated with surface irregularities and vehicle dynamics. Besides collaborating to visual appearance and motion-induced variability, we employ a multimodal preprocessing pipeline (\autoref{sec:preprocessing}) also to mitigate geometric and sensor placement variability, accounting for differences in camera mounting, IMU sensor positioning, sensor quality, and small variations in vehicle configuration (e.g., suspension tuning or cabin height). Finally, a fusion module (\autoref{subsec:fusion-module}) will combine these representations to ensure robustness under domain shifts, including adverse environments and out-of-distribution conditions. This structure ensures that a corresponding methodological component concretely addresses every challenge identified in the Introduction.


Specifically, given an image $I \in \mathbb{R}^{H \times W \times C}$ with spatial resolution $H \times W$ and $C$ color channels, and a set of inertial sensor readings 
$S = \{ s_t \}_{t=1}^{T}$, where $s_t \in \mathbb{R}^{d}$ represents $d$-dimensional acceleration and angular velocity measurements collected over a temporal window of length $T$, 
our method processes each data stream independently to obtain two feature representations: $f(I)$ for the visual branch and $f(S)$ for the inertial branch. 

These representations are subsequently integrated by a multimodal fusion module
\begin{equation}
z = \mathcal{F}(f(I), f(S)),
\end{equation}
which learns joint cross-modal dependencies.
The fused vector $z$ is then fed to a fully connected classifier that outputs the posterior probabilities over the $K$ road-type classes:
\begin{equation}
Y = \delta(Wz + b),
\end{equation}
where $W$ and $b$ are trainable parameters, and $\delta(\cdot)$ denotes the softmax activation. We will detail each processing stream below.

\subsection{Vision processing stream}
\label{subsec:vision-processing-stream}
The vision stream is based on the EfficientNet-B0~\citep{tan2019efficientnet} backbone, pre-trained on ImageNet~\citep{deng2009imagenet} and fine-tuned for the road surface classification task. This stream extracts features from exterior camera images, providing contextual information that complements the inertial modality. We replace the original classification head of EfficientNet-B0 with a new parameterized linear layer corresponding to the number of classes in the dataset (see Section~\ref{sec:dataset}). 
Formally,
\begin{equation}
f(I) = \phi_{\text{vis}}(I) \in \mathbb{R}^{D_{\text{vis}}},
\end{equation}
where $\phi_{\text{vis}}(\cdot)$ denotes the visual encoder (initialized from EfficientNet-B0), and $D_{\text{vis}} = 1280$ is the dimensionality of the visual embedding produced by the final projection layer. This stream, therefore, directly operationalizes the visual appearance variability described in \autoref{sec:introduction}, providing robustness to illumination changes, heterogeneous asphalt appearances, weather artifacts, and surface-texture diversity.

\subsection{Inertial processing stream}
\label{subsec:inertial-processing-stream}
The inertial stream processes accelerometer and gyroscope data using a hybrid CNN-BLSTM architecture that learns vibration patterns and long-term temporal dependencies in IMU signals. The 1-D convolutional layer detects characteristic excitation patterns arising from road surface irregularities, producing a sequence of latent feature maps that encode local vibrations. These maps are then processed by a bidirectional long short-term memory (BLSTM) network~\citep{hochreiter1997long}, which models temporal correlations and captures how vibration responses evolve over time. 
Formally,
\begin{equation}
f(S) = \phi_{\text{imu}}(S) \in \mathbb{R}^{D_{\text{imu}}},
\end{equation}
where $\phi_{\text{imu}}(\cdot)$ denotes the inertial encoder composed of the CNN-BLSTM hybrid architecture, and $D_{\text{imu}} = 256$ is the dimensionality of the resulting inertial embedding vector. This stream provides complementary motion dynamics that reinforce the visual cues in the subsequent multimodal fusion stage and directly address the motion-induced variability discussed earlier, including vibration signatures from surface irregularities, mechanical resonance, and speed-dependent vehicle dynamics.

\subsection{Multimodal preprocessing}
\label{sec:preprocessing}

To mitigate the geometric and sensor-placement variability described before, we apply extensive preprocessing and data-augmentation routines to both modalities. These steps explicitly account for differences in camera mounting angles, IMU positioning across the chassis, minor variations in vehicle configuration, and sensor-quality inconsistencies such as calibration drift or optical degradation. By modeling these factors during training, the framework learns feature representations that remain stable across heterogeneous vehicle setups and sensing conditions.


\begin{figure}[h!]
\centering

\begin{subfigure}{0.48\columnwidth}
    \centering
    \includegraphics[width=\columnwidth]{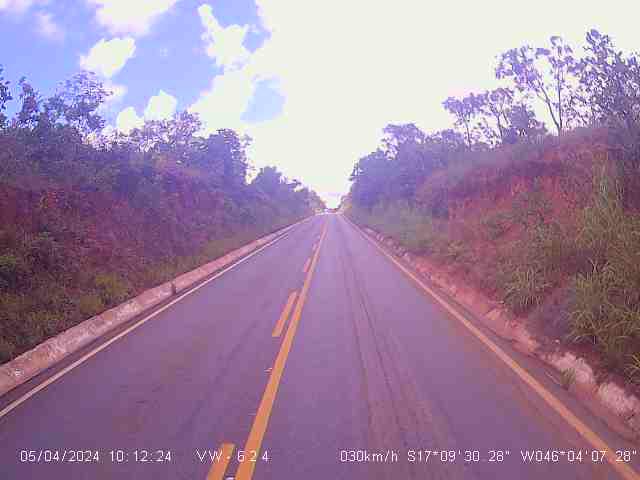}
\end{subfigure}
\hfill
\begin{subfigure}{0.48\columnwidth}
    \centering
    \includegraphics[width=\columnwidth]{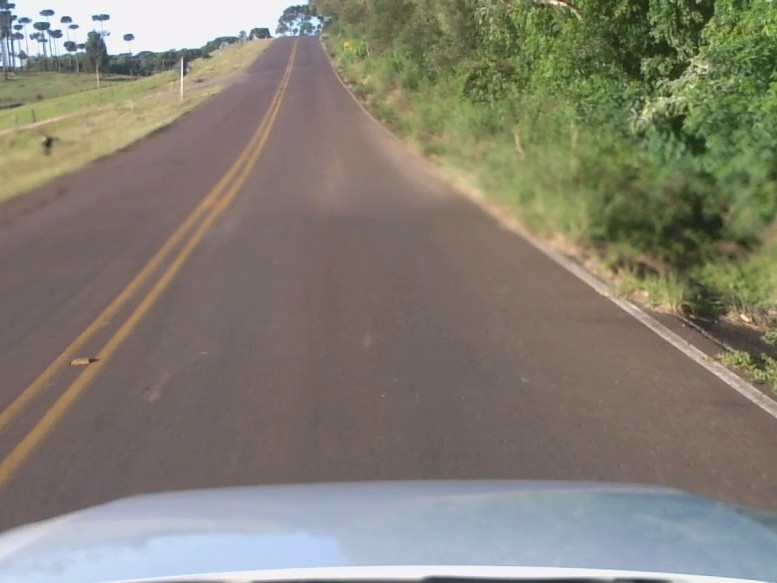}
\end{subfigure}
\begin{subfigure}{0.48\columnwidth}
    \centering
    \includegraphics[width=\columnwidth]{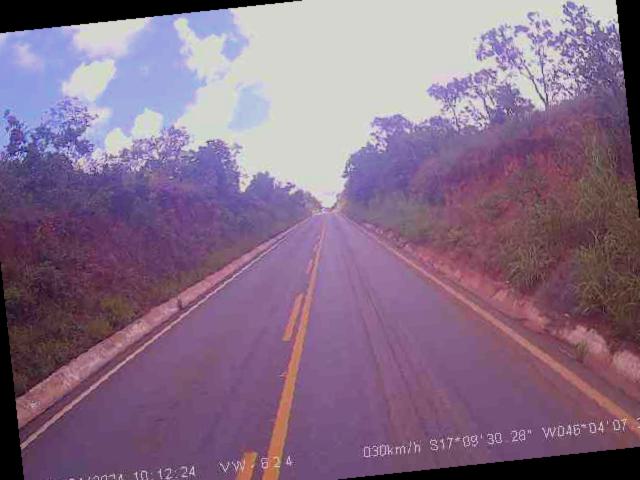}
    \caption{}
\end{subfigure}
\hfill
\begin{subfigure}{0.48\columnwidth}
    \centering
    \includegraphics[width=\columnwidth]{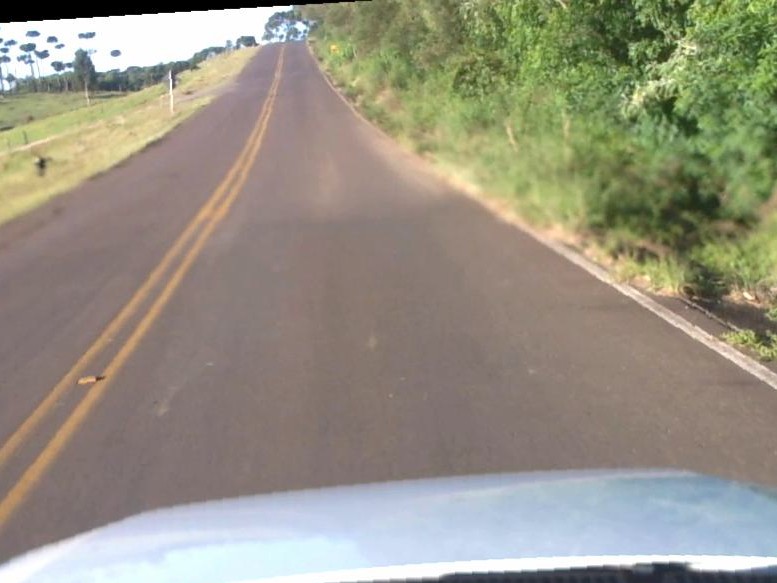}
    \caption{}
\end{subfigure}
\caption{Examples of basic preprocessing operations applied to the input RGB frames. The top row shows the original images, while the bottom row illustrates simple transforms such as resizing, center cropping, and mild color/geometry adjustments, common on computer vision pipelines.}
	\label{fig:vision-transforms}
\end{figure}

\begin{figure}[h!]
\centering

\begin{subfigure}{0.48\columnwidth}
    \centering
    \includegraphics[width=\columnwidth]{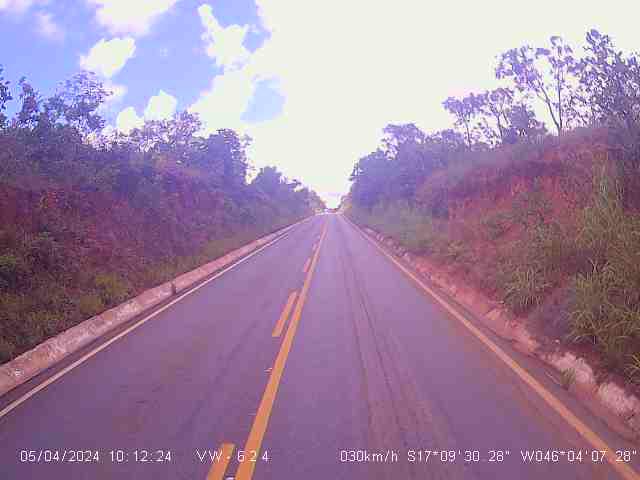}
\end{subfigure}
\hfill
\begin{subfigure}{0.48\columnwidth}
    \centering
    \includegraphics[width=\columnwidth]{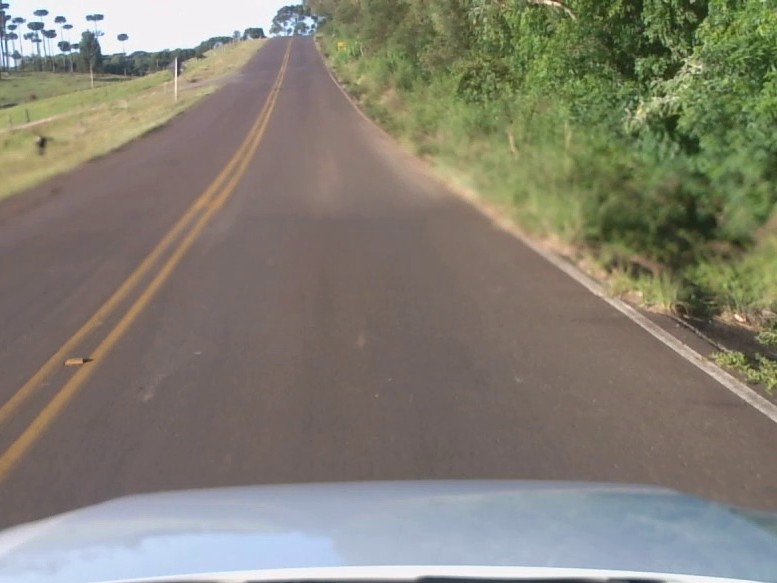}
\end{subfigure}

\begin{subfigure}{0.48\columnwidth}
    \centering
    \includegraphics[width=\columnwidth]{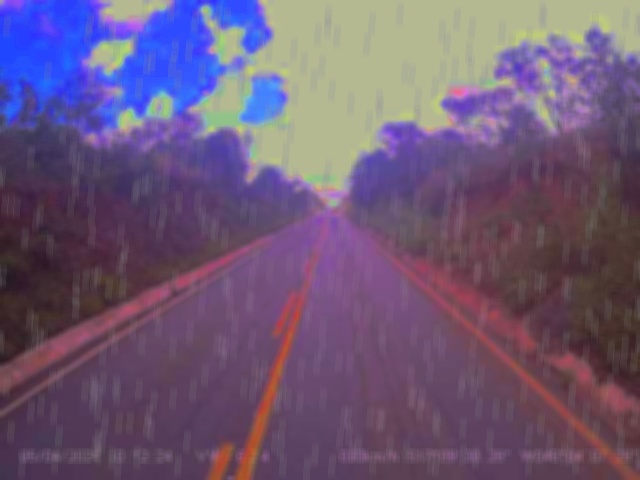}
    \caption{}
\end{subfigure}
\hfill
\begin{subfigure}{0.48\columnwidth}
    \centering
    \includegraphics[width=\columnwidth]{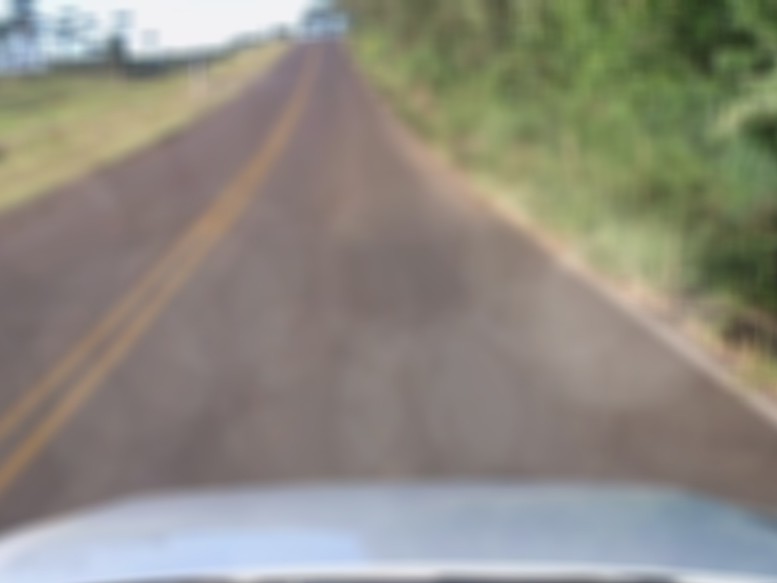}
    \caption{}
\end{subfigure}
\caption{Examples of data augmentation using the Automold library. The top row shows original images, and the bottom row displays augmented versions that simulate environmental effects such as brightness variation, rain, a dirty lens, and motion distortion.}
	\label{fig:vision-transforms-automold}
\end{figure}

\paragraph{Vision preprocessing and augmentation.} We apply standard Computer Vision transforms to the vision module, resizing the image to $256\times256$ and applying (a) center crop, (b) rotation, (c) motion blur, and (d) color jittering. These transforms were empirically selected to increase robustness against (a) variations in input resolution (e.g., Full HD versus HQ cameras), (b) differences in camera mounting positions, (c) vehicle motion speed, and (d) sensor quality or color temperature deviations. Additionally, we use the Automold\footnote{Available at \url{https://github.com/UjjwalSaxena/Automold--Road-Augmentation-Library}} library to simulate environmental effects such as brightness variation, shadows, rain, fog, solar flare, and speed distortion, each applied with a $70\%$ probability. Examples of these augmentations are illustrated in Figures~\ref{fig:vision-transforms} and~\ref{fig:vision-transforms-automold}.

These transformations explicitly address the appearance variability identified in \autoref{sec:introduction} by simulating changes in input resolution, color temperature deviations, shadows, glare, and weather-related visibility distortions. In doing so, they encourage the vision stream to learn representations that remain consistent across diverse lighting and environmental conditions.

\paragraph{IMU preprocessing and augmentation.} For the inertial signals, we apply temporal domain augmentations that enhance robustness to differences in sensor placement and calibration. These include random jittering, scaling, and magnitude warping, also implemented using Automold. Each transformation perturbs the temporal or amplitude characteristics of the signal while preserving its underlying vibration patterns. This directly addresses sensor-quality and calibration variability, ensuring that the learned representation generalizes across different IMU placements, hardware configurations, and calibration states.

\subsection{Fusion module}
\label{subsec:fusion-module}

This module complements the robustness objectives introduced in \autoref{sec:introduction} by integrating the variability-mitigation strategies of the previous sections. It fuses the intermediate representations from the visual and inertial streams in an \textit{adaptive} manner, adjusting the contributions of each modality based on signal quality and contextual reliability. To achieve this, we employ a lightweight bidirectional cross-attention mechanism followed by an element-wise gating function.

\paragraph{Tokenization of modality-specific embeddings.}
While the visual and inertial encoders each produce a single global embedding,
token-level interactions are required for the cross-attention mechanism.
To enable this without relying on spatial feature maps or raw temporal
sequences, we introduce a lightweight tokenizer for each modality. 
Given the global embeddings 
$f(I) \in \mathbb{R}^{D_{\text{vis}}}$ and 
$f(S) \in \mathbb{R}^{D_{\text{imu}}}$,
the tokenizers apply a layer-normalized linear projection followed by a reshape,
expanding each embedding into $n$ latent tokens of dimension $d$:
\begin{align}
V &= W_{\text{vis}}\,\mathrm{LN}(f(I)) + b_{\text{vis}}, \\
A &= W_{\text{imu}}\,\mathrm{LN}(f(S)) + b_{\text{imu}},
\end{align}
where each linear projection is subsequently reshaped into $n$ latent tokens of dimension $d$. This transformation yields small sets of modality-specific latent tokens that serve as the query, key, and value inputs for subsequent bidirectional cross-attention. In practice, we use $n=6$ tokens per modality and $d=512$ as the shared latent dimension.

\paragraph{Cross-attention.}
Given the visual tokens $V \in \mathbb{R}^{n \times d}$ and inertial tokens 
$A \in \mathbb{R}^{n \times d}$ produced by the tokenizers, we align the two modalities through a bidirectional cross-attention mechanism. In the first direction, the vision tokens query the inertial tokens to identify 
motion patterns consistent with the observed scene; in the second direction, the inertial tokens query the vision tokens to localize visual features that explain the measured vibrations. Formally,
\begin{align}
V' &= \text{MSA}(Q = V, K = A, V = A), \\
A' &= \text{MSA}(Q = A, K = V, V = V),
\end{align}
where $\text{MSA}(\cdot)$ denotes a multi-head attention block with residual and feed-forward layers configured for cross-modal queries. 
This bidirectional exchange enables each modality to refine its representation using complementary cues from the other before the pooling and gating stages.

\begin{table*}[t]
\centering
\caption{Overview of the \textbf{ROAD} dataset. Each subset differs in modality, duration, sampling characteristics, and class distribution between \textbf{A}sphalt, \textbf{B}elgian blocks, and \textbf{O}ff-road, enabling complementary evaluation of multimodal fusion, vision-only robustness, and out-of-distribution generalization.}
\label{tab:road-overview}

\begin{tabular}{lllllll}
\toprule
\textbf{Subset} & 
\textbf{Modality} & 
\textbf{Duration} & 
\textbf{Sampling} & 
\textbf{Frames} & 
\textbf{Dist. (\%)} & 
\textbf{Purpose} \\
\midrule

\#1 Sensor Fusion &
\makecell[l]{Camera +\\5$\times$IMU} &
10h 40m &
\makecell[l]{30\,fps\\400\,Hz} &
$\approx$1.15M &
\makecell[l]{\textbf{A}: 79.67\\\textbf{B}: 8.55\\\textbf{O}: 11.78} &
\makecell[l]{Multimodal learning and\\ correlation studies} \\
\midrule
\#2 Vision Only &
RGB images &
13h 34m &
30\,fps &
$\approx$1.47M &
\makecell[l]{\textbf{A}: 72.74\\\textbf{B}: 10.53\\\textbf{O}: 16.73} &
\makecell[l]{Visual robustness under\\adverse conditions} \\
\midrule
\#3 Synthetic &
Simulation &
53m &
30\,fps &
$\approx$95.4k &
\makecell[l]{\textbf{A}: 61.3\\\textbf{B}: 6.6\\\textbf{O}: 32.1} &
\makecell[l]{Out-of-distribution\\generalization} \\
\bottomrule
\end{tabular}
\end{table*}

\paragraph{Pooling and gating.}
Following the bidirectional cross-attention, each modality outputs a set of context-enriched tokens 
$V' \in \mathbb{R}^{n \times d}$ and $A' \in \mathbb{R}^{n \times d}$. To summarize these token sets into single representative vectors, we employ a learnable attention-based pooling operation. 
For each modality, scalar attention weights are produced through a linear projection and normalized via a softmax function:
\begin{equation}
w_v = \text{softmax}(W_p V'), \quad 
w_a = \text{softmax}(W_p A'),
\end{equation}
and the pooled representations are then obtained as:
\begin{equation}
v^* = \sum_i w_{v,i} V'_i, \quad
a^* = \sum_i w_{a,i} A'_i,
\end{equation}
where $W_p$ denotes the learnable projection matrix.
These pooled vectors are subsequently fused through an adaptive gating function:
\begin{equation}
g = \sigma(W_g [v^*; a^*] + b_g), \quad 
z = g \odot v^* + (1 - g) \odot a^*,
\end{equation}
where $[v^*; a^*]$ denotes concatenation, $\odot$ the element-wise product, and $\sigma(\cdot)$ the sigmoid activation.
The gate vector $g \in (0,1)^d$ dynamically balances the contributions of each modality for each latent dimension, enabling the model to adjust its reliance on vision or inertial cues based on context.

This fusion mechanism is inherently \textit{adaptive}: when one modality becomes unreliable or out-of-distribution (e.g., during low visibility or high sensor noise), the gating function suppresses its contribution, allowing the other modality to dominate. Such behavior is essential to ensure robustness for real-world applications.

\section{The \textit{ROAD} Dataset}\label{sec:dataset}

To support research on multimodal RSC in Latin American contexts, we introduce the \textit{ROAD} dataset. It comprises real-world and synthetic data, explicitly designed for sensor fusion between cameras and inertial sensors. This dataset is composed of three subsets, each serving a distinct research purpose while remaining complementary in coverage and scenario diversity. The dataset is available at \url{https://road-dataset.github.io}. We detail the subsets below.

\subsection{Multimodal sensor fusion dataset}
\label{road:subset-1}

\begin{figure}[h!]
\centering

\begin{subfigure}{0.32\columnwidth}
    \centering
    \includegraphics[width=\columnwidth]{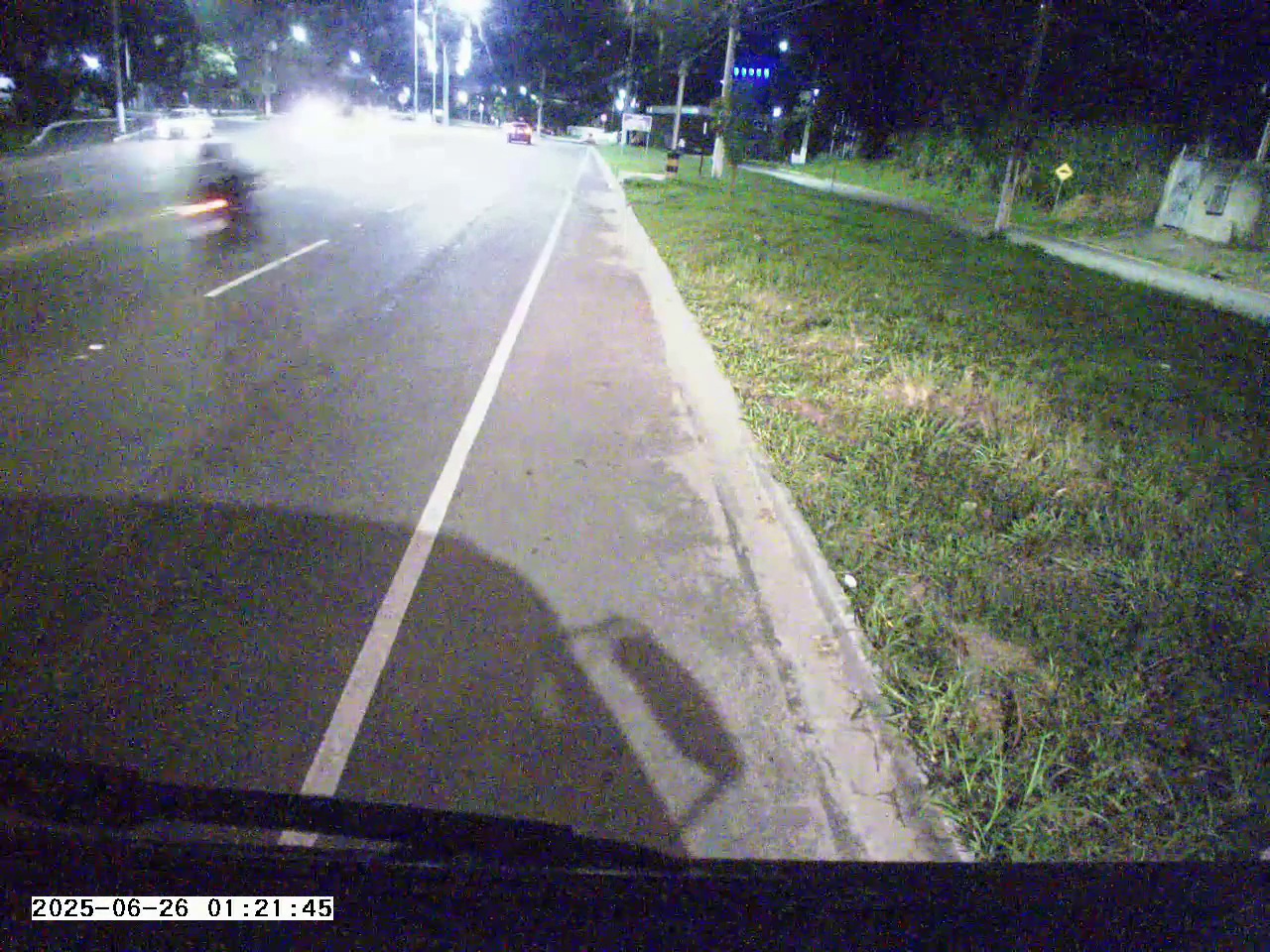}
\end{subfigure}
\hfill
\begin{subfigure}{0.32\columnwidth}
    \centering
    \includegraphics[width=\columnwidth]{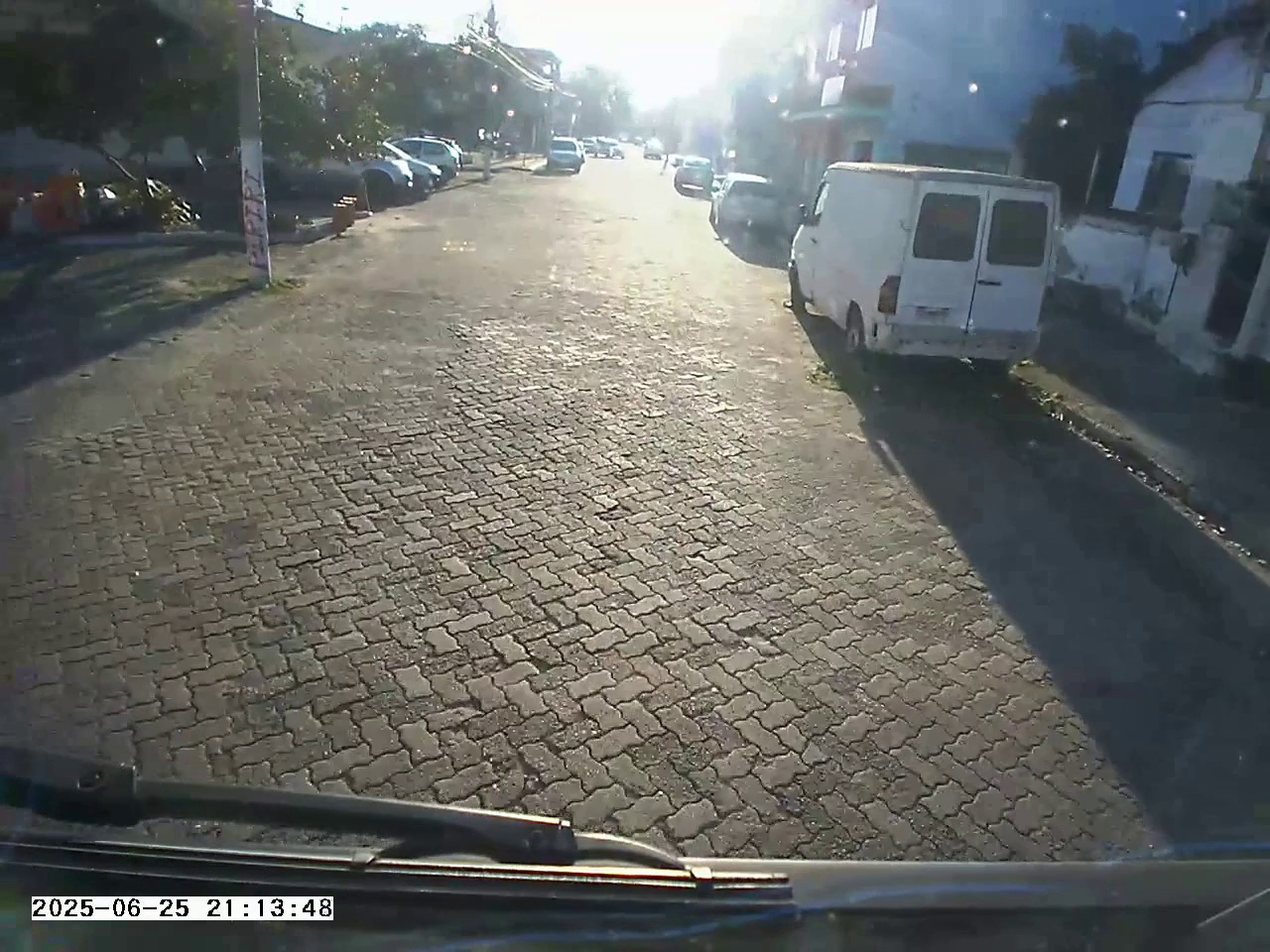}
\end{subfigure}
\hfill
\begin{subfigure}{0.32\columnwidth}
    \centering
    \includegraphics[width=\columnwidth]{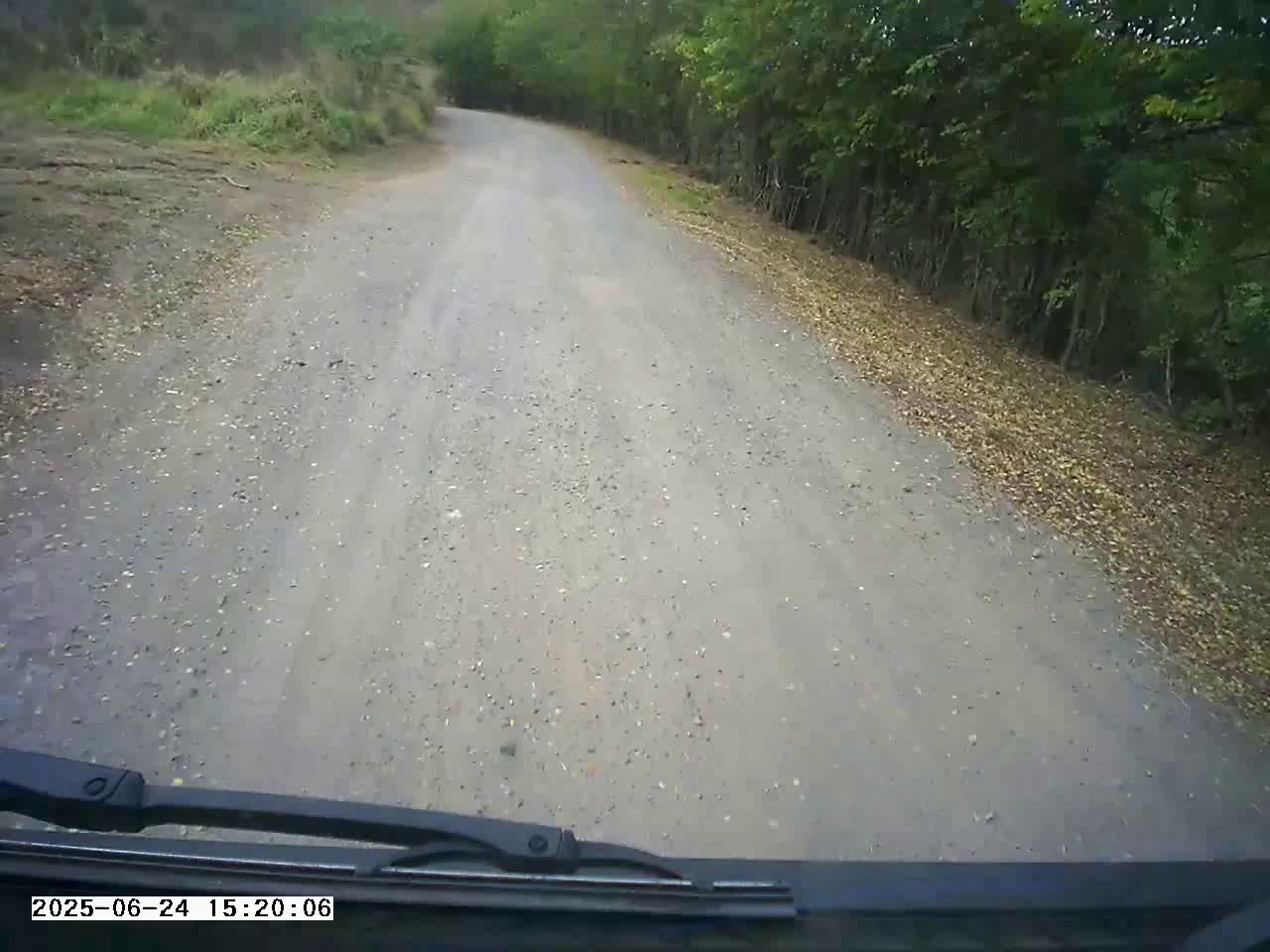}
\end{subfigure}

\vspace{0.4em}

\begin{subfigure}{0.32\columnwidth}
    \centering
    \includegraphics[width=\columnwidth]{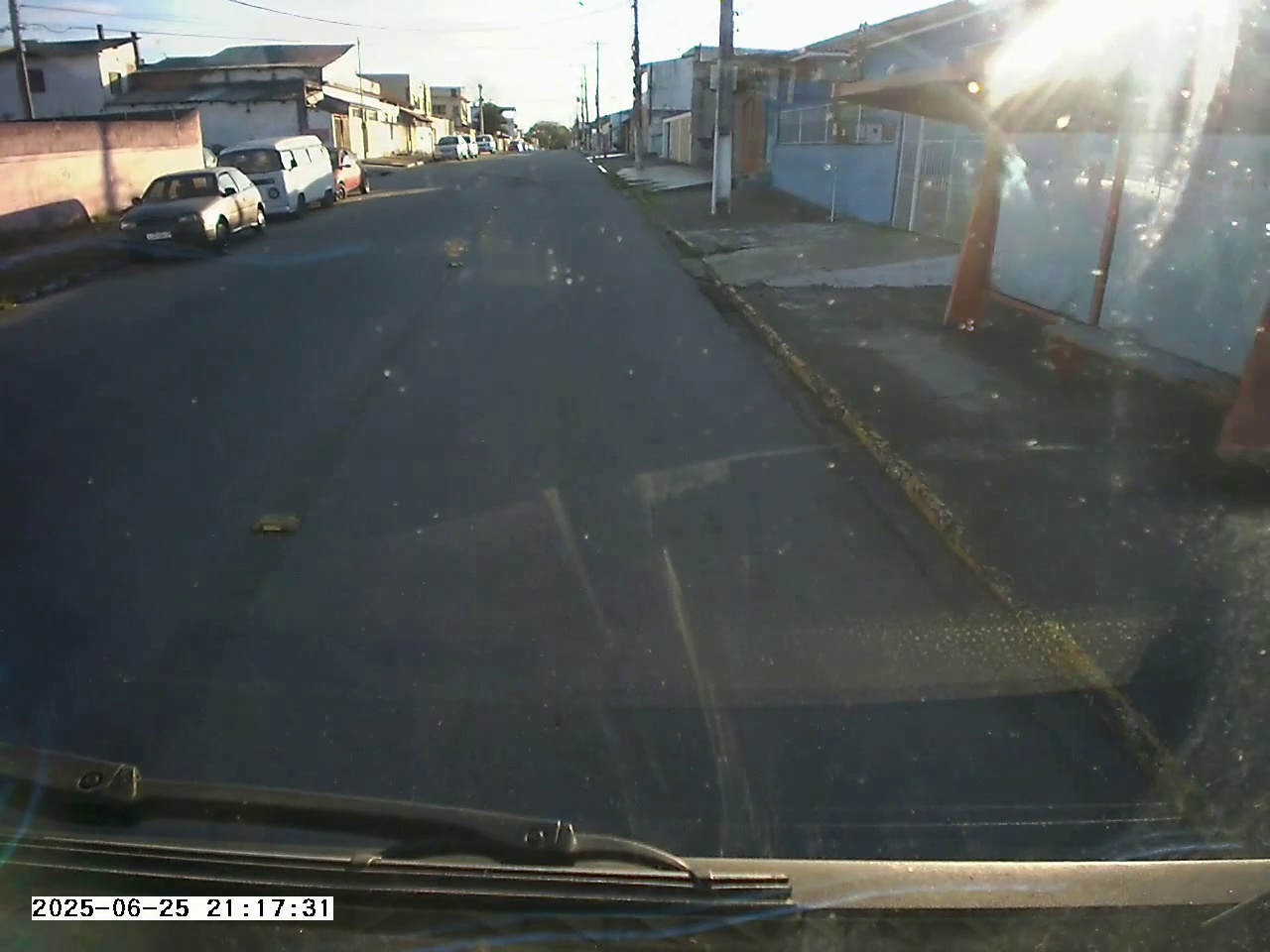}
    \caption{Asphalt}
\end{subfigure}
\hfill
\begin{subfigure}{0.32\columnwidth}
    \centering
    \includegraphics[width=\columnwidth]{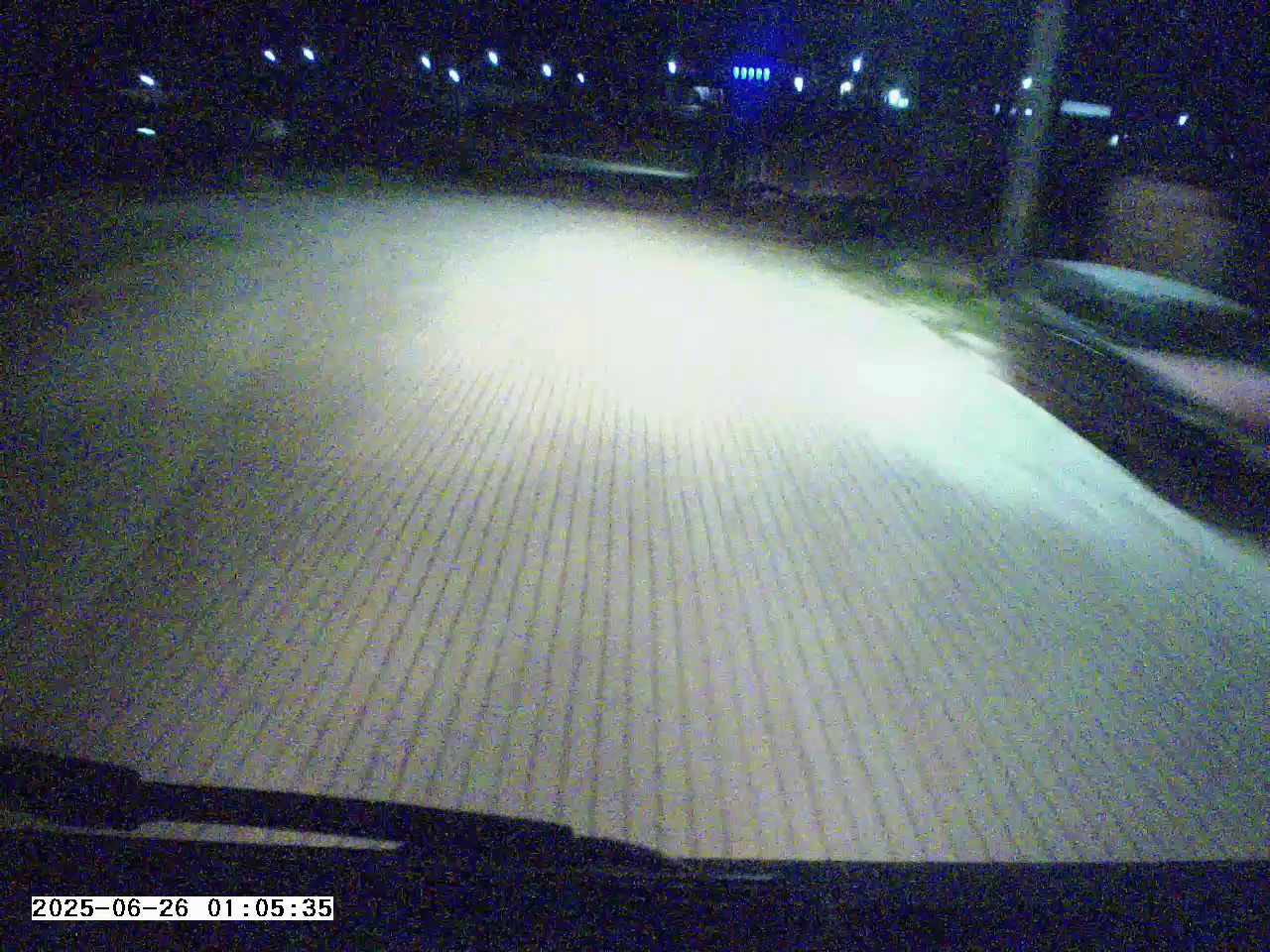}
    \caption{Belgian Blocks}
\end{subfigure}
\hfill
\begin{subfigure}{0.32\columnwidth}
    \centering
    \includegraphics[width=\columnwidth]{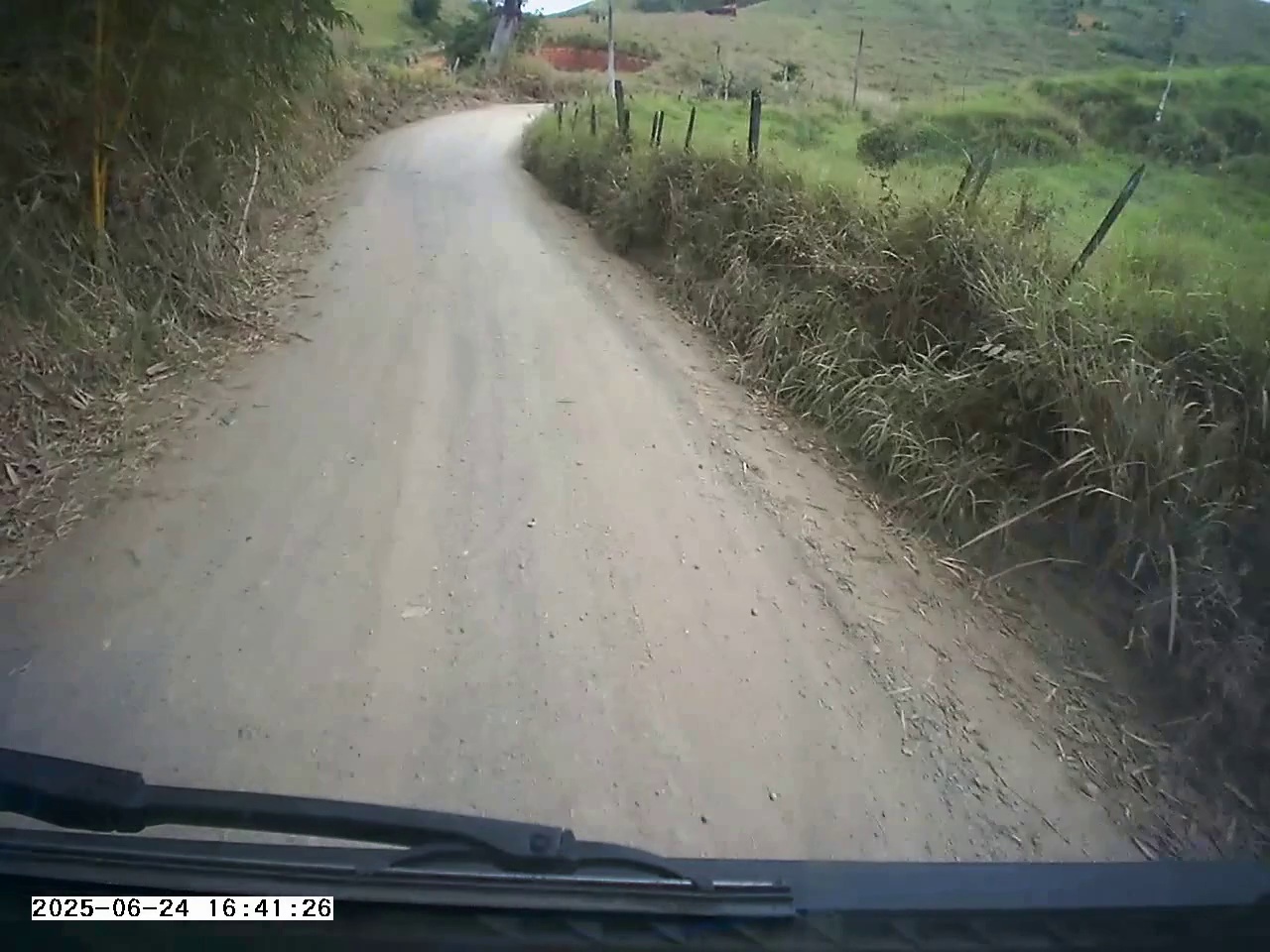}
    \caption{Off-road}
\end{subfigure}
\caption{Representative samples from the proposed multimodal sensor fusion dataset, showcasing the three road surface classes available in our dataset: \textit{Asphalt}, \textit{Belgian Blocks}, and \textit{Off-road}. Images were captured on public roads using a fixed, industry-standard camera mounting configuration.}
	\label{fig:multimodal-dataset}
\end{figure}

Subset~\#1 was collected using a Volkswagen Constellation test truck operating on public roads in multiple cities in the state of Rio de Janeiro, Brazil, capturing diverse road surfaces and also including challenging scenarios, such as rain, night, and a combination of night and rain. We show samples of this dataset on \autoref{fig:multimodal-dataset}. This subset was collected on the span of multiple days, with multiple professional test drivers from the Volkswagen Truck and Bus company. Data were recorded using both an RGB camera \textit{Sony SNCCH210/B} sampled at 30~fps and five \textit{VC MEMS 3711F11} inertial sensors sampled at 400~Hz, all timestamp-aligned through the \textit{Siemens Simcenter SCADAS Recorder}, a golden standard industry-level datalogger. The total recording time amounts to approximately 10 hours and 40 minutes.

\begin{figure}[b!]
	\centering
		\includegraphics[width=\columnwidth]{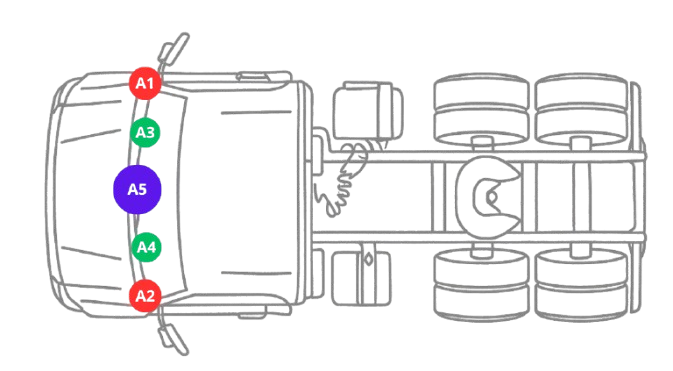}
	\caption{Schematic visualization of the placement of the five IMU units around the truck. The truck illustration was generated using a Generative AI tool for clarity and does not correspond to a specific vehicle model.}
	\label{fig:camera-placements}
\end{figure}

We've placed five IMU units strategically positioned to capture vehicle dynamics and vibration patterns across the truck structure. We detail below the locations of the sensors, as we also show in \autoref{fig:camera-placements}.

\begin{itemize}
    \item A1 – Front axle (right side)
    \item A2 – Front axle (left side)
    \item A3 – Side member (right side)
    \item A4 – Side member (left side)
    \item A5 – Inside the cabin
\end{itemize}

Frame-level annotations were produced using \url{CVAT.ai}, with two annotators performing the initial labeling and validation conducted by a team of five R\&D engineers and researchers. The data on this subset is divided into \textit{Asphalt} (79.67\%), \textit{Belgian Blocks} (8.55\%), and \textit{Off-road} (11.78\%)\footnote{We adopt a different class nomenclature from PVS, which defines \textit{Asphalt}, \textit{Cobblestone}, and \textit{Dirt road}, as our dataset covers a broader range of surface appearances. In particular, not all \textit{Belgian Blocks} correspond to \textit{Cobblestone} surfaces, and not all \textit{Dirt roads} can be considered \textit{Off-road}.}. This collection provides a foundation for multimodal learning, enabling research that explores correlations between visual cues and vibration signatures.

\subsection{Vision-only dataset}
\label{road:subset-2}

\begin{figure}[h!]
\centering

\begin{subfigure}{0.32\columnwidth}
    \centering
    \includegraphics[width=\columnwidth]{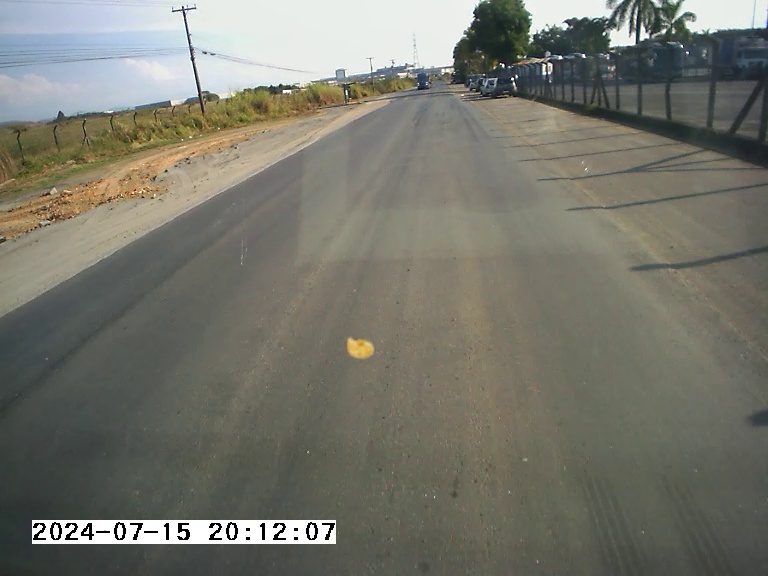}
\end{subfigure}
\hfill
\begin{subfigure}{0.32\columnwidth}
    \centering
    \includegraphics[width=\columnwidth]{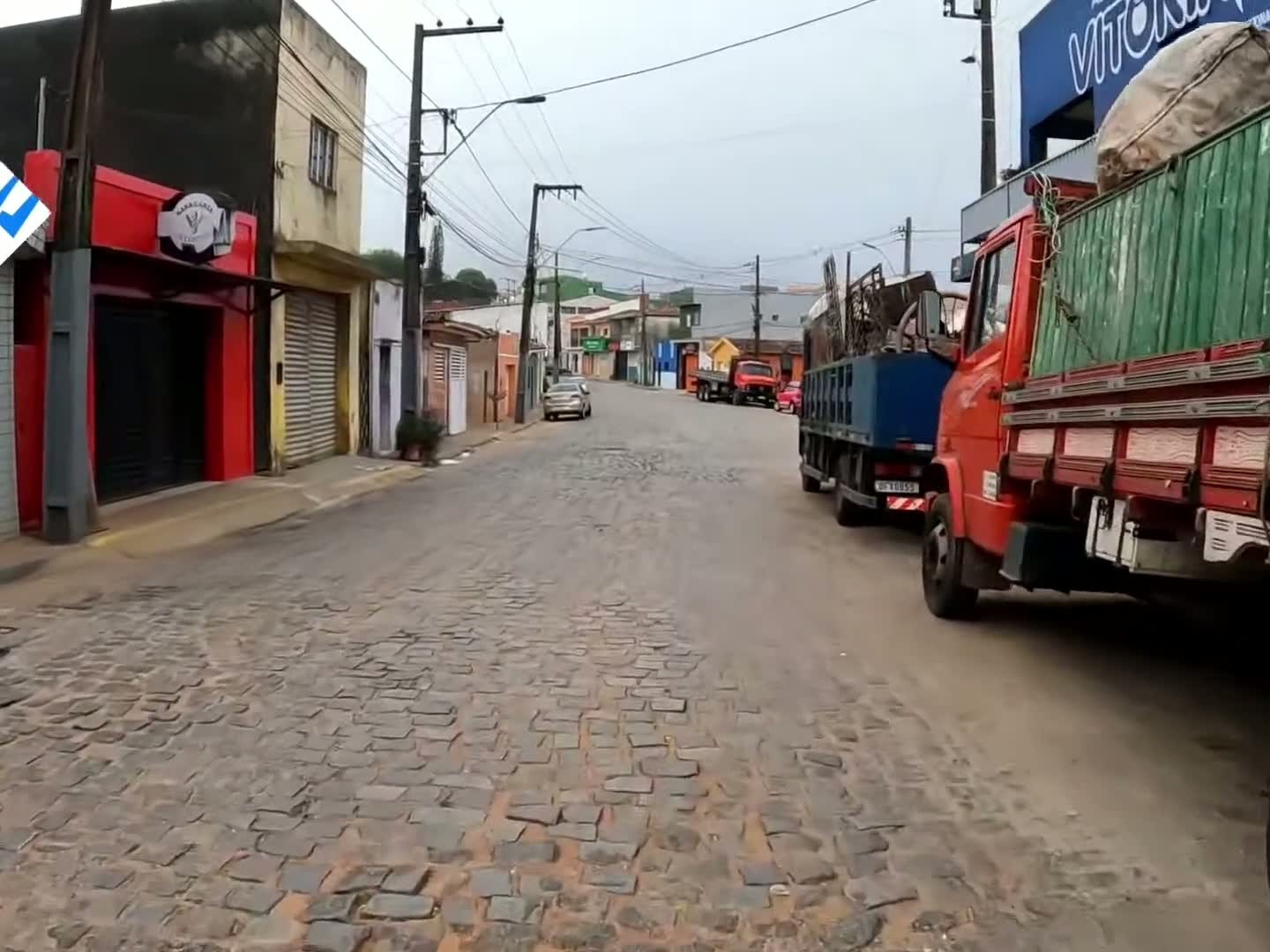}
\end{subfigure}
\hfill
\begin{subfigure}{0.32\columnwidth}
    \centering
    \includegraphics[width=\columnwidth]{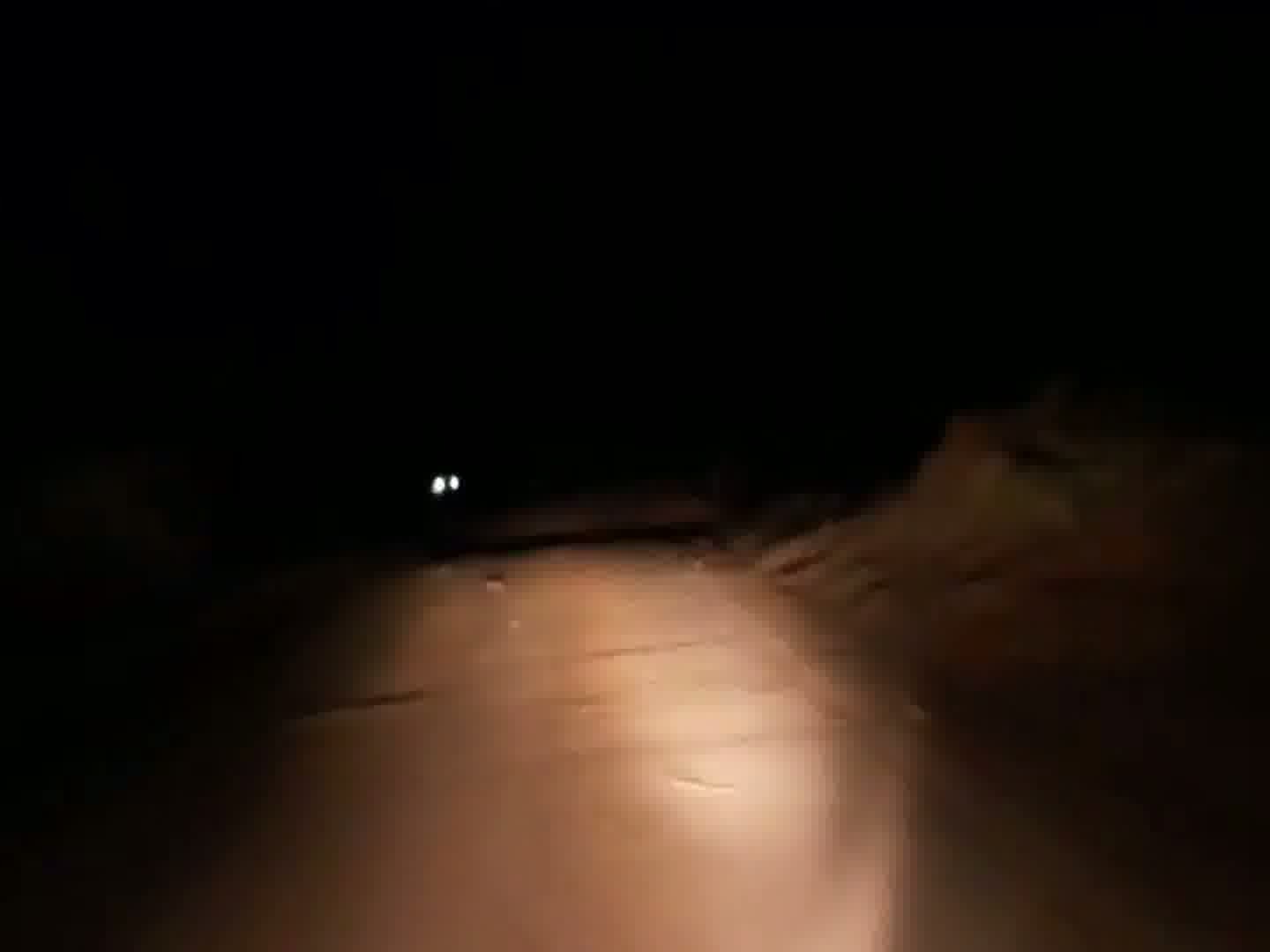}
\end{subfigure}

\vspace{0.4em}

\begin{subfigure}{0.32\columnwidth}
    \centering
    \includegraphics[width=\columnwidth]{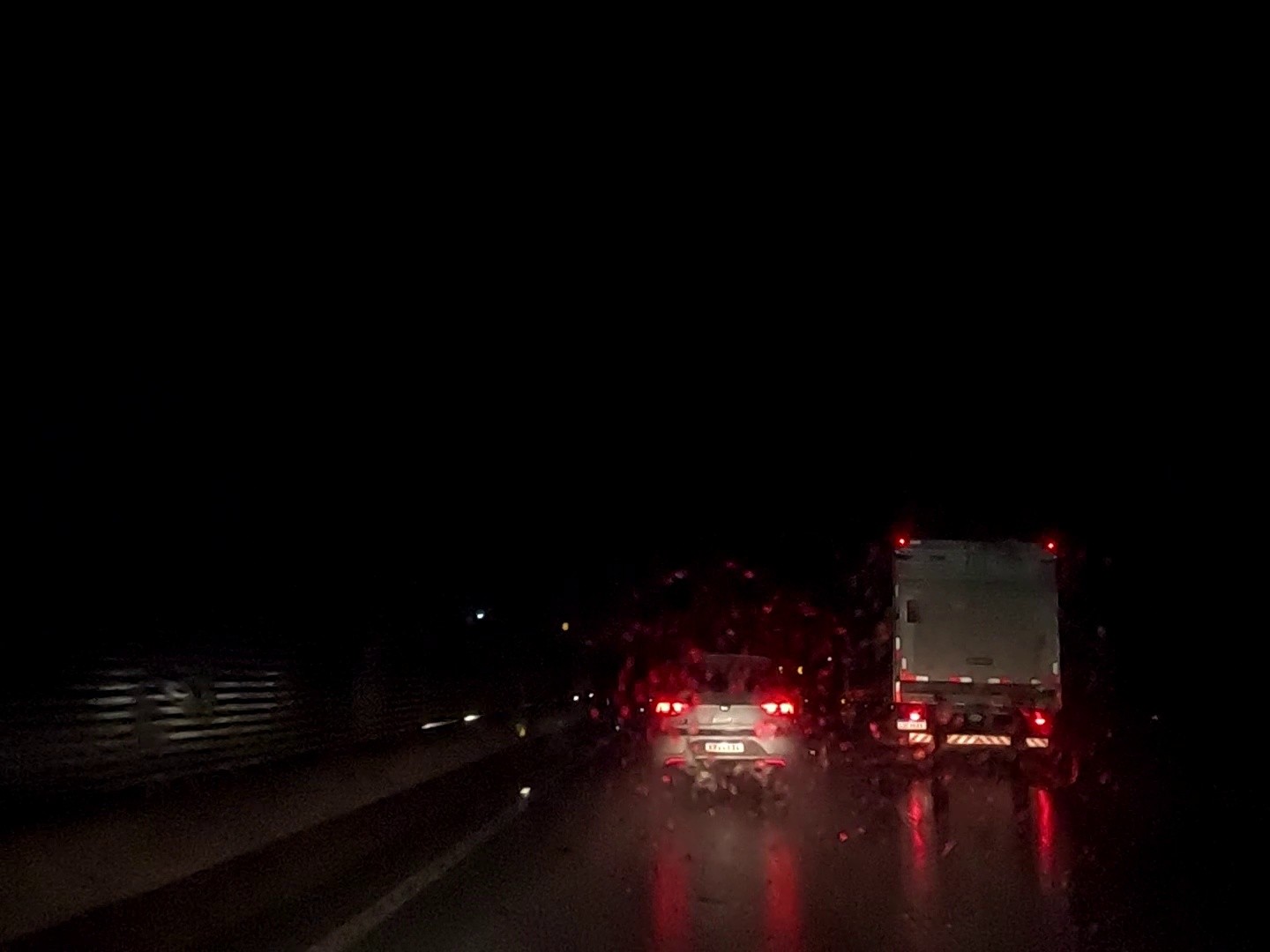}
    \caption{Asphalt}
\end{subfigure}
\hfill
\begin{subfigure}{0.32\columnwidth}
    \centering
    \includegraphics[width=\columnwidth]{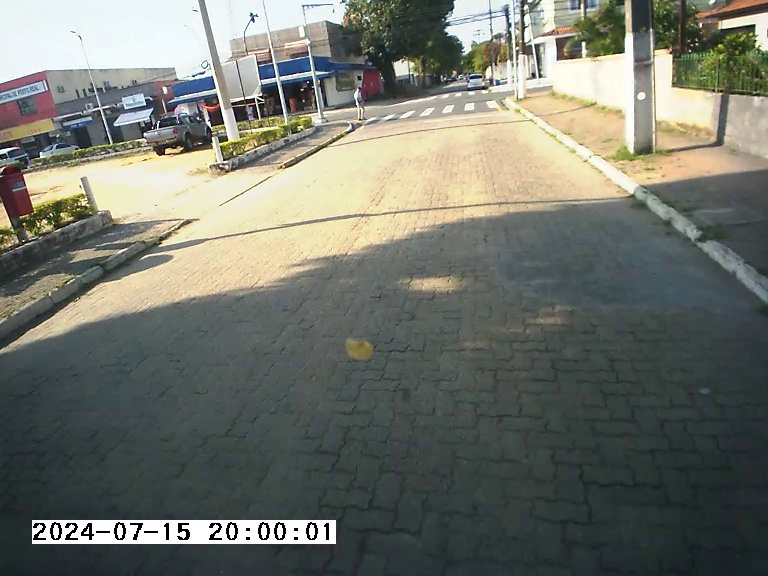}
    \caption{Belgian Blocks}
\end{subfigure}
\hfill
\begin{subfigure}{0.32\columnwidth}
    \centering
    \includegraphics[width=\columnwidth]{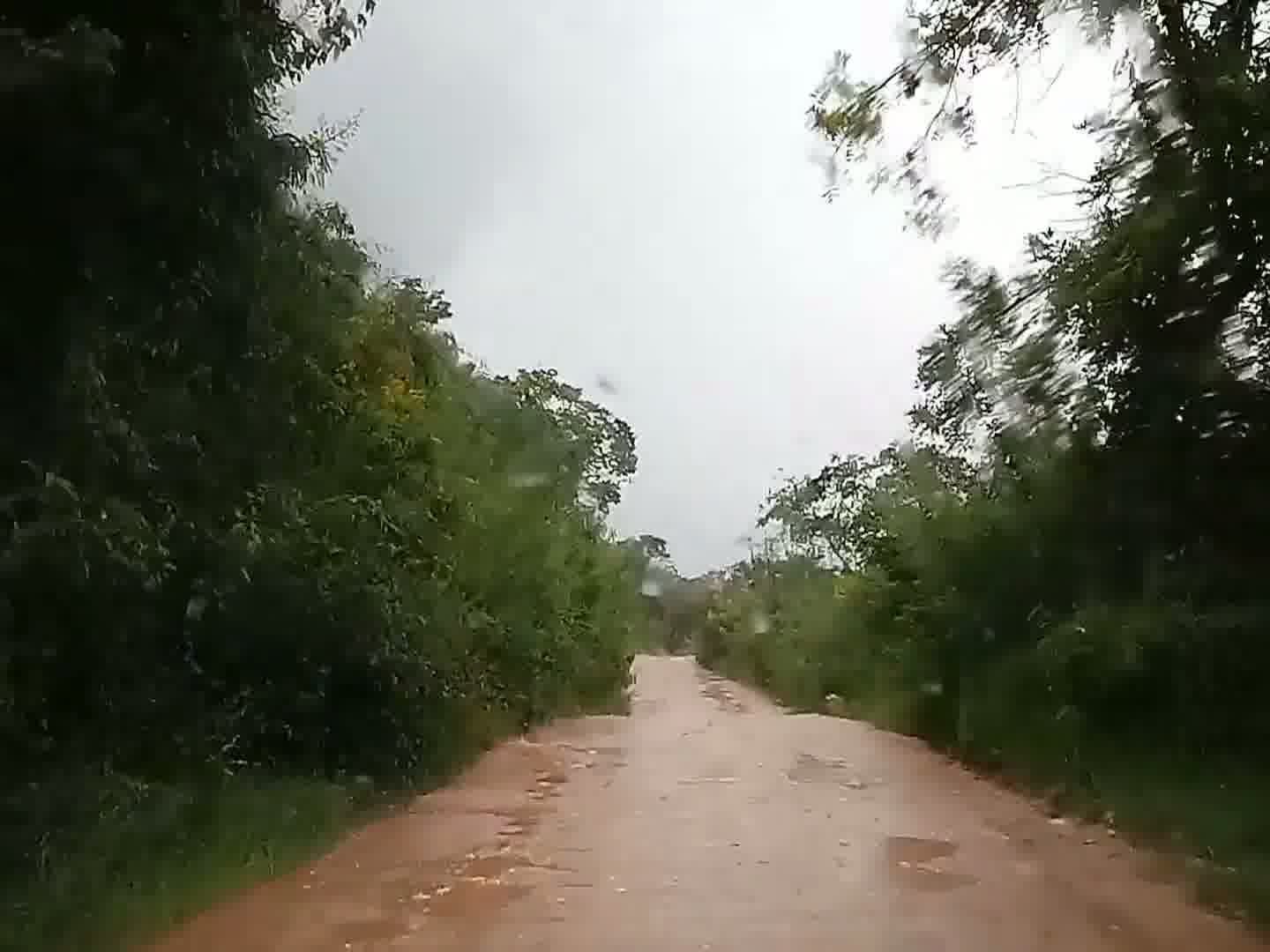}
    \caption{Off-road}
\end{subfigure}
\caption{Representative samples from the vision-only subset, illustrating the three road surface classes (\textit{Asphalt}, \textit{Belgian Blocks}, and \textit{Off-road}) under diverse illumination and weather conditions. This subset is designed to evaluate vision-only performance in challenging real-world scenarios.}
	\label{fig:vision-dataset}
\end{figure}

Subset~\#2 focuses exclusively on visual information, especially in challenging scenarios, allowing evaluations to answer questions such as ``\textit{would a vision-only model perform well under darker settings?}''. Therefore, this subset is strongly focused on these challenging conditions. It is composed of 1,465,200 frames, totaling about 13 hours and 34 minutes of data. Recordings cover diverse conditions, including rain, night, rain~+~night, different Asphalt appearances, Belgian block structures, and off-road rural environments. We show samples of this subset in \autoref{fig:vision-dataset}.

The data were collected across multiple regions in Brazil and enriched with point-of-view driving videos from across Latin America, increasing geographic and infrastructural diversity. All external point-of-view recordings were obtained from publicly available sources with permissive licenses and contain only road-front view scenes without identifying individuals. Recordings were made with heterogeneous devices, including dashcams, GoPro Hero 13 Black units, and smartphone cameras, capturing realistic variability in field conditions. Annotations followed the same dual-annotation and validation protocol as Subset~\#1. The data on this subset is divided into \textit{Asphalt} (72.74\%), \textit{Belgian Blocks} (10.53\%), and \textit{Off-road} (16.73\%). This subset complements the sensor-fusion dataset by providing a broader benchmark for evaluating vision-only perception models, especially under adverse illumination and weather conditions.

\subsection{Synthetic data for out-of-distribution scenarios}
\label{road:subset-3}

\begin{figure}[b!]
\centering

\begin{subfigure}{0.32\columnwidth}
    \centering
    \includegraphics[width=\columnwidth]{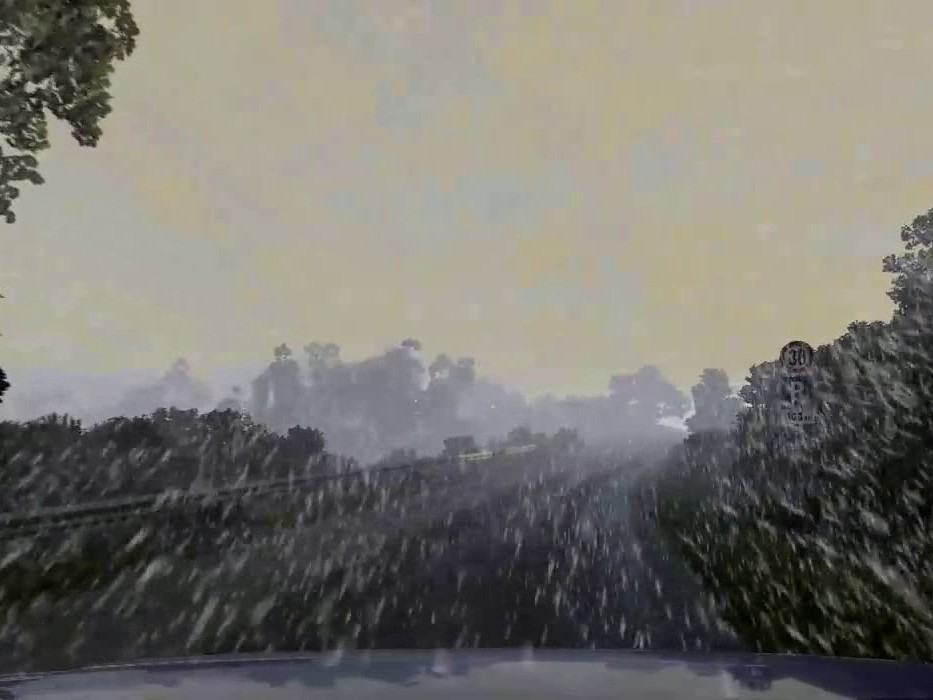}
\end{subfigure}
\hfill
\begin{subfigure}{0.32\columnwidth}
    \centering
    \includegraphics[width=\columnwidth]{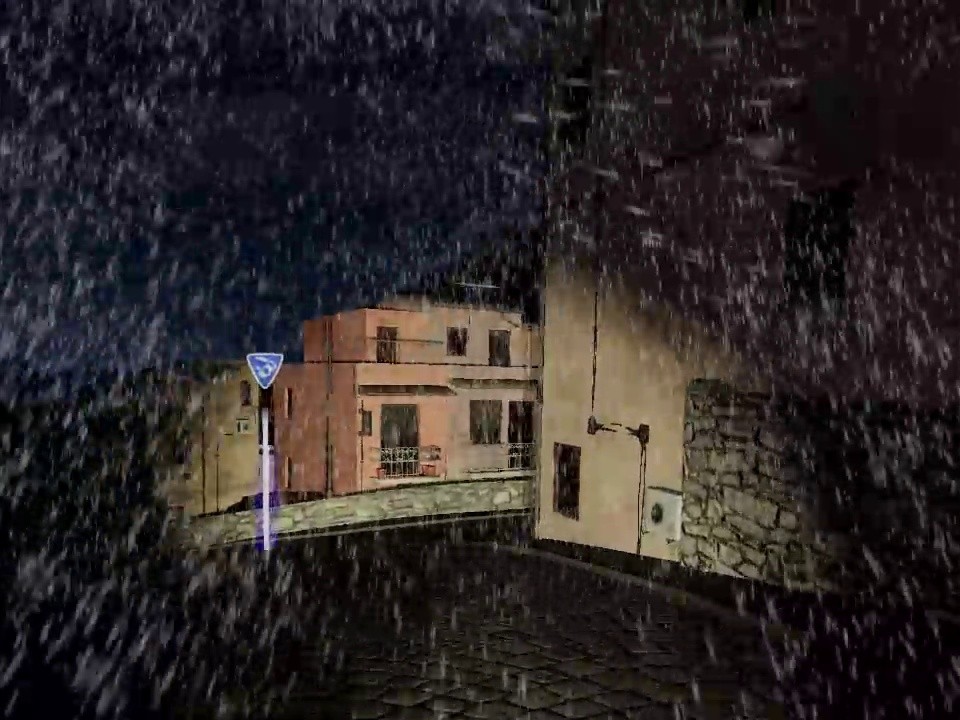}
\end{subfigure}
\hfill
\begin{subfigure}{0.32\columnwidth}
    \centering
    \includegraphics[width=\columnwidth]{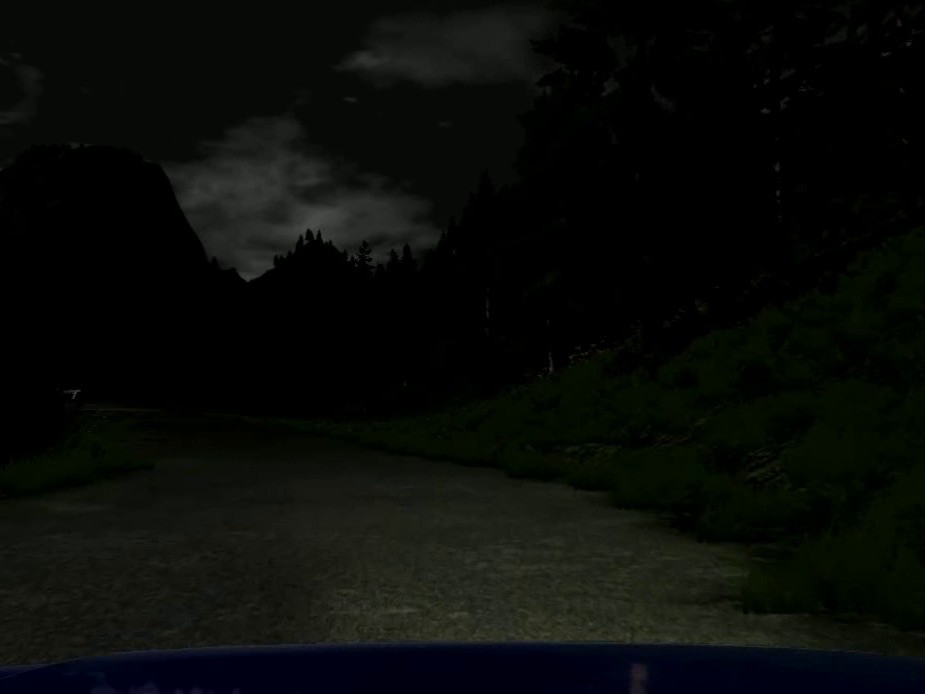}
\end{subfigure}

\vspace{0.4em}

\begin{subfigure}{0.32\columnwidth}
    \centering
    \includegraphics[width=\columnwidth]{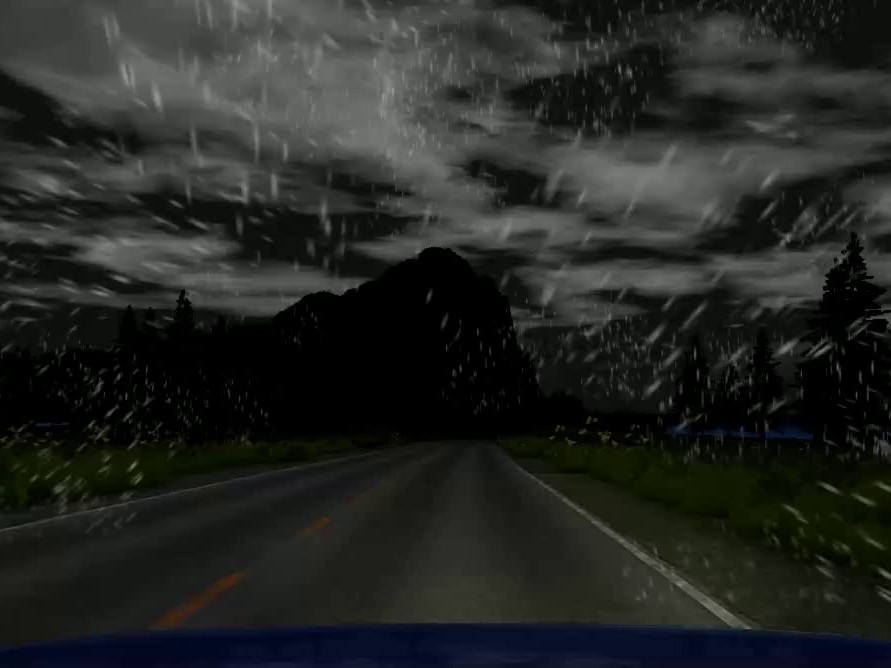}
    \caption{Asphalt}
\end{subfigure}
\hfill
\begin{subfigure}{0.32\columnwidth}
    \centering
    \includegraphics[width=\columnwidth]{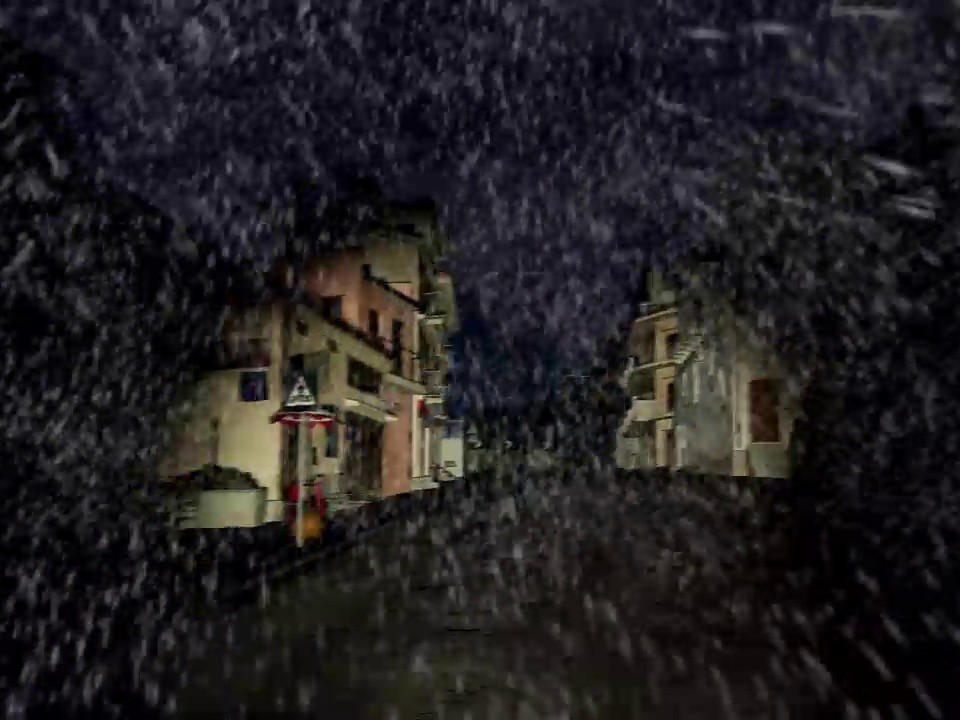}
    \caption{Belgian Blocks}
\end{subfigure}
\hfill
\begin{subfigure}{0.32\columnwidth}
    \centering
    \includegraphics[width=\columnwidth]{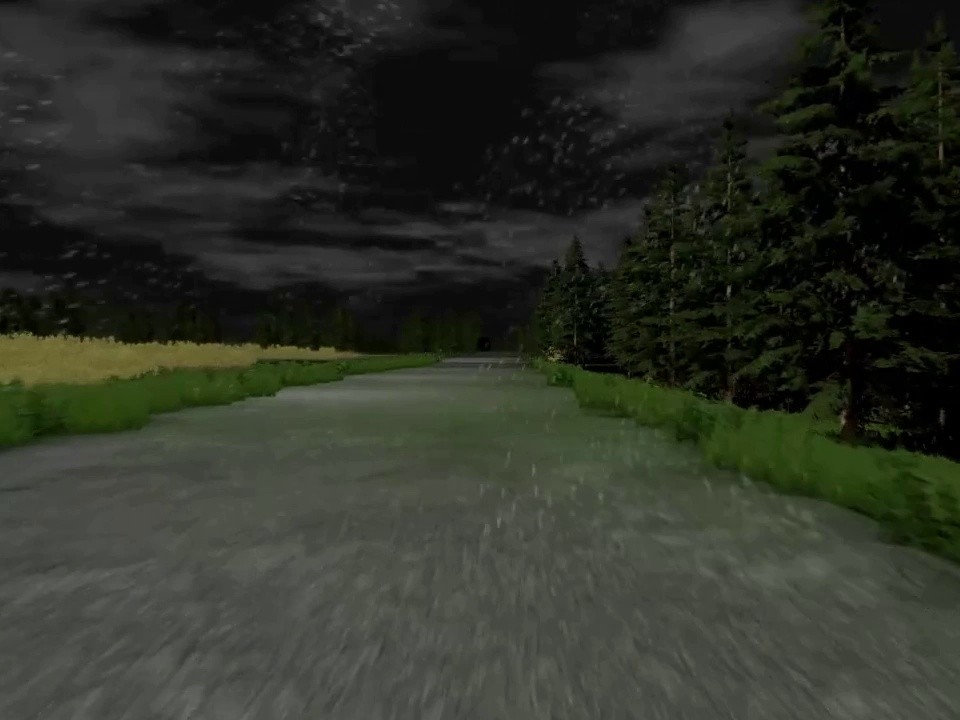}
    \caption{Off-road}
\end{subfigure}
\caption{Representative samples from the synthetically generated subset, illustrating the three road surface classes (Asphalt, Belgian Blocks, and Off-road) under extreme environmental conditions such as heavy rain and low-illumination scenarios.}
	\label{fig:synthetic-dataset}
\end{figure}

Subset~\#3 was developed to address environmental conditions that are difficult to capture consistently in real-world environments, such as challenging night-rain conditions. Using the \url{BeamNG.tech} simulation platform, we generated synthetic scenes to maintain visual coherence and controlled environmental diversity.

This subset comprises approximately 53~minutes of Full~HD recordings and enables controlled experiments on domain adaptation and out-of-distribution robustness. Synthetic generation also provides access to precise ground-truth metadata, including weather, surface type, and lighting conditions, enabling detailed supervision during model training and evaluation. Data in this subset is divided into \textit{Asphalt} (61.3\%), \textit{Belgian Blocks} (6.6\%), and \textit{Off-road} (32.1\%). We show samples of this subset in \autoref{fig:synthetic-dataset}.

\sisetup{
  round-mode=places,
  round-precision=1,
  table-number-alignment=center,
  detect-weight=true,
  detect-family=true
}

\begin{table*}[h!]
\centering
\caption{Quantitative comparison of our proposed multimodal framework against two state-of-the-art baselines on the PVS dataset and on Subset \#1 of the ROAD dataset. Results are reported as overall accuracy and per-class F1-scores (\textuparrow  indicates higher is better). The table highlights the strong performance of our method across both benchmarks, with particularly notable gains on the more challenging classes of the ROAD dataset.}
\label{tab:results}
\begin{tabular}{
    l
    c
    l S[table-format=2.1]
    c
    l S[table-format=3.1]
}
\toprule
\multirow{2}{*}{\textbf{Method}} &
  \multicolumn{3}{c}{\textbf{PVS Dataset}} &
  \multicolumn{3}{c}{\textbf{ROAD Dataset} (Subset \#1)} \\
\cmidrule(lr){2-4} \cmidrule(lr){5-7}
& {\textbf{Acc. (\%) \textuparrow}} & \multicolumn{2}{c}{\textbf{F1-score (\%)  \textuparrow}} 
& {\textbf{Acc. (\%) \textuparrow}} & \multicolumn{2}{c}{\textbf{F1-score (\%)  \textuparrow}} \\
\midrule
\multirow{3}{*}{\footnotesize \citet{menegazzo2021road}} 
  & \multirow{3}{*}{92.7} & Asphalt     & 98.6 & \multirow{3}{*}{87.0} & Asphalt        & 93.1 \\
  &                       & Cobblestone & 86.1 &                       & Belgian Blocks & 0.0  \\
  &                       & Dirt road   & 90.8 &                       & Off-road       & 0.0  \\
\cmidrule(r){1-7}
\multirow{3}{*}{\citet{van2025hybrid}} 
  & \multirow{3}{*}{94.2} & Asphalt     & 99.0 & \multirow{3}{*}{86.6} & Asphalt        & 92.8 \\
  &                       & Cobblestone & 92.3 &                       & Belgian Blocks & 0.0  \\
  &                       & Dirt road   & 91.2 &                       & Off-road       & 46.3 \\
\cmidrule(r){1-7}
\multirow{3}{*}{Ours} 
  & \multirow{3}{*}{\textbf{95.6}} & Asphalt     & \textbf{99.2} & \multirow{3}{*}{\textbf{98.2}} & Asphalt        & \textbf{98.9} \\
  &                       & Cobblestone & \textbf{92.7} &                       & Belgian Blocks & \textbf{87.6} \\
  &                       & Dirt road   & \textbf{93.7} &                       & Off-road       & \textbf{100.0} \\
\bottomrule
\end{tabular}
\end{table*}

\section{Experiments}
\label{sec:experiments}

This section details the experimental design used to evaluate the proposed multimodal framework. We structure the discussion around four components: (i) experimental goals, (ii) datasets and evaluation protocol, (iii) baseline methods, and (iv) implementation details. This organization enhances clarity and ensures reproducibility, following common practices in computer vision and multimodal learning research.

\subsection{Experimental goals}

Our experiments are designed to answer the following research questions:
\begin{itemize}
    \item \textbf{RQ1 — Cross-dataset generalization:} How does the proposed framework perform on an established road surface benchmark \citep{menegazzo2021road}?
    \item \textbf{RQ2 — Effectiveness of multimodal fusion:} When evaluated on ROAD Subset~\#1, which provides long, continuous, and challenging real-world variability, does combining camera and IMU information improve robustness compared to prior approaches evaluated on more constrained datasets?
    \item \textbf{RQ3 — Modality contribution:} To what extent does the IMU stream affect overall performance, and how does the model behave when operating in a vision-only mode?
    \item \textbf{RQ4 — Robustness under adverse visual conditions:} How well does the vision-only configuration generalize to ROAD Subset~\#2 and synthetically augmented training using Subset~\#3, which explicitly includes adverse illumination, weather variability, and heterogeneous capture conditions?
\end{itemize}

Each experiment below addresses one or more of these questions.

\subsection{Datasets and evaluation protocol}

All ROAD subsets use a 70/20/10 split for training, validation, and testing, respectively, ensuring that frames belonging to the same continuous driving segment remain within the same split to prevent temporal leakage.

\paragraph{Experiment 1 — Benchmark on PVS (RQ1).}
We first evaluate our method on the Passive Vehicular Sensors (PVS) dataset~\citep{menegazzo2021road}, a widely used benchmark for RSC using IMU and RGB cameras. We follow the dataset's original evaluation protocol and adapt the input formats to match our architecture. This experiment validates whether the proposed model generalizes beyond our newly collected dataset.

\paragraph{Experiment 2 — Multimodal evaluation on ROAD Subset~\#1 (RQ2).}
ROAD Subset~\#1 contains synchronized RGB images and five IMU streams captured from a test truck operating across varied environments (nighttime, rain, mixed night+rain).  
For this experiment, we use both modalities (camera + IMU) and train the model end-to-end using a 70/20/10 split, without external data (Subsets \#2 or \#3). This setup evaluates the benefits of adaptive fusion under real-world variability.

\paragraph{Experiment 3 — Modality ablation (RQ3).}
To isolate the contribution of the IMU stream, we train a vision-only baseline under identical settings used in Experiment~2. The goal of this experiment is to quantify the standalone strength of visual cues, the marginal contribution of IMU signals, and the system's ability to operate under sensor degradation.

\paragraph{Experiment 4 — Vision-only robustness on Subsets \#2 and \#3 (RQ4).}
To evaluate generalization under severe appearance variability, we train a vision-only model using all vision-capable subsets (\#1 - \#3), ignoring IMU channels from \#1, and test on Subsets \#1 and \#2. This experiment isolates visual robustness in conditions such as heavy rain, low illumination, and heterogeneous camera devices.

\subsection{Baseline methods}

We compare the proposed approach with two state-of-the-art baselines commonly used in road surface classification. \citet{menegazzo2021road}, an IMU-centric deep learning approach, and \citet{van2025hybrid}, which uses hybrid feature selection to improve prediction pipeline. Baselines were implemented as closely as possible to the original descriptions. All methods are evaluated on the same ROAD Subset~\#1 test set for fair comparison.

\subsection{Evaluation metrics}

We report performance using three standard metrics for imbalanced multi-class classification. First, we use Accuracy, considering:

\begin{equation}
    \text{Accuracy} = 
    \frac{\sum_{i=1}^{K} \text{TP}_i}
    {\sum_{i=1}^{K} (\text{TP}_i + \text{FP}_i + \text{FN}_i)},
\end{equation}
where $K$ is the number of classes. Given how ROAD is imbalanced, with a strong dominance of \textit{Asphalt}, we also evaluate the models using macro-averaged F1-score, which provides a class-balanced measure:

\begin{align}
    \text{Precision}_i &= 
    \frac{\text{TP}_i}{\text{TP}_i + \text{FP}_i}, \\
    \text{Recall}_i &= 
    \frac{\text{TP}_i}{\text{TP}_i + \text{FN}_i}, \\
    \text{F1}_i &= 
    \frac{2 \cdot \text{Precision}_i \cdot \text{Recall}_i}
         {\text{Precision}_i + \text{Recall}_i}, \\
    \text{Macro-F1} &= \frac{1}{K} \sum_{i=1}^{K} \text{F1}_i.
\end{align}

Lastly, we also include normalized confusion matrices for qualitative error analysis, defined as:
\begin{equation}
    C'_{ij} = 
    \frac{C_{ij}}
         {\sum_{k=1}^{K} C_{ik}},
\end{equation}
where $C_{ij}$ is the number of samples of class $i$ predicted as $j$. These matrices highlight systematic biases such as majority-class overprediction and cross-class ambiguity.

\subsection{Implementation details}

All models were implemented using PyTorch 2.7.1 with CUDA 12.8 and trained on a workstation equipped with an NVIDIA GeForce RTX~5070 GPU.  We used AdamW \citep{loshchilov2017decoupled} as the optimizer, with a learning rate of $10^{-3}$ and weight decay of~$2\times10^{-4}$.  Models were trained for up to~50~epochs with a batch size of~32, 
using early stopping based on validation accuracy to prevent overfitting. All preprocessing and data augmentation steps described in Section~\ref{sec:preprocessing} were applied online during training.

\section{Results and Discussion}
This section presents a comprehensive analysis of the experimental results obtained with the proposed framework. We first analyze the performance of the multimodal approach under sensor-fusion settings, comparing quantitative metrics and qualitative behavior, directly addressing RQ1 and RQ2 (\autoref{subsec:sensor-fusion-results}). Within this analysis, we also perform an ablation study to investigate the contribution of the IMU stream and sensor-degraded configurations, directly addressing RQ3. Finally, we evaluate the robustness of the vision-only configuration under increasingly adverse and heterogeneous visual conditions, supported by both quantitative and qualitative analyses (\autoref{subsec:vision-only-results}).

\subsection{Sensor fusion results}
\label{subsec:sensor-fusion-results}
\paragraph{Quantitative analysis.} \autoref{tab:results} summarizes the quantitative analysis of our proposed framework against two other state-of-the-art approaches for road surface classification. As described in \autoref{sec:experiments}, we evaluate all models using overall accuracy and macro-averaged F1-score as performance metrics. Evaluations are conducted on the PVS dataset \citep{menegazzo2021road}, a well-established benchmark, as well as on ROAD Subset~\#1 (\autoref{road:subset-1}), which is specifically designed for multimodal sensor fusion analysis. It is important to highlight that this subset is imbalanced: \textit{Asphalt} accounts for 79.7\% of all samples, \textit{Off-road} for 11.78\%, and \textit{Belgian Blocks} for 8.55\%. This distribution mirrors real-world driving conditions but poses a challenge for learning minority classes, particularly for models that tend to overfit to dominant visual patterns. As we show below, this imbalance strongly influences the behavior of the baseline methods. This comparison directly evaluates the model’s generalization ability on established benchmarks (\textbf{RQ1}) and its robustness under multimodal conditions on ROAD Subset \#1 (\textbf{RQ2}).

Our method achieves an overall accuracy of 95.6\% on PVS, outperforming the previous state-of-the-art by 1.4 percentage points compared to \citet{van2025hybrid} and 2.9 percentage points compared to \citet{menegazzo2021road}. Consistently high F1-scores across all classes indicate that the model not only maintains precision on \textit{Asphalt}, the dominant class, but also has an enhanced discrimination on the more challenging classes of \textit{Cobblestone} and \textit{Dirt road}. These results confirm the model’s ability to generalize beyond our dataset and outperform prior methods on an external benchmark (\textbf{RQ1}).

On the proposed ROAD dataset, which contains a broader range of environmental and surface variability, the gap between existing methods and ours becomes more pronounced. While previous approaches exhibit severe drops in minority-class performance (e.g., class-wise F1 below 50\% on \textit{Off-road} and close to 0\% on \textit{Belgian Blocks}), our framework remains more balanced, reaching 98.2\% overall accuracy and a class-wise F1 of 100\% for \textit{Off-road} on the Subset~\#1 test split. Although this value may appear high, it is consistent with the evaluation protocol adopted in this work, where data are split at the level of continuous driving segments to avoid temporal leakage. Moreover, the same trend holds across baselines evaluated under identical splits, suggesting that the improvement reflects robustness to realistic variability rather than artifacts of the evaluation protocol. We interpret this behavior as being consistent with the design of the multimodal fusion mechanism, which is intended to promote temporal stability by incorporating inertial cues when visual information becomes unreliable. In contrast, prior methods show inflated F1-scores for the dominant \textit{Asphalt} class, consistent with majority-class overprediction under imbalance. Overall, these results support the benefit of combining visual and inertial cues under challenging and diverse real-world conditions (\textbf{RQ2}).

\begin{figure*}[b!]
\centering

\begin{subfigure}{0.31\textwidth}
    \centering
    \includegraphics[width=\linewidth]{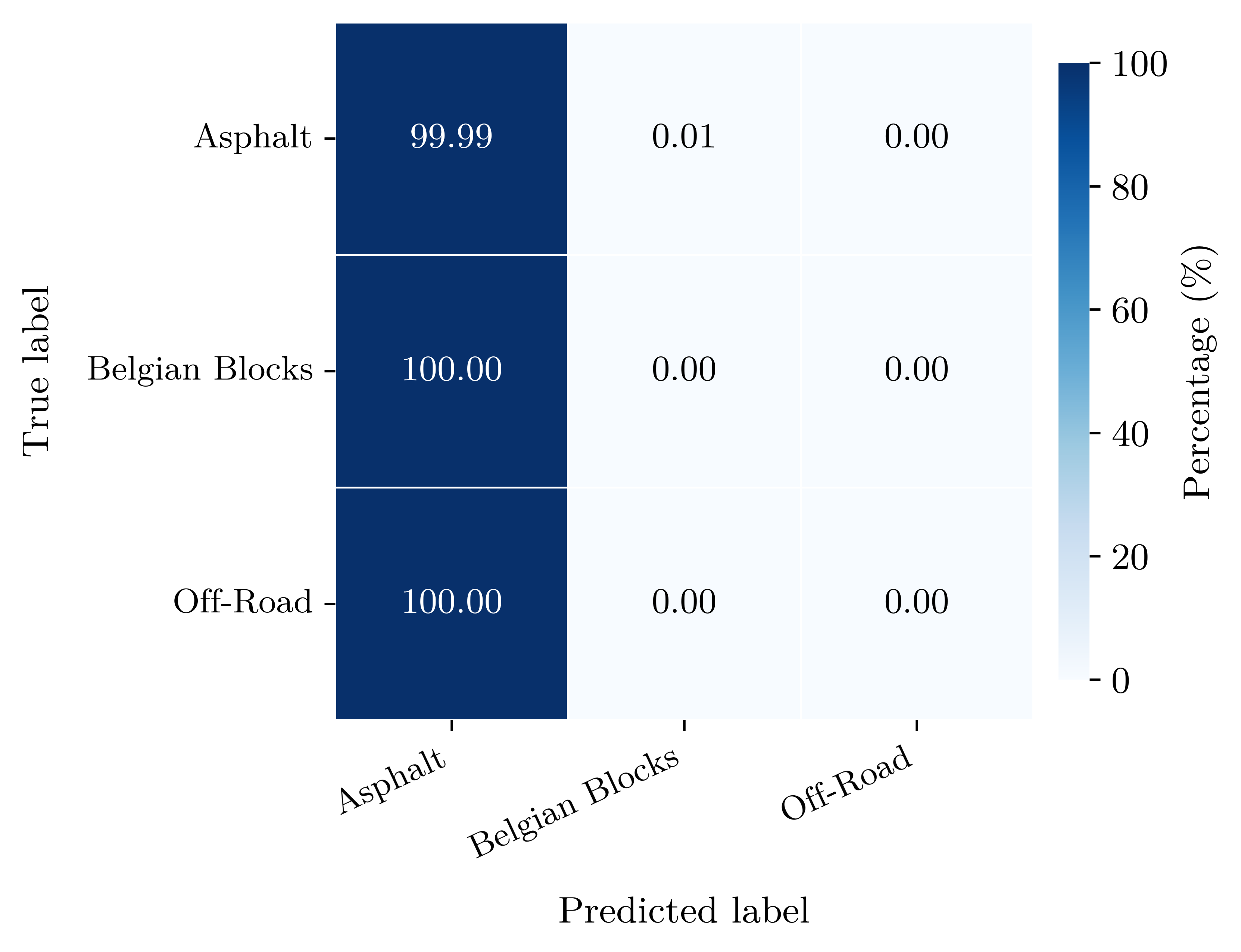}
    \caption{\footnotesize \citet{menegazzo2021road}}
    \label{fig:confusion-matrices:a}
\end{subfigure}
\hfill
\begin{subfigure}{0.31\textwidth}
    \centering
    \includegraphics[width=\linewidth]{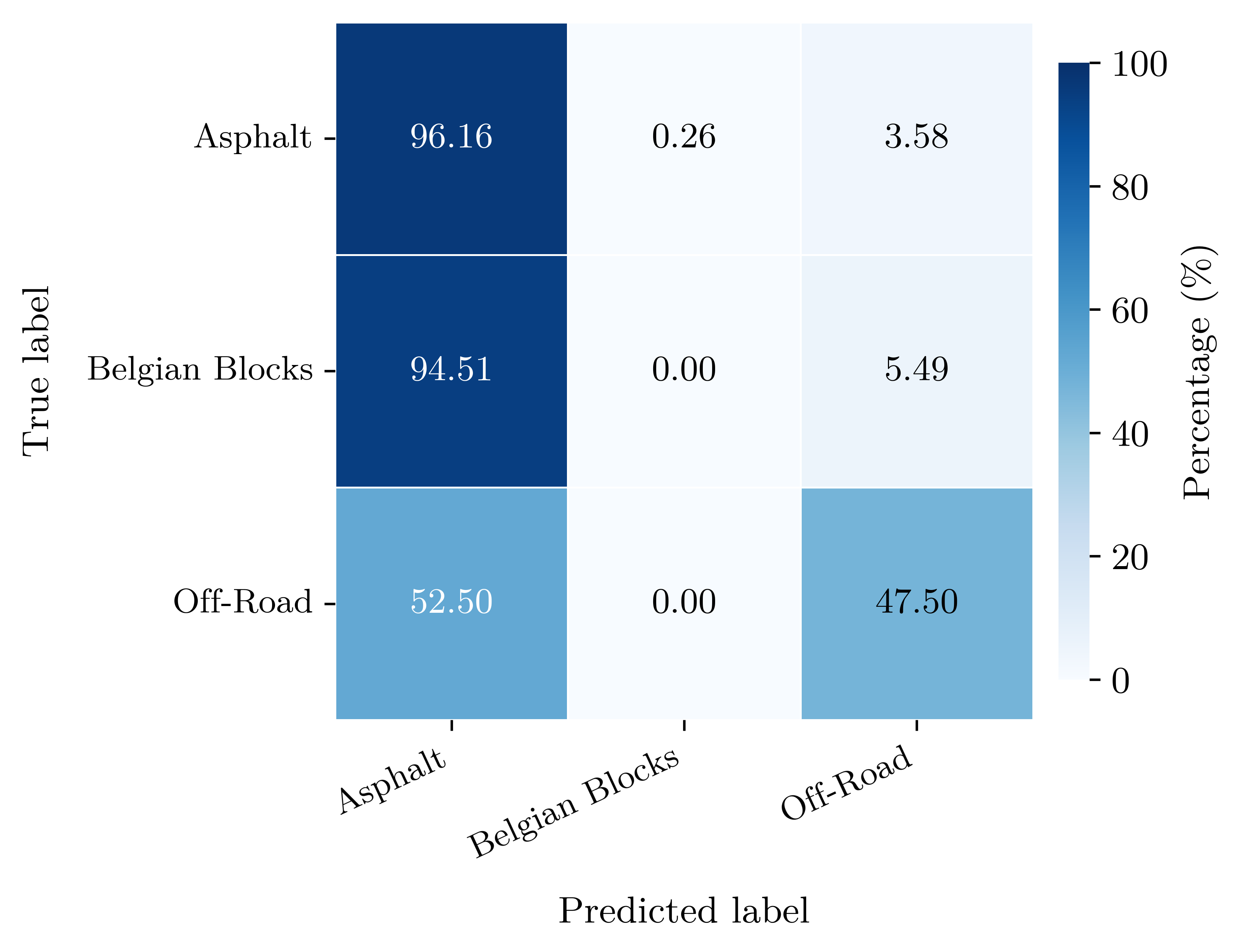}
    \caption{\citet{van2025hybrid}}
    \label{fig:confusion-matrices:b}
\end{subfigure}
\hfill
\begin{subfigure}{0.31\textwidth}
    \centering
    \includegraphics[width=\linewidth]{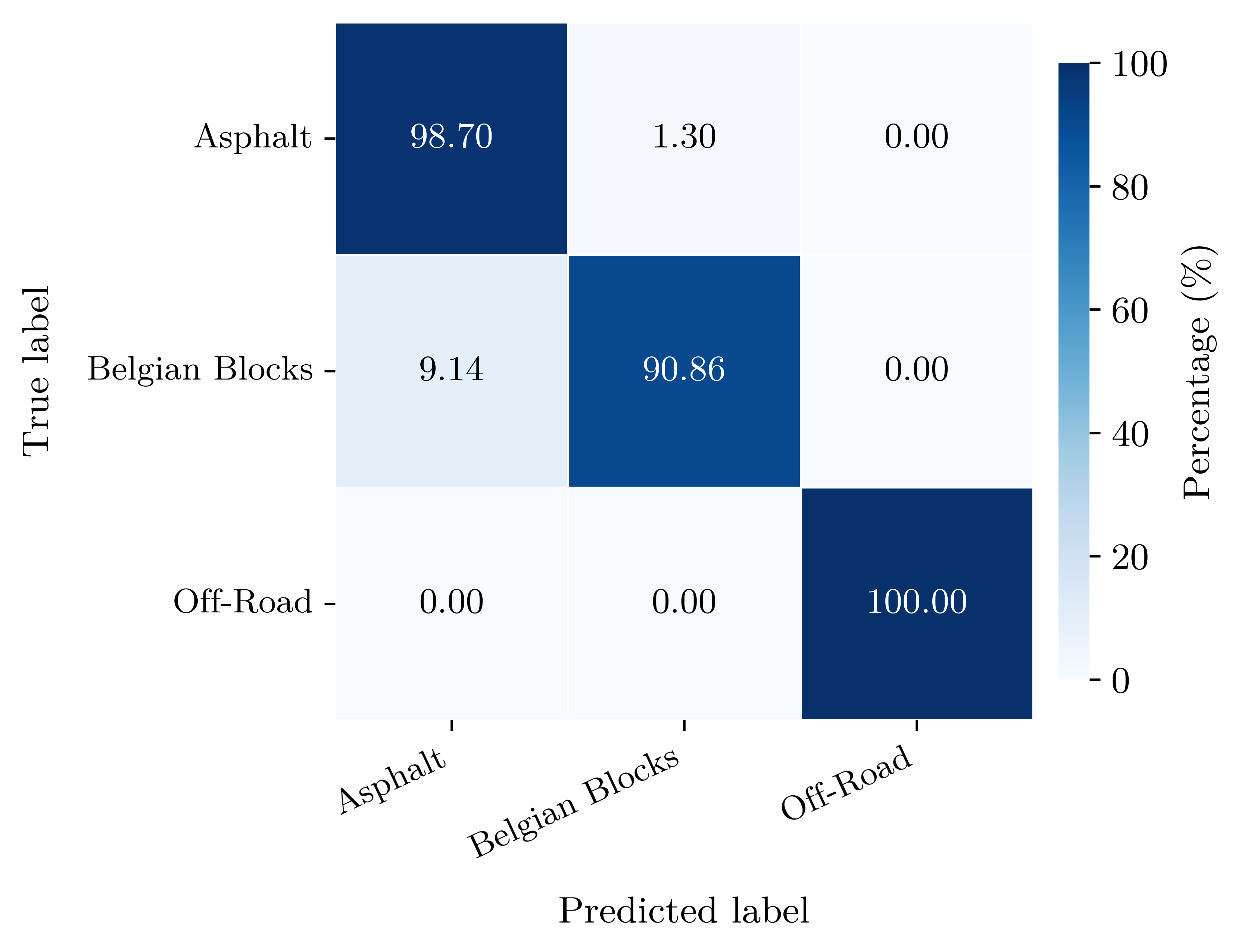}
    \caption{Ours}
    \label{fig:confusion-matrices:c}
\end{subfigure}

\caption{Normalized confusion matrices comparing two state-of-the-art baselines with our proposed framework, all evaluated on the same test split of ROAD Subset \#1. The visual comparison highlights systematic overprediction of the majority class (\textit{Asphalt}) by prior methods and the more balanced class discrimination achieved by our approach.}
	\label{fig:confusion-matrices}
\end{figure*}

\paragraph{Confusion matrices.} The confusion matrices in \autoref{fig:confusion-matrices} provide additional insights into the quantitative results of each model on the ROAD dataset. As we show in \autoref{fig:confusion-matrices:a} and \autoref{fig:confusion-matrices:b}, current baselines tend to overpredict the \textit{Asphalt} class, misclassifying nearly all other classes. This pattern is consistent with our discussion of the class-wise F1-scores, suggesting a tendency toward overprediction of the dominant class under challenging conditions, which directly relates to the robustness issues examined in \textbf{RQ2}.

In contrast, our proposed approach (\autoref{fig:confusion-matrices:c}) achieves a more balanced class distribution. While close to 10\% of \textit{Belgian Blocks} samples are misclassified as \textit{Asphalt}, our approach successfully identifies all \textit{Off-road} samples and significantly improves recognition of minority classes. This demonstrates that our approach enhances feature discrimination for non-asphalt surfaces, even in uncontrolled settings, effectively reducing the model's bias toward the majority class and improving generalization across heterogeneous road conditions.

\begin{figure*}[b!]
\centering

\begin{subfigure}{\textwidth}
    \centering
    \begin{subfigure}{0.31\textwidth}
        \centering
        \includegraphics[width=\linewidth]{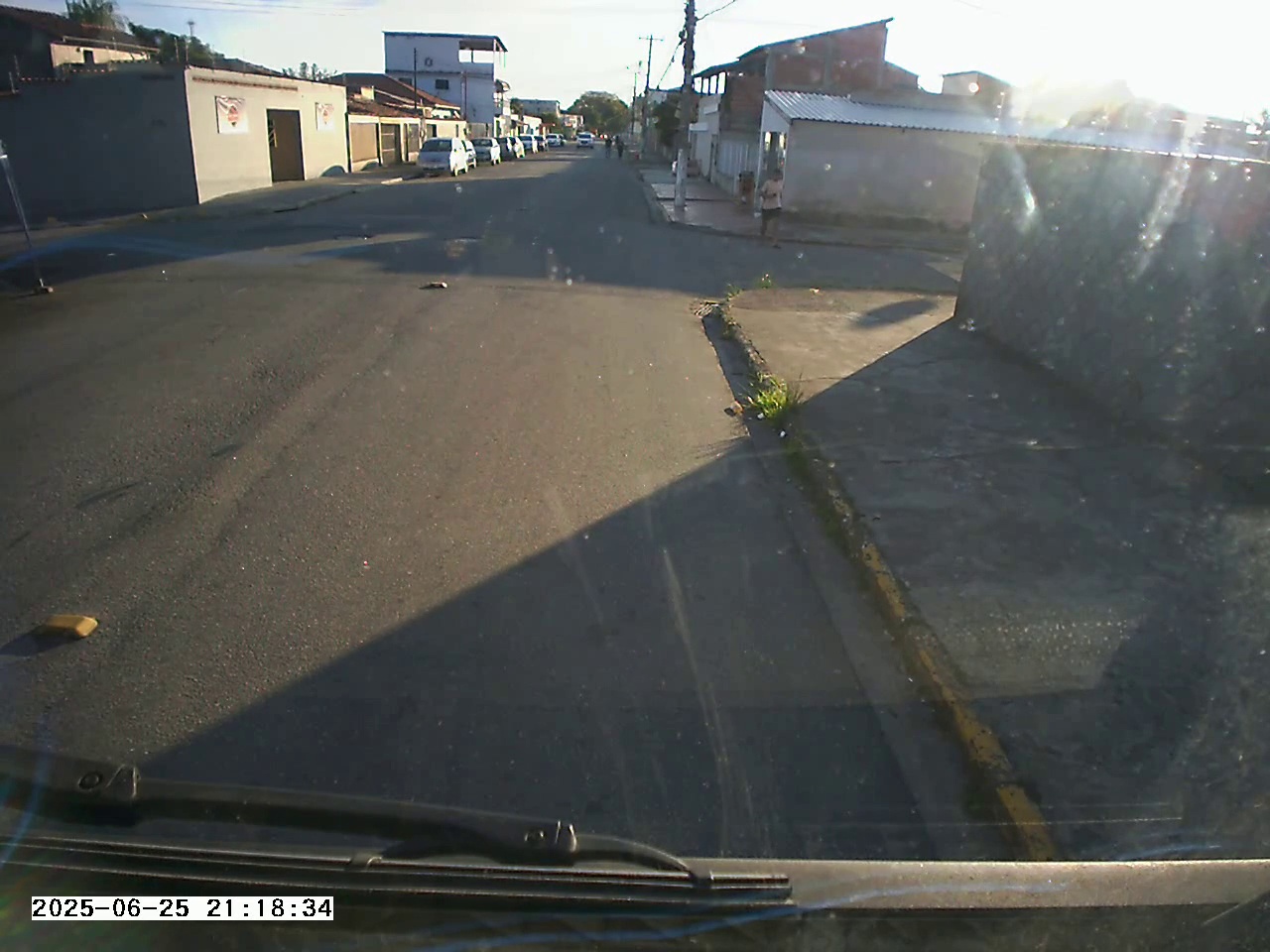}
    \end{subfigure}
    \hfill
    \begin{subfigure}{0.31\textwidth}
        \centering
        \includegraphics[width=\linewidth]{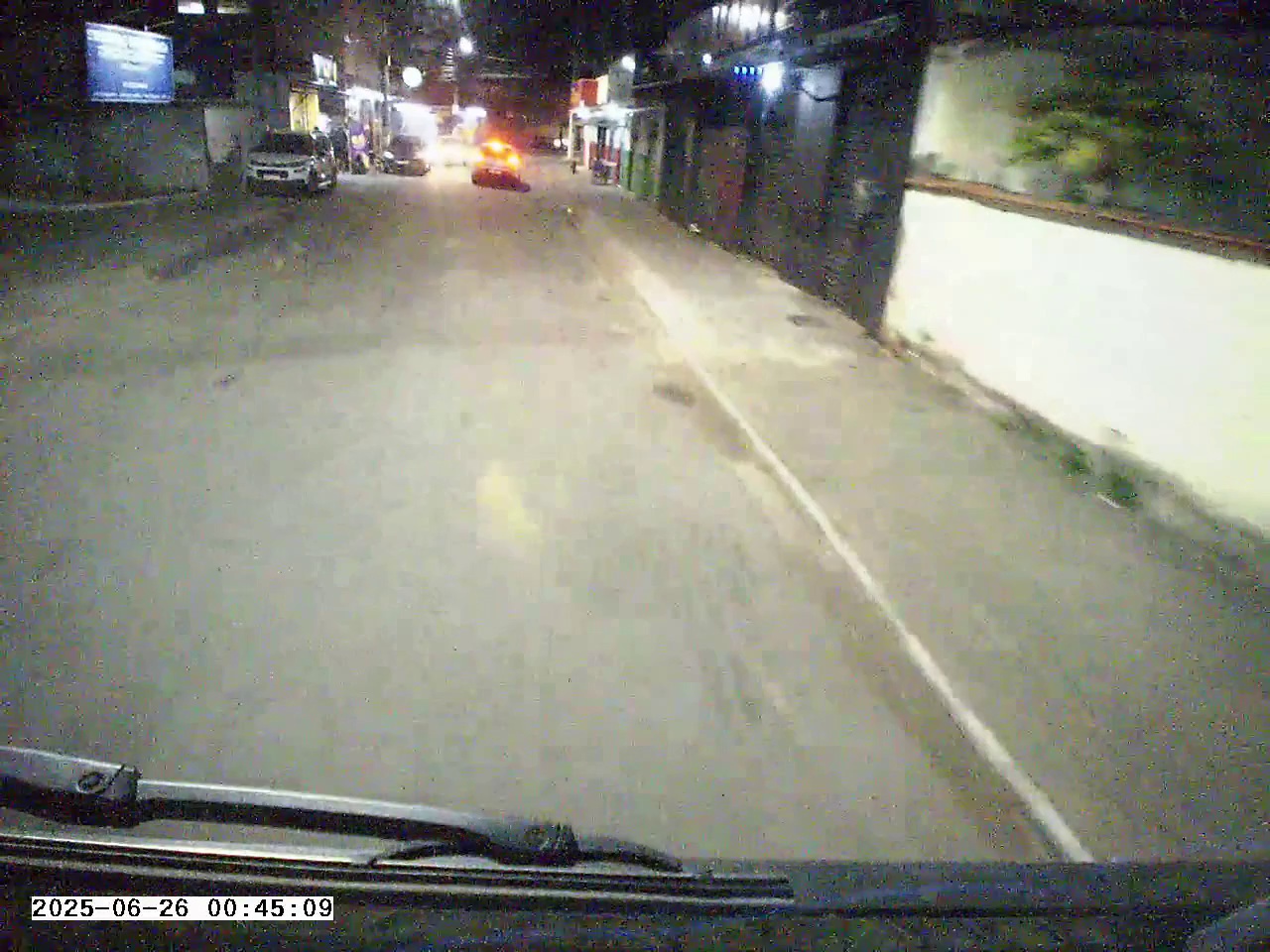}
    \end{subfigure}
    \hfill
    \begin{subfigure}{0.31\textwidth}
        \centering
        \includegraphics[width=\linewidth]{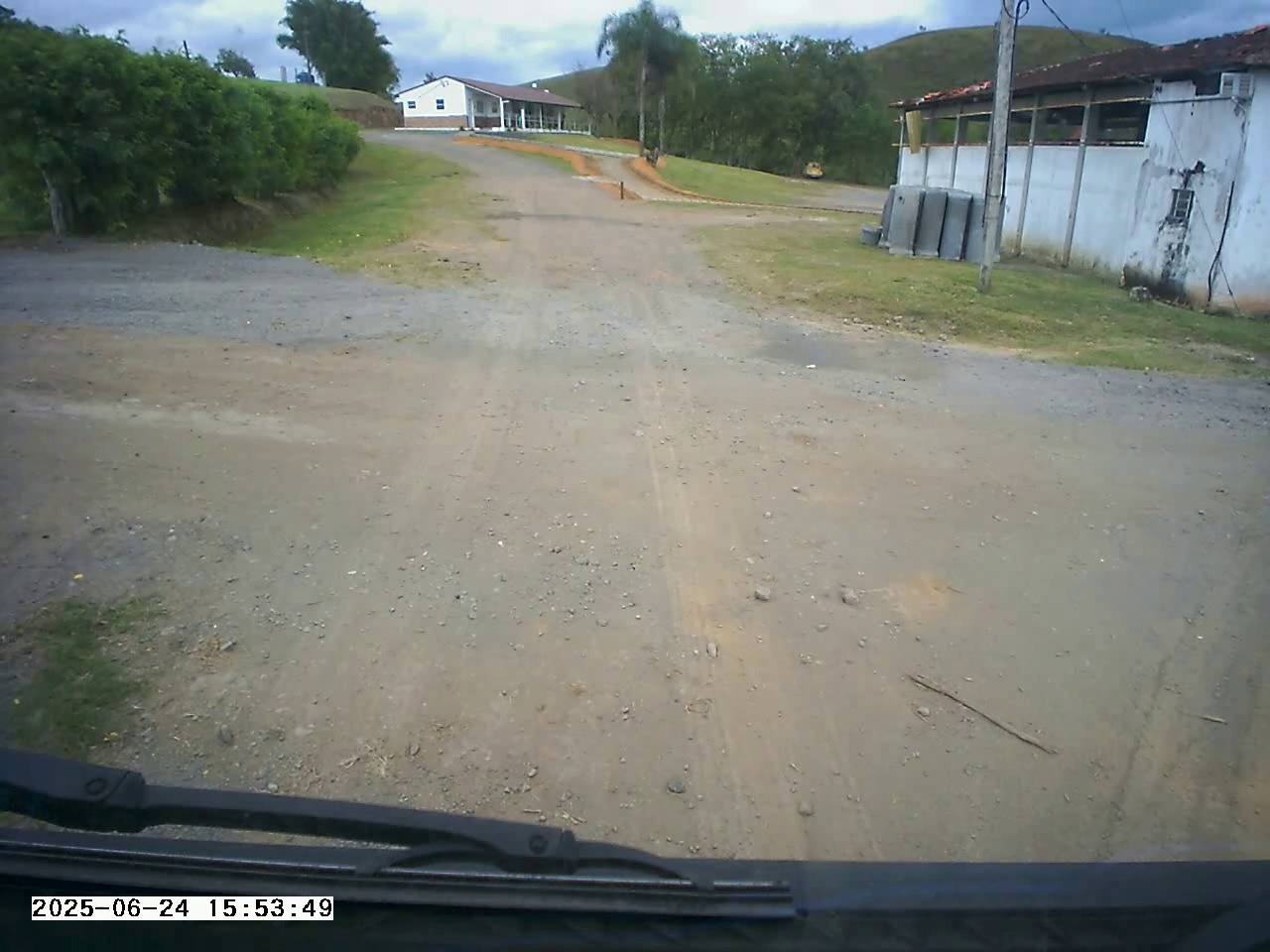}
    \end{subfigure}
    \caption{Correct classification on the ROAD dataset, Subset \#1.}
    \label{fig:success}
\end{subfigure}

\vspace{1em}

\begin{subfigure}{\textwidth}
    \centering
    \begin{subfigure}{0.31\textwidth}
        \centering
        \includegraphics[width=\linewidth]{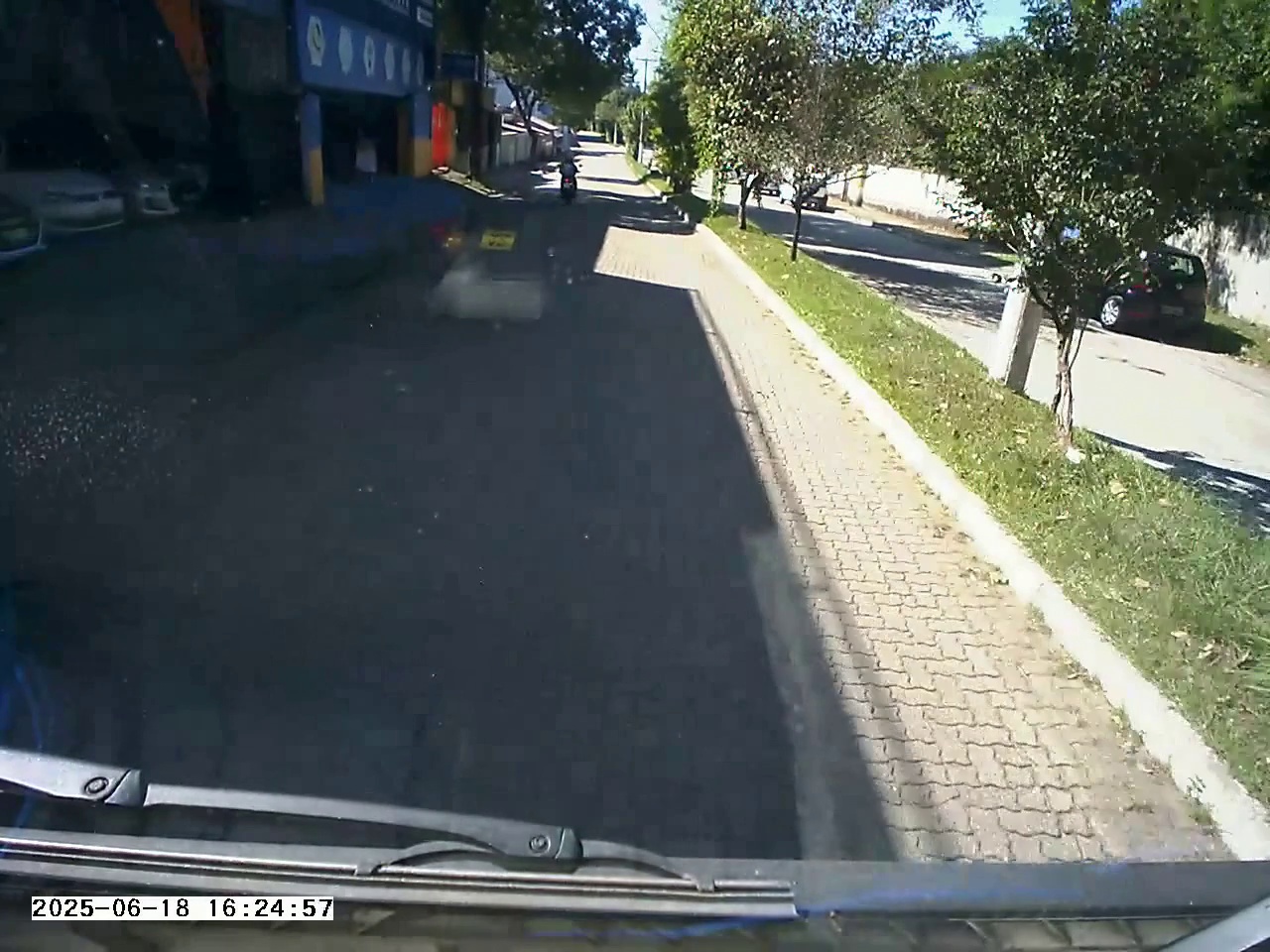}
    \end{subfigure}
    \hfill
    \begin{subfigure}{0.31\textwidth}
        \centering
        \includegraphics[width=\linewidth]{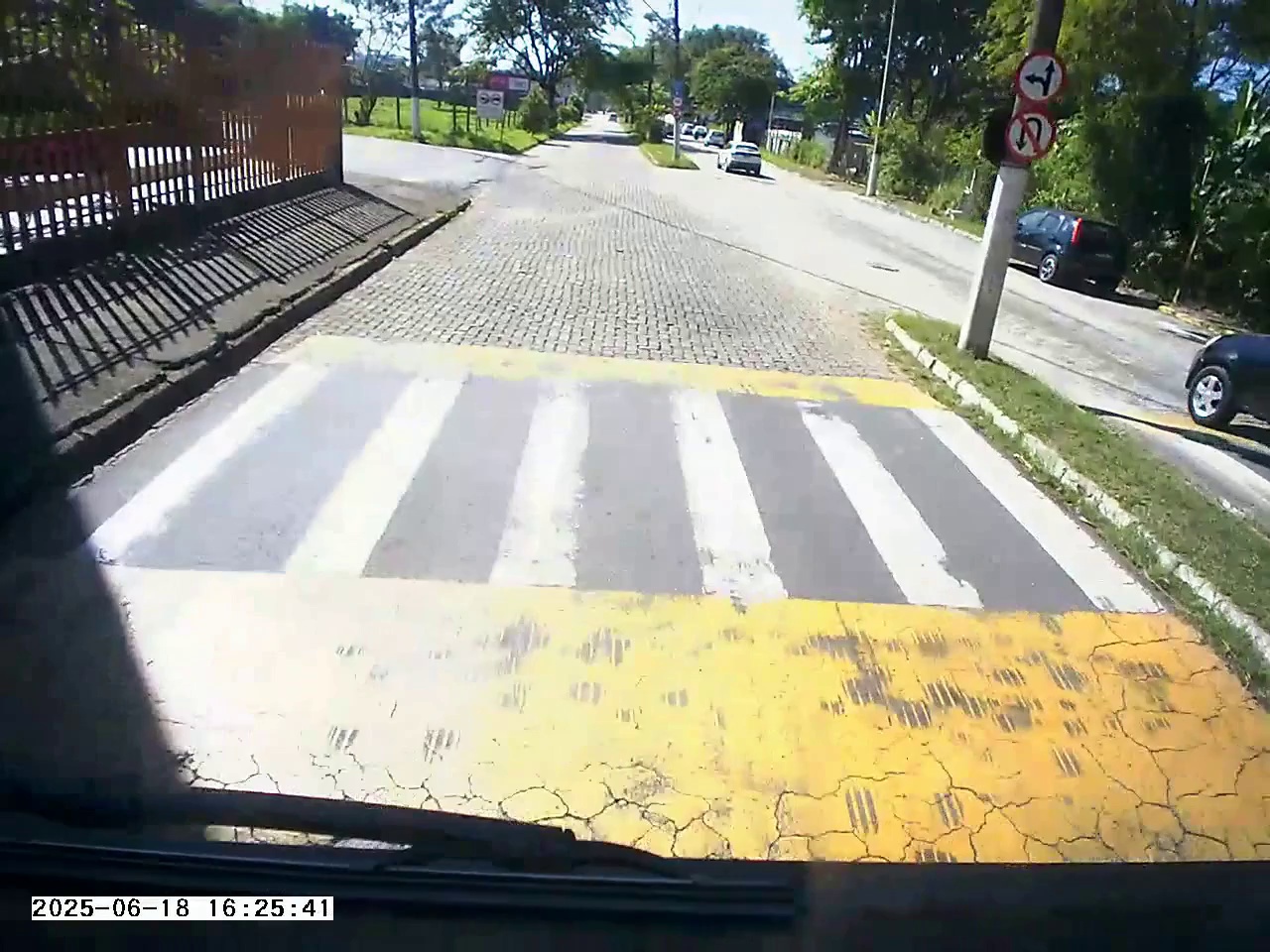}
    \end{subfigure}
    \hfill
    \begin{subfigure}{0.31\textwidth}
        \centering
        \includegraphics[width=\linewidth]{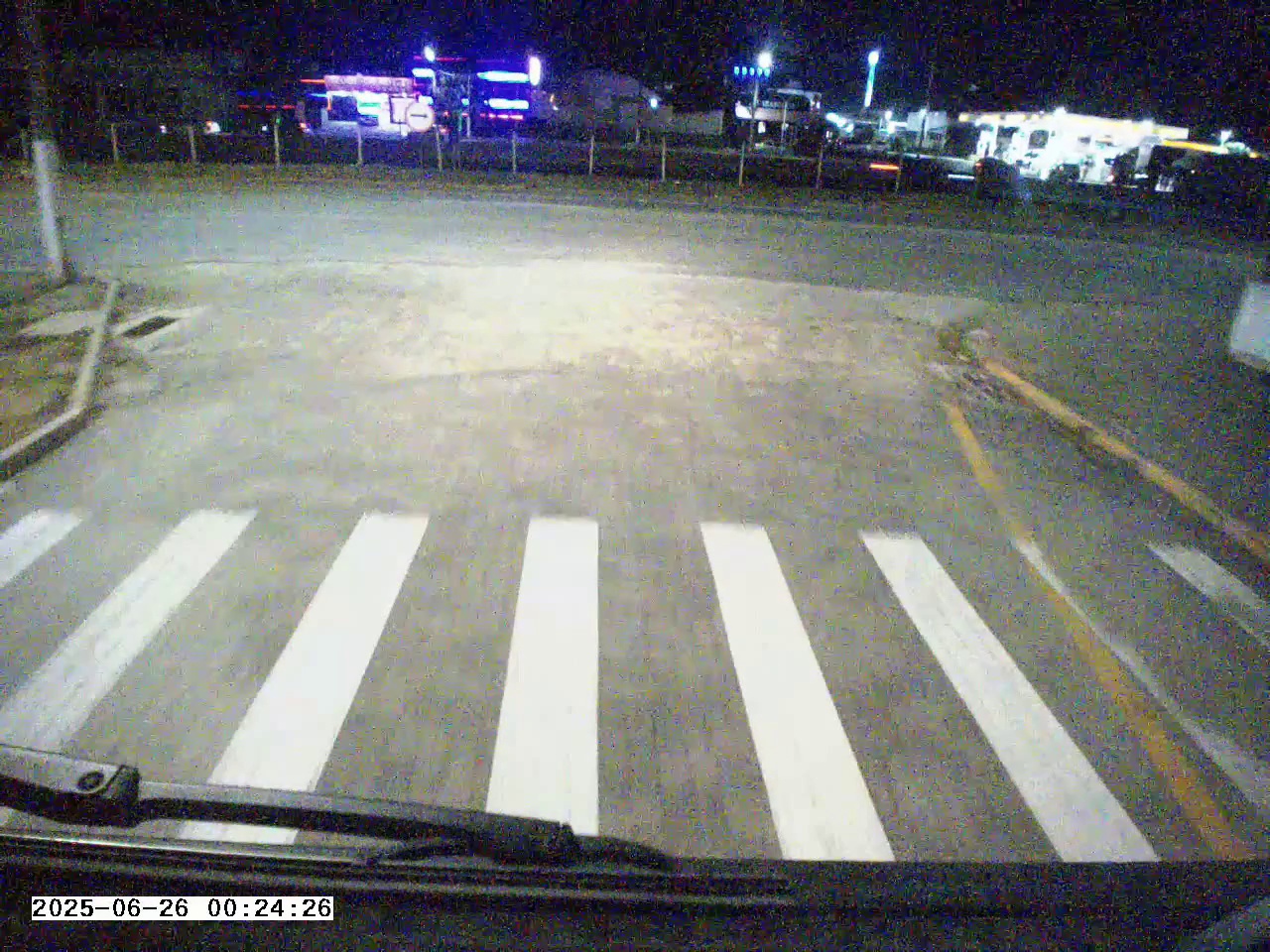}
    \end{subfigure}
    \caption{Misclassifications on the ROAD dataset, Subset \#1. From left to right, the model predictions and their ground-truth (GT) are Asphalt (\textbf{GT}: Belgian Blocks), Belgian Blocks (\textbf{GT}: Asphalt), and Asphalt (\textbf{GT}: Belgian Blocks).}
    \label{fig:failure}
\end{subfigure}

\caption{Representative examples of correct classifications (a) and misclassifications (b) produced by our model on Subset \#1 of the ROAD dataset. These samples illustrate the model’s robustness under challenging conditions, such as nighttime scenes, shadows, and heterogeneous surface textures, as well as typical failure cases occurring in transitional regions between \textit{Asphalt} and \textit{Belgian blocks}.}
\label{fig:qualitative-analysis-subset-1}
\end{figure*}

\paragraph{Qualitative analysis.}
\label{para:fusion_quali}

We further conducted qualitative analyses to illustrate the behavior of the proposed model under diverse real-world conditions. Representative examples of correct predictions are shown in \autoref{fig:success}. From left to right, these include a correct classification of \textit{Asphalt} despite the presence of strong lens flare and pronounced shadows producing a duotone appearance; a nighttime \textit{Asphalt} scene affected by poor sensor quality, resulting in severe motion blur and Gaussian noise; and a correct classification of \textit{Off-road} in a visually challenging scenario where the grayish appearance of crushed stone closely resembles \textit{Asphalt}, a common characteristic of rural roads in Brazil.

Conversely, \autoref{fig:failure} presents representative failure cases. From left to right, these include shadowed regions in which darker areas visually resemble \textit{Asphalt}; frames with heterogeneous textures containing both \textit{Asphalt} and \textit{Belgian Blocks} simultaneously; and a severely degraded \textit{Belgian Blocks} sample exhibiting a combination of adverse lighting, blur, and noise within the same frame. Most misclassifications occur during transitions between \textit{Asphalt} and \textit{Belgian Blocks}, where consecutive frames contain overlapping textures of both surfaces. We attribute this behavior primarily to the temporal offset between the camera and IMU streams: while the IMU captures the vehicle’s instantaneous motion, the camera’s field of view often corresponds to the upcoming region of the road.

In other words, because the camera is mounted with a forward-facing angle and the IMU reflects instantaneous vehicle dynamics, the temporal correspondence between modalities is imperfect during surface transitions. This observation highlights the importance of precise multimodal synchronization. This suggests that future work could benefit from investigating finer temporal alignment.

We also noticed that some failure cases tend to occur in short temporal bursts, with multiple consecutive samples misclassified. This behavior suggests that the fusion model can enter brief periods of instability, likely triggered by imperfect camera-IMU alignment or IMU misclassifications, which propagate for a moment before the model recovers. We show examples of this issue in \autoref{appendix_a}.

\paragraph{Ablation study.}
We examine the contribution of the inertial stream by comparing the full multimodal model with a vision-only variant that preserves the same training protocol and data augmentation strategy. As shown in \autoref{tab:ablation}, both configurations achieve comparable performance, with the vision-only variant reaching slightly higher accuracy on both datasets. Importantly, this result does not imply that road surface classification can be effectively addressed by arbitrary vision-only approaches. Rather, it indicates that, within our framework, strong visual representations and training on diverse conditions already capture much of the discriminative information, while the IMU stream primarily contributes complementary robustness in ambiguous or degraded scenarios. This analysis directly addresses the role of modality contribution examined in \textbf{RQ3}.

These marginal differences suggest that, even in challenging scenarios present in our dataset, the IMU stream contributes minimally to overall accuracy and therefore points to future research focusing on visual modalities. The comparable performance also highlights that the fusion architecture remains effective even when operating in a vision-only mode, confirming that the framework does not over-rely on the IMU stream.
 
However, this outcome is consistent with the intended design of the system: the inertial modality is not expected to boost performance under ideal conditions but to provide complementary motion cues that enhance temporal consistency and cross-condition performance when visual information becomes unreliable. By integrating sensor data that is invariant to illumination and appearance, the fusion architecture offers greater resilience in out-of-distribution scenarios, such as unseen weather conditions, camera vibration, or partial sensor failure, where purely vision-based models typically degrade. These observations confirm the role of the IMU stream as a robustness enhancer rather than a primary accuracy driver (\textbf{RQ3}). Given this discovery, we also proceed to evaluate our vision model on the vision-capable subsets.

\begin{table}[h!]
\centering
\caption{Accuracy comparison between our full multimodal model (camera + IMU) and our vision-only baseline on PVS and ROAD Subset \#1. Results show minimal contribution from the IMU stream.}
\label{tab:ablation}
\begin{tabular}{lccc}
\toprule
\textbf{Approach} & \textbf{PVS (\%)} & \textbf{Subset \#1 (\%)} \\ 
\midrule
Full model (Camera + IMU)      & 95.6 & 98.2 \\
Vision-only           & 95.4 & 98.4 \\
\bottomrule
\end{tabular}
\end{table}

\subsection{Vision-only results}
\label{subsec:vision-only-results}
\paragraph{Quantitative analysis.} We show in \autoref{tab:ablation} a summary of our vision-only baseline compared with our proposed sensor fusion approach. For PVS \citep{menegazzo2021road}, the vision-only model performs 0.2\% worse, and for ROAD Subset \#1 (\autoref{road:subset-1}), 0.2\% better. These are marginal differences that indicate how comparable these models are. This contextualizes the behavior of the model when relying solely on visual cues (\textbf{RQ4}).

We also evaluate the vision-only approach, expanding the data used for training and testing to go beyond Subset \#1 and including the subsets tailored for vision-only analysis, Subsets \#2 (\autoref{road:subset-2}) and \#3 (\autoref{road:subset-3}), with \#3 used only for training. As we show in \autoref{tab:results-visiononly}, for all subsets, our vision-only approach achieves 96.6\% accuracy, which is lower than its performance on Subset \#1 but justifiable given the increased number of challenging scenarios in Subset \#2.

\begin{figure}[t!]
	\centering
		\includegraphics[width=\columnwidth]{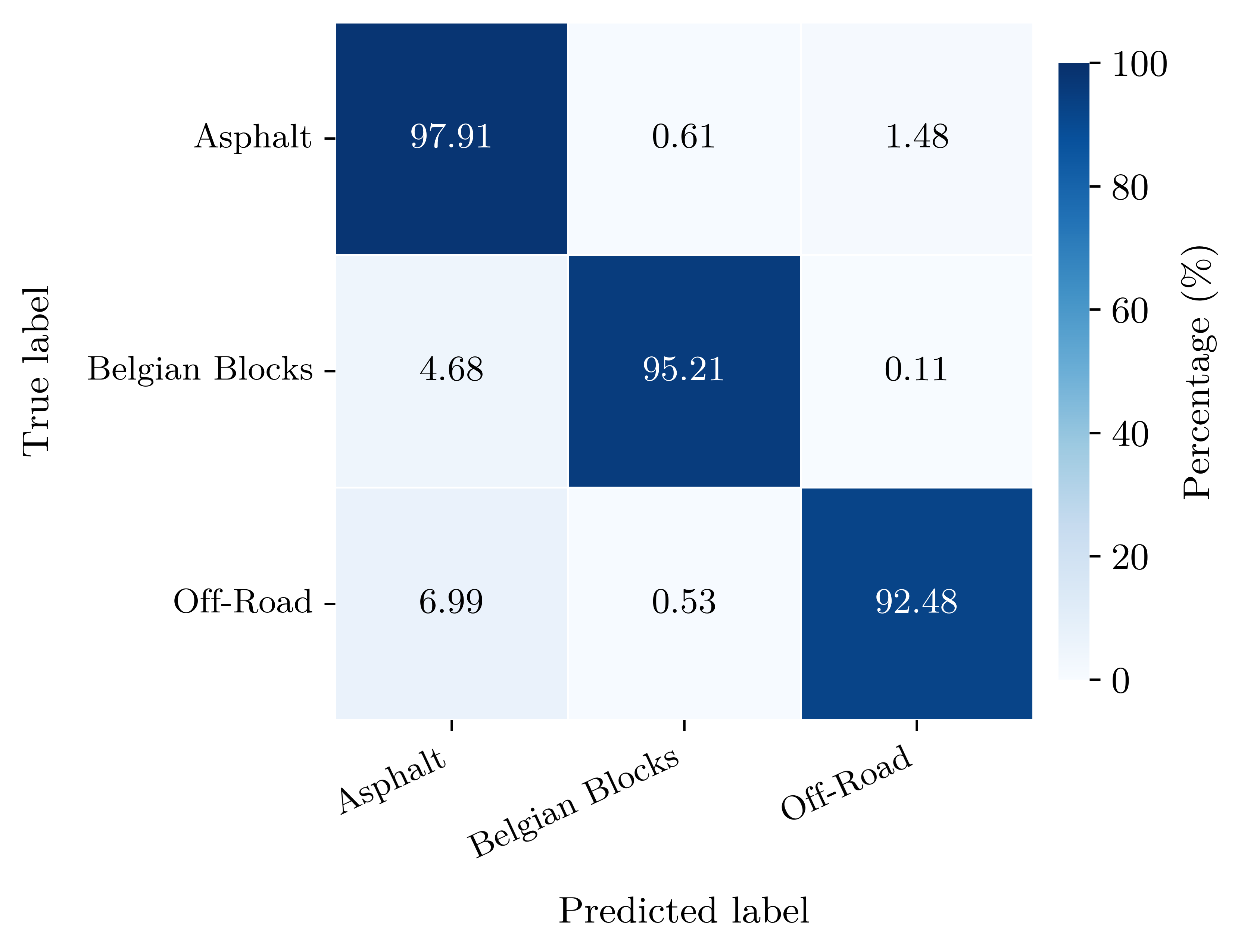}
	\caption{Normalized confusion matrix for the vision-only model evaluated on the ROAD dataset. Although the model achieves high accuracy (>90\%) across all classes, the matrix reveals residual confusions between \textit{Asphalt} and \textit{Belgian Blocks} and occasional errors in \textit{Off-road}, reflecting the sensitivity of a vision-only approach to subtle appearance variations and transitional surface regions.}
	\label{fig:confusion-matrix-vision-only}
\end{figure}

\paragraph{Confusion matrix.} \autoref{fig:confusion-matrix-vision-only} illustrates the normalized confusion matrix for the vision-only baseline on ROAD. Following the quantitative analysis, the model achieves high classification accuracy across all surface types, with correctly identified samples exceeding 92\% in every class. This provides a detailed view of the vision-only model’s behavior across classes (\textbf{RQ4}).

Most errors still occur between \textit{Asphalt} and \textit{Belgian Blocks}, but for this approach, we now see misclassifications in \textit{Off-road} as well, with some confusion with \textit{Asphalt}. These errors highlight the limitations of the visual modality when appearance becomes ambiguous (\textbf{RQ4}) and confirm that, while a vision-only model is capable of strong performance in stable visual scenarios, it remains sensitive to subtle appearance changes and inter-class similarities that arise in transitional or mixed-surface regions.

\begin{figure*}[b!]
\centering

\begin{subfigure}{\textwidth}
    \centering
    \begin{subfigure}{0.32\textwidth}
        \centering
        \includegraphics[width=\linewidth]{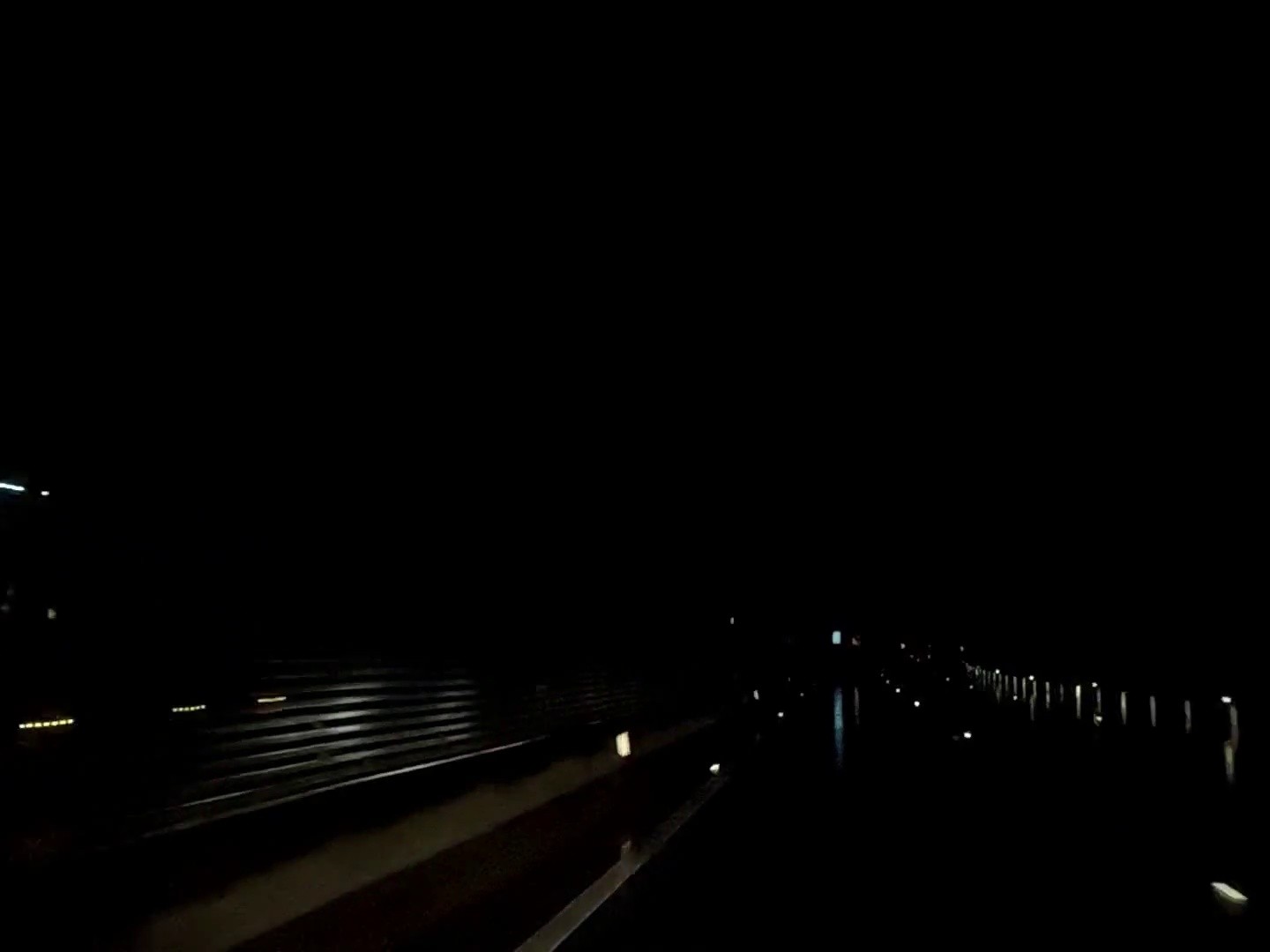}
    \end{subfigure}
    \hfill
    \begin{subfigure}{0.32\textwidth}
        \centering
        \includegraphics[width=\linewidth]{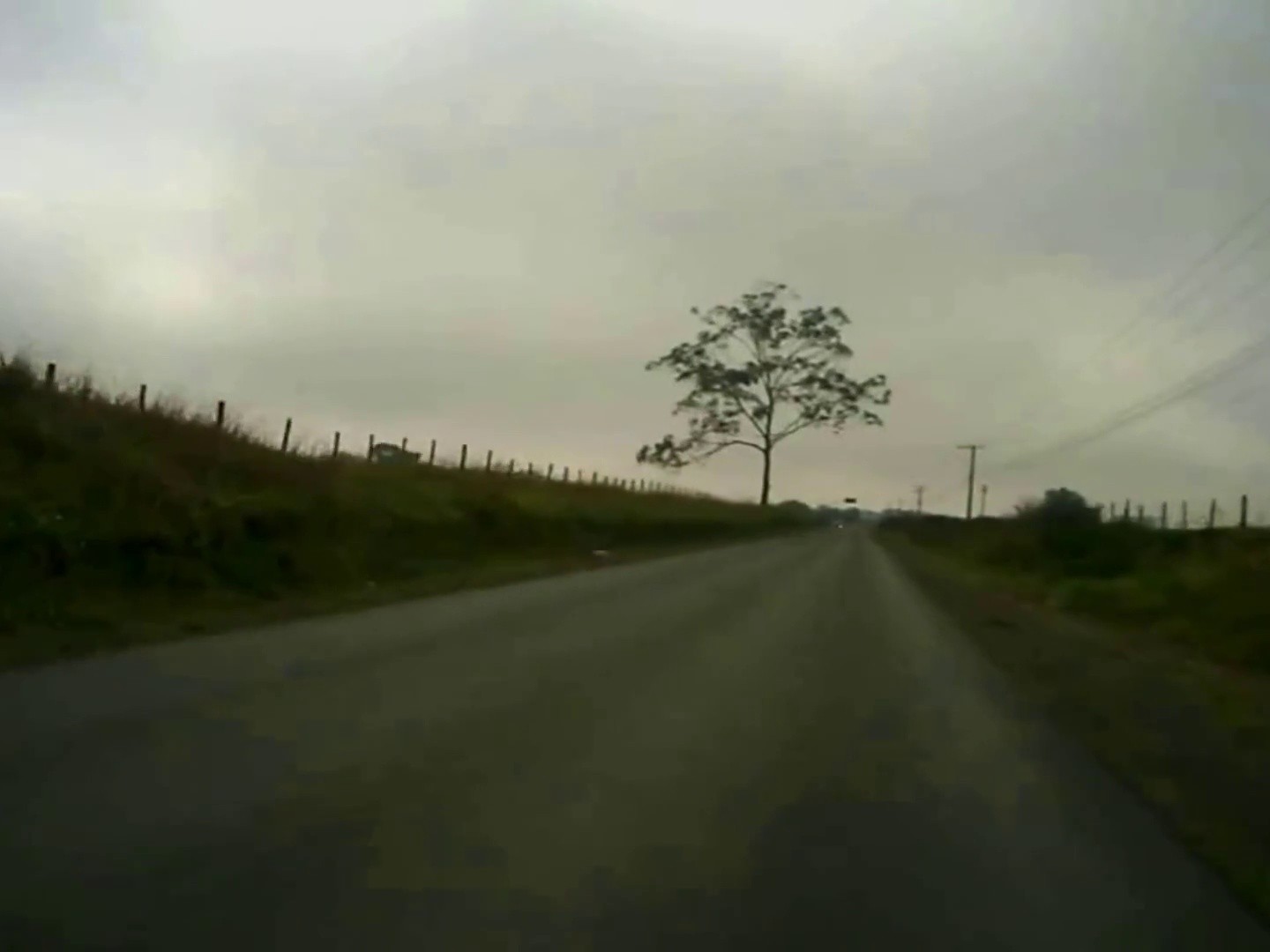}
    \end{subfigure}
    \hfill
    \begin{subfigure}{0.32\textwidth}
        \centering
        \includegraphics[width=\linewidth]{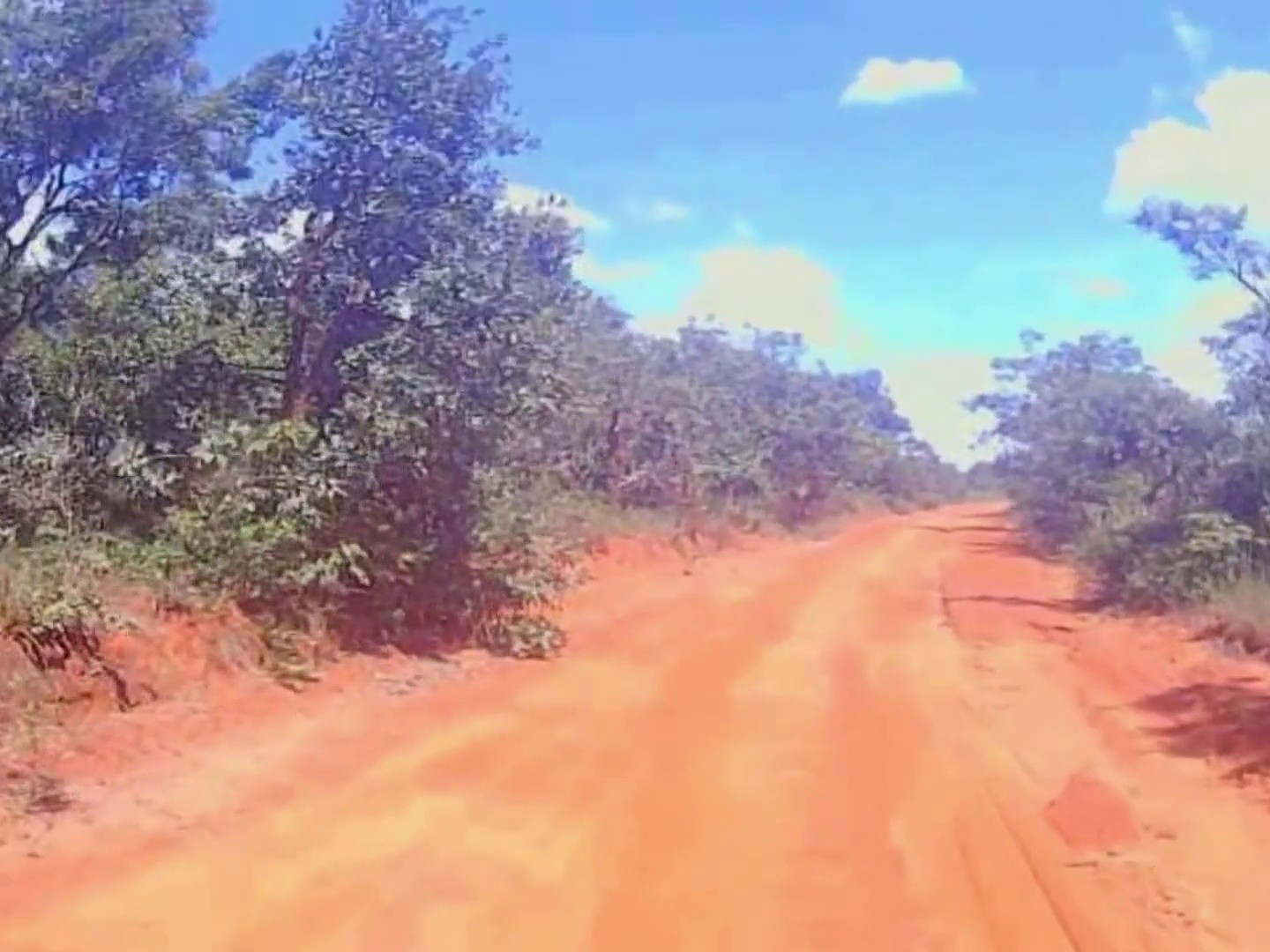}
    \end{subfigure}
    \caption{Correct classifications on the ROAD dataset, Subset \#2.}
    \label{fig:success-vision-all}
\end{subfigure}

\vspace{1em}

\begin{subfigure}{\textwidth}
    \centering
    \begin{subfigure}{0.32\textwidth}
        \centering
        \includegraphics[width=\linewidth]{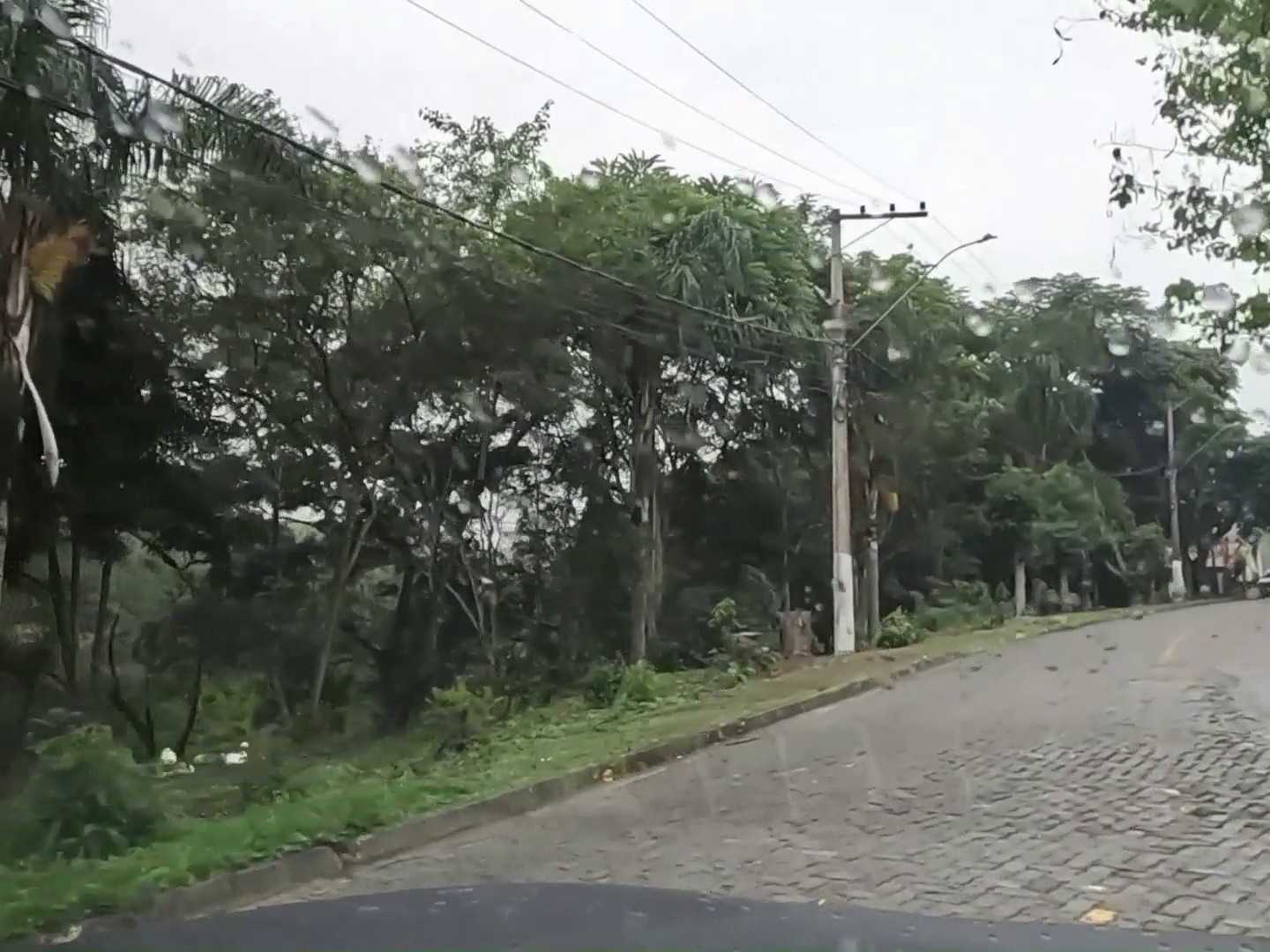}
    \end{subfigure}
    \hfill
    \begin{subfigure}{0.32\textwidth}
        \centering
        \includegraphics[width=\linewidth]{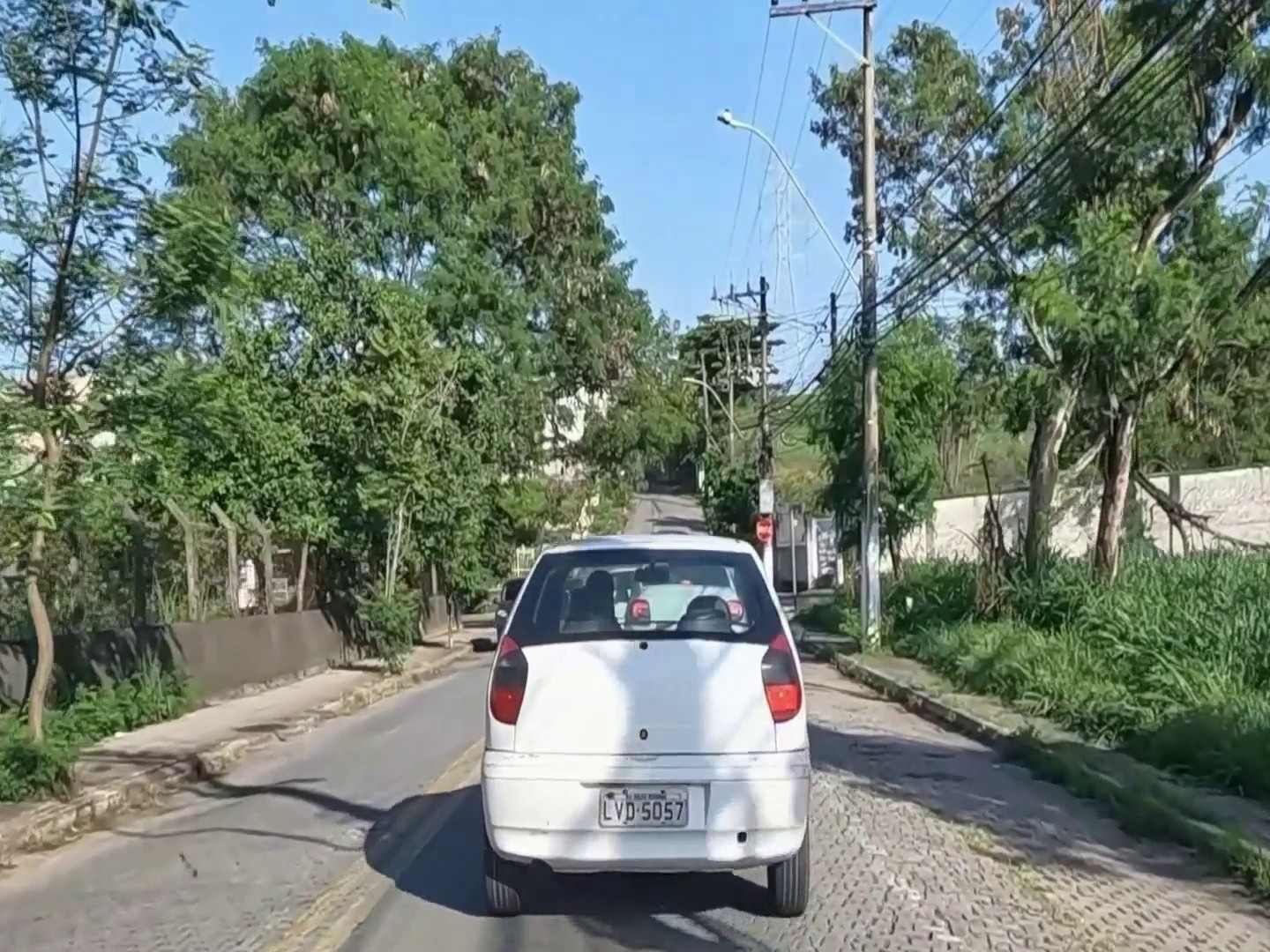}
    \end{subfigure}
    \hfill
    \begin{subfigure}{0.32\textwidth}
        \centering
        \includegraphics[width=\linewidth]{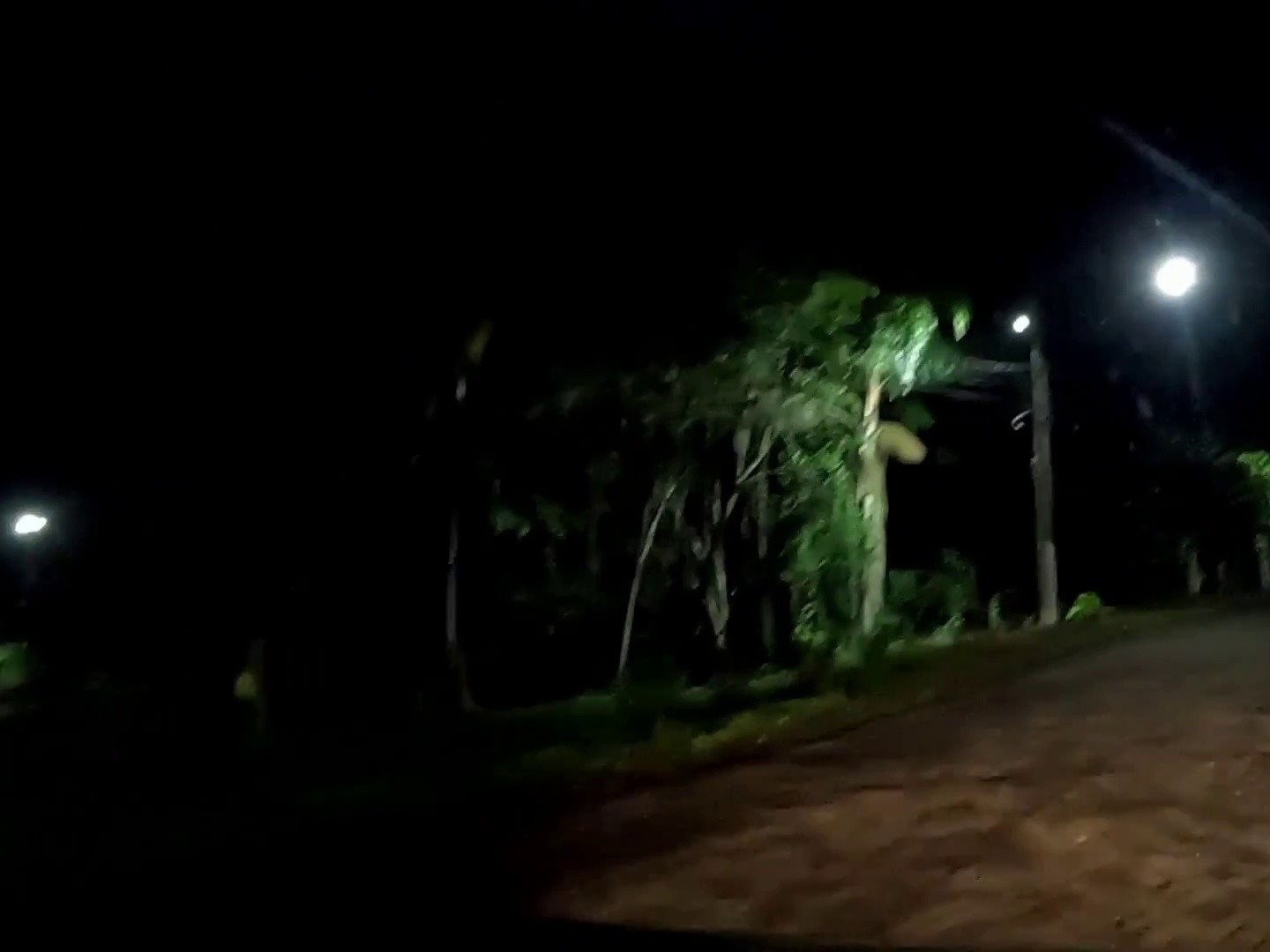}
    \end{subfigure}
    \caption{Misclassifications on the ROAD dataset, Subset \#2. From left to right, the model predictions and their ground-truth (GT) are Asphalt (\textbf{GT}: Belgian Blocks), Belgian Blocks (\textbf{GT}: Asphalt), and Off-road (\textbf{GT}: Belgian Blocks).}
    \label{fig:failure-vision-all}
\end{subfigure}

\caption{Representative examples of correct classifications (a) and misclassifications (b) produced by the vision-only model on Subset \#2 of the ROAD dataset. These samples illustrate the model’s behavior across diverse illumination and weather conditions, including extremely dark scenes, overexposed daytime frames, and transitional regions where multiple surface types coexist.}
\label{fig:qualitative-analysis-all}
\end{figure*}

\paragraph{Qualitative analysis.}
Representative qualitative examples are shown in \autoref{fig:qualitative-analysis-all}. The model demonstrates robust performance across diverse visual conditions, as illustrated in \autoref{fig:success-vision-all}. From left to right, these examples include an extremely dark nighttime scene with strong asphalt reflections, an underexposed image in which low illumination suppresses visual cues, and an overexposed daytime scene where excessive brightness degrades image quality. In all cases, the model correctly predicts the road surface type despite severe appearance distortions. Misclassification cases are presented in \autoref{fig:failure-vision-all}. From left to right, these include a scene with strong dashboard reflections on the windshield, which likely leads to the misclassification of \textit{Belgian Blocks} as \textit{Asphalt}; an ambiguous frame in which \textit{Asphalt} and \textit{Belgian Blocks} are simultaneously visible, resulting in label uncertainty; and a sample affected by artificial lighting and defocus, producing a warmer color tone that causes the model to misclassify the region as \textit{Off-road}.

\begin{table*}[b!]
\centering
\caption{Performance of our multimodal and vision-only models on ROAD Subset \#1 and on all vision-only subsets. The vision-only approach generalizes well across the expanded dataset, despite the increased difficulty of Subset \#2. The $\times$ in the first row indicates that the corresponding evaluation setting is not applicable, due to the absence of IMU data in all subsets.}
\label{tab:results-visiononly}
\begin{tabular}{
    l
    c
    l S[table-format=2.1]
    c
    l S[table-format=3.1]
}
\toprule
\multirow{2}{*}{\textbf{Method}} &
  \multicolumn{3}{c}{\textbf{ROAD Dataset} (Subset \#1)} &
  \multicolumn{3}{c}{\textbf{ROAD Dataset} (All subsets)} \\
\cmidrule(lr){2-4} \cmidrule(lr){5-7}
& {\textbf{Acc. (\%) \textuparrow}} & \multicolumn{2}{c}{\textbf{F1-score (\%)  \textuparrow}} 
& {\textbf{Acc. (\%) \textuparrow}} & \multicolumn{2}{c}{\textbf{F1-score (\%)  \textuparrow}} \\
\midrule
\multirow{3}{*}{Ours (Camera + IMU)} 
  & \multirow{3}{*}{98.2} & Asphalt     & 98.9 & \multirow{3}{*}{$\times$} & Asphalt        & $\times$ \\
  &                       & Cobblestone & 87.6 &                       & Belgian blocks & $\times$ \\
  &                       & Dirt road   & 100.0 &                       & Off-road       & $\times$ \\
\cmidrule(r){1-7}
\multirow{3}{*}{Ours (Vision-only)} 
  & \multirow{3}{*}{\textbf{98.4}} & Asphalt     & \textbf{99.0} & \multirow{3}{*}{96.6} & Asphalt        & 97.7 \\
  &                       & Cobblestone & 89.7 &                       & Belgian blocks & \textbf{95.3} \\
  &                       & Dirt road   & \textbf{100.0} &                       & Off-road       & 93.2 \\
\bottomrule
\end{tabular}
\end{table*}

With respect to the proposed framework, this work clarifies how multimodal fusion between RGB images and inertial signals can be effectively leveraged for road surface classification under realistic driving conditions. By combining modality-specific encoders with a lightweight bidirectional cross-attention mechanism and an adaptive gating strategy, the framework enables selective information exchange between vision and IMU streams while remaining resilient to domain shifts. Experimental evidence shows that this design allows the model to adapt its reliance on each modality according to signal reliability, preserving performance under challenging conditions such as low illumination, heterogeneous textures, and surface transitions. Importantly, the framework remains functional in degraded-sensor settings without architectural changes, providing a unified formulation that supports both multimodal and vision-only operation within the same learning paradigm.

\section{Conclusion}
This work introduces a new multimodal framework for camera--IMU road surface classification under realistic driving conditions. By combining modality-specific encoders with a lightweight bidirectional cross-attention mechanism and an adaptive gating strategy, the framework enables selective information exchange between vision and IMU streams while remaining resilient to domain shifts. Experimental results indicate that this design allows the model to adapt its reliance on each modality according to signal reliability, preserving performance under challenging conditions such as low illumination, heterogeneous textures, and surface transitions.

To support research on multimodal fusion under diverse driving conditions, we also introduce the ROAD dataset, structured into three complementary subsets: (i) a sensor-fusion subset with tightly synchronized camera and IMU streams for evaluating multimodal learning; (ii) a large-scale collection of unconstrained real-world driving scenarios providing broad environmental variability; and (iii) a synthetic subset built from high-fidelity simulation, enabling training on scenarios that are difficult to reproduce on public roads. By incorporating synthetic data, ROAD reflects a growing trend in intelligent transportation systems research toward simulation-enhanced training pipelines and provides a valuable resource for studying domain transfer, robustness, and failure modes.

Across experiments on both the PVS benchmark and the ROAD dataset, the proposed framework achieves state-of-the-art performance, surpassing the previous state of the art \citep{van2025hybrid} by 1.4 percentage points on PVS and by 11.6 percentage points on the ROAD benchmark (Subset~\#1), reaching 98.2\% accuracy while also attaining markedly higher class-wise F1-scores on minority and challenging surface classes.

Overall, our findings contribute to the development of more reliable road surface understanding modules, which are foundational components for a wide range of perception and decision-making systems, including predictive maintenance, vehicle dynamics optimization, driver-assistance pipelines, and road condition monitoring. Together, the proposed framework and dataset provide a solid basis for advancing multimodal research in these domains.

Future work includes expanding the synthetic subset with additional weather conditions, increasing the diversity of real-world acquisitions, and exploring temporal modeling approaches that capture longer-range dependencies in camera-IMU interactions.

\section*{Acknowledgements}
This research was funded by the Fundação de Desenvolvimento da Pesquisa – Fundep, Mover - Linha VI, and supported by the following industry partners: Volkswagen Truck \& Bus, Volkswagen, Stellantis (FCA and PSA), and Embeddo.
\clearpage

\appendix
\section{Extended qualitative analysis of sensor fusion model}
\label{appendix_a}

\begin{figure*}[B!]
\centering

    \centering
    \begin{subfigure}{0.24\textwidth}
        \centering
        \includegraphics[width=\linewidth]{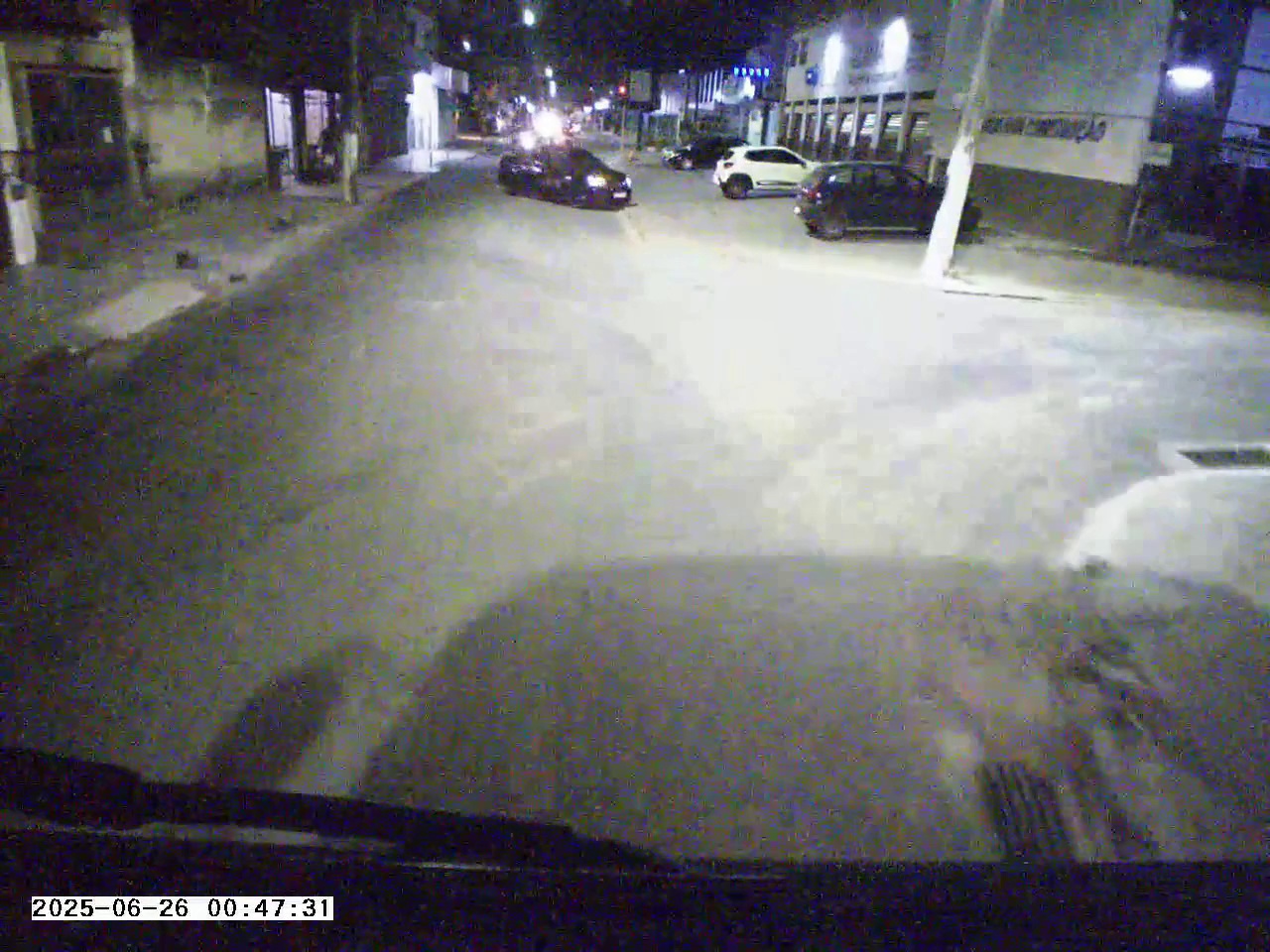}
    \end{subfigure}
    \hfill
    \begin{subfigure}{0.24\textwidth}
        \centering
        \includegraphics[width=\linewidth]{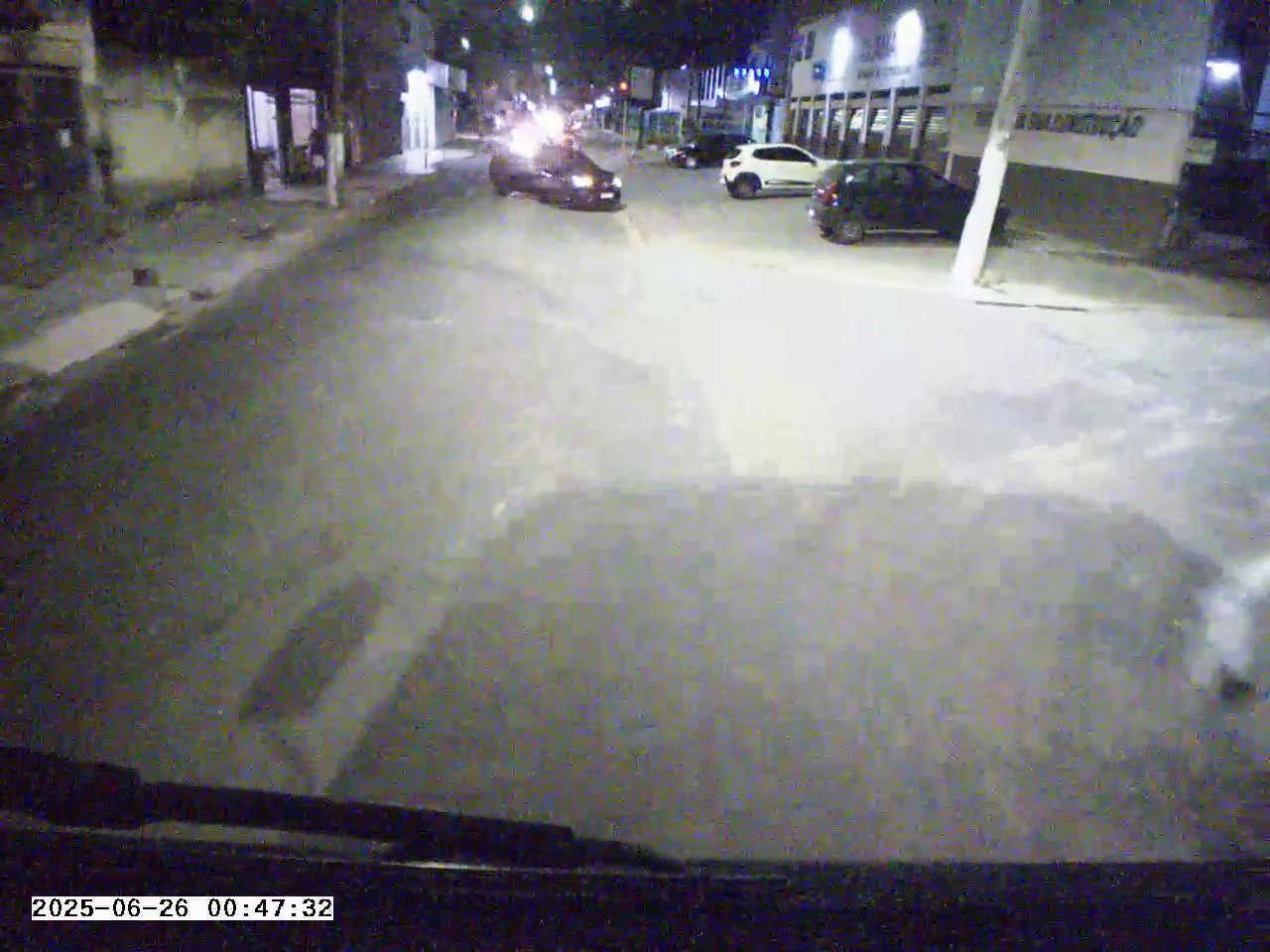}
    \end{subfigure}
    \hfill
    \begin{subfigure}{0.24\textwidth}
        \centering
        \includegraphics[width=\linewidth]{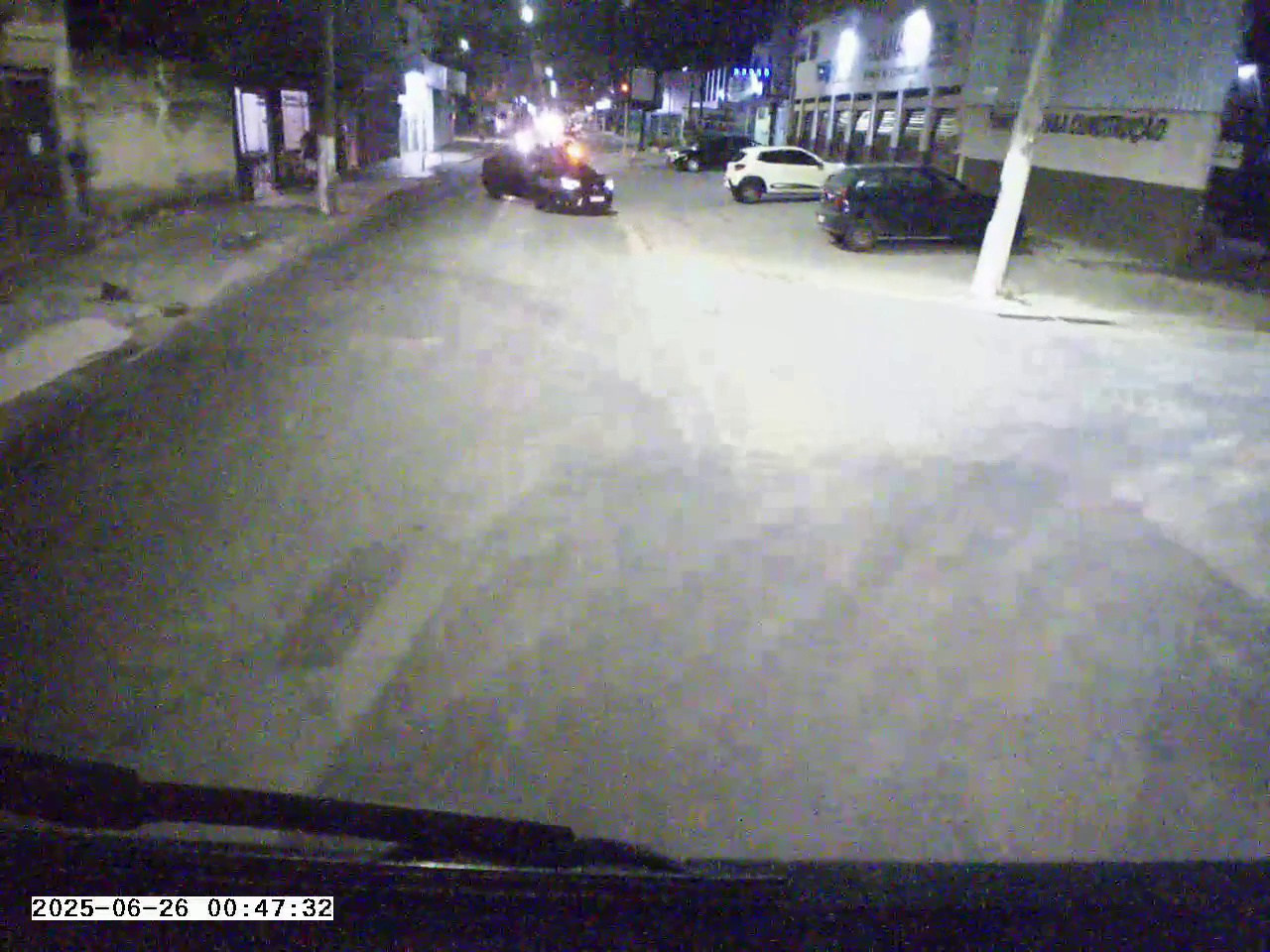}
    \end{subfigure}
    \hfill
    \begin{subfigure}{0.24\textwidth}
        \centering
        \includegraphics[width=\linewidth]{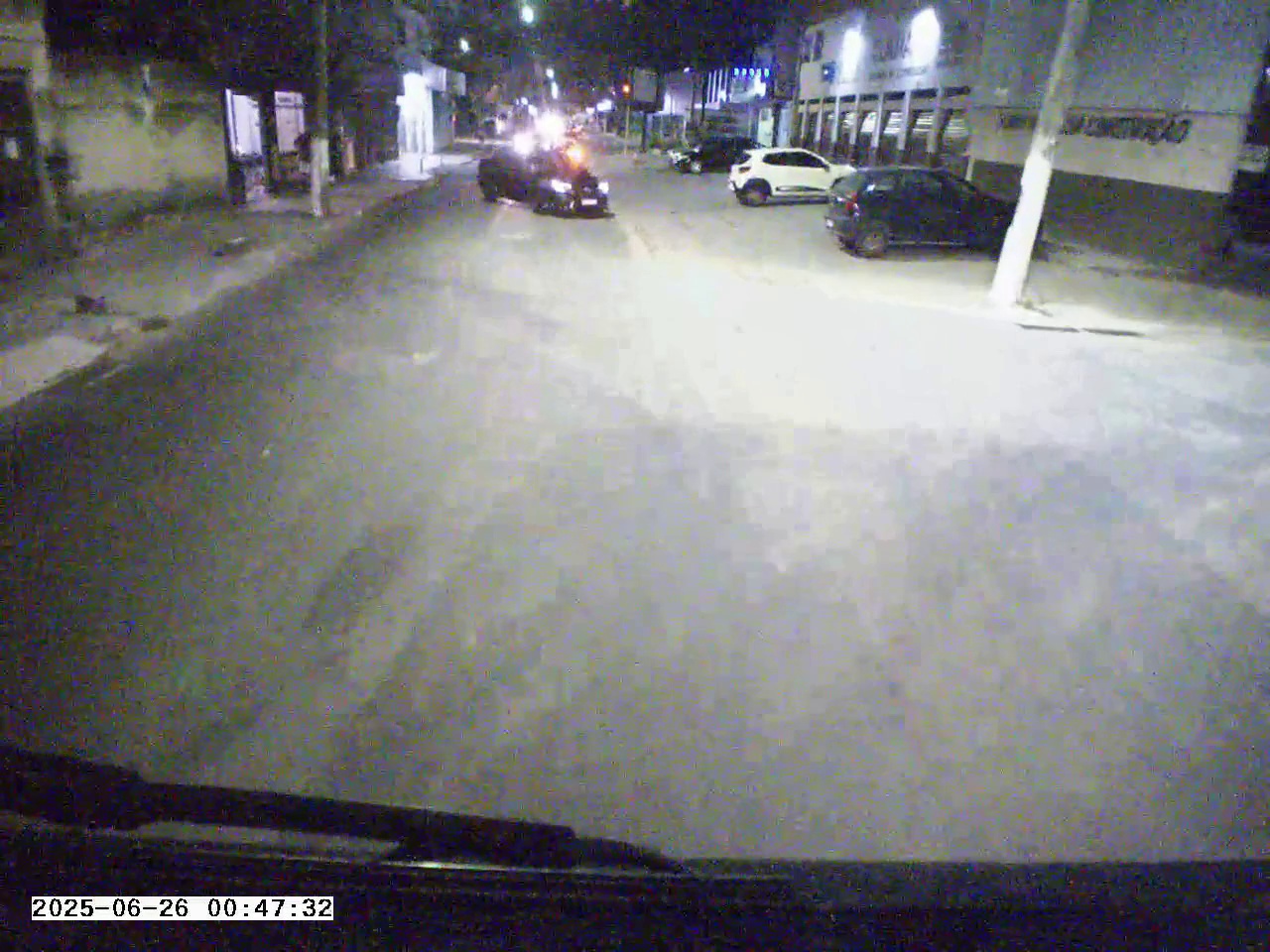}
    \end{subfigure}

    \centering
    \begin{subfigure}{0.24\textwidth}
        \centering
        \includegraphics[width=\linewidth]{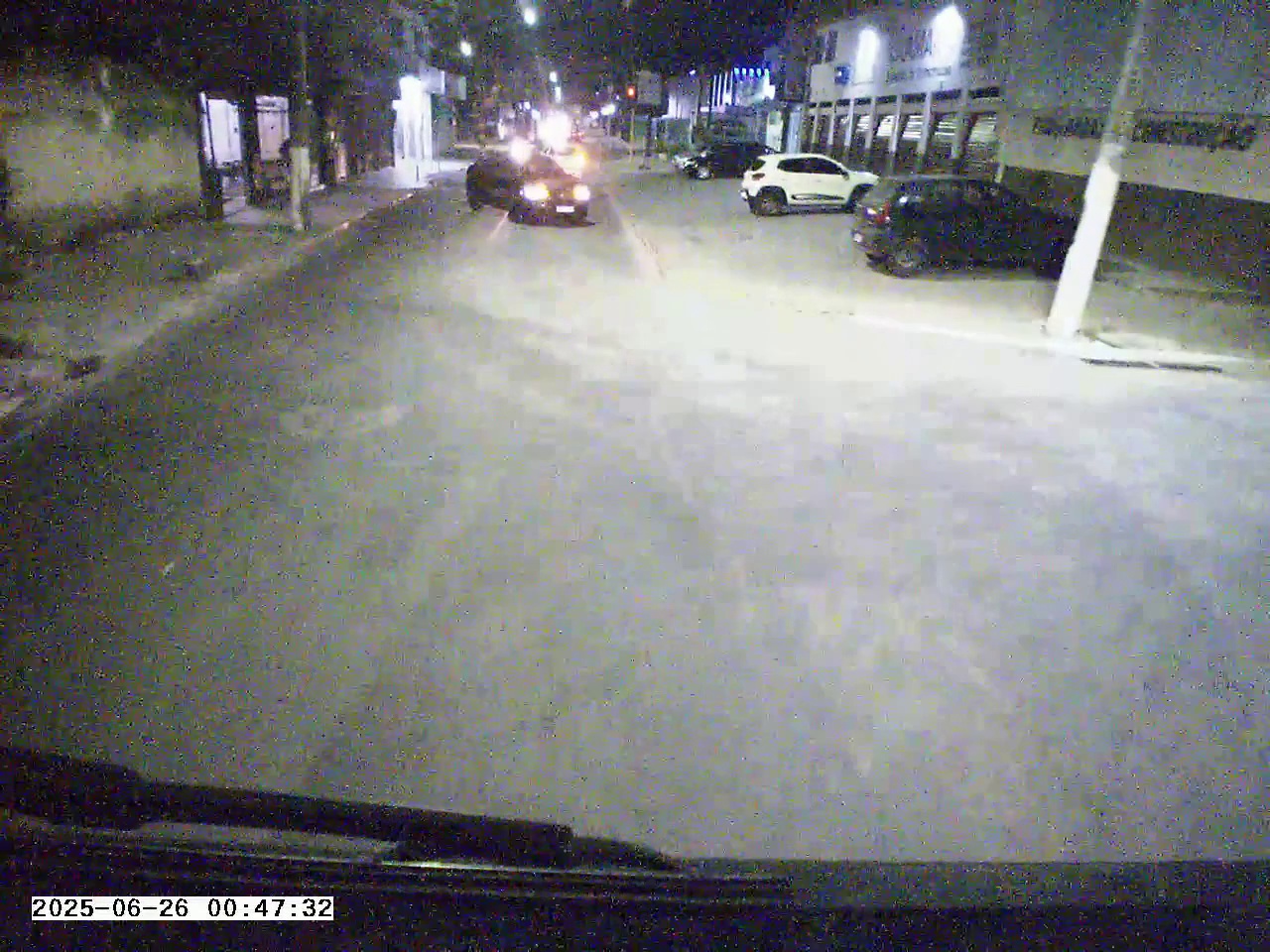}
    \end{subfigure}
    \hfill
    \begin{subfigure}{0.24\textwidth}
        \centering
        \includegraphics[width=\linewidth]{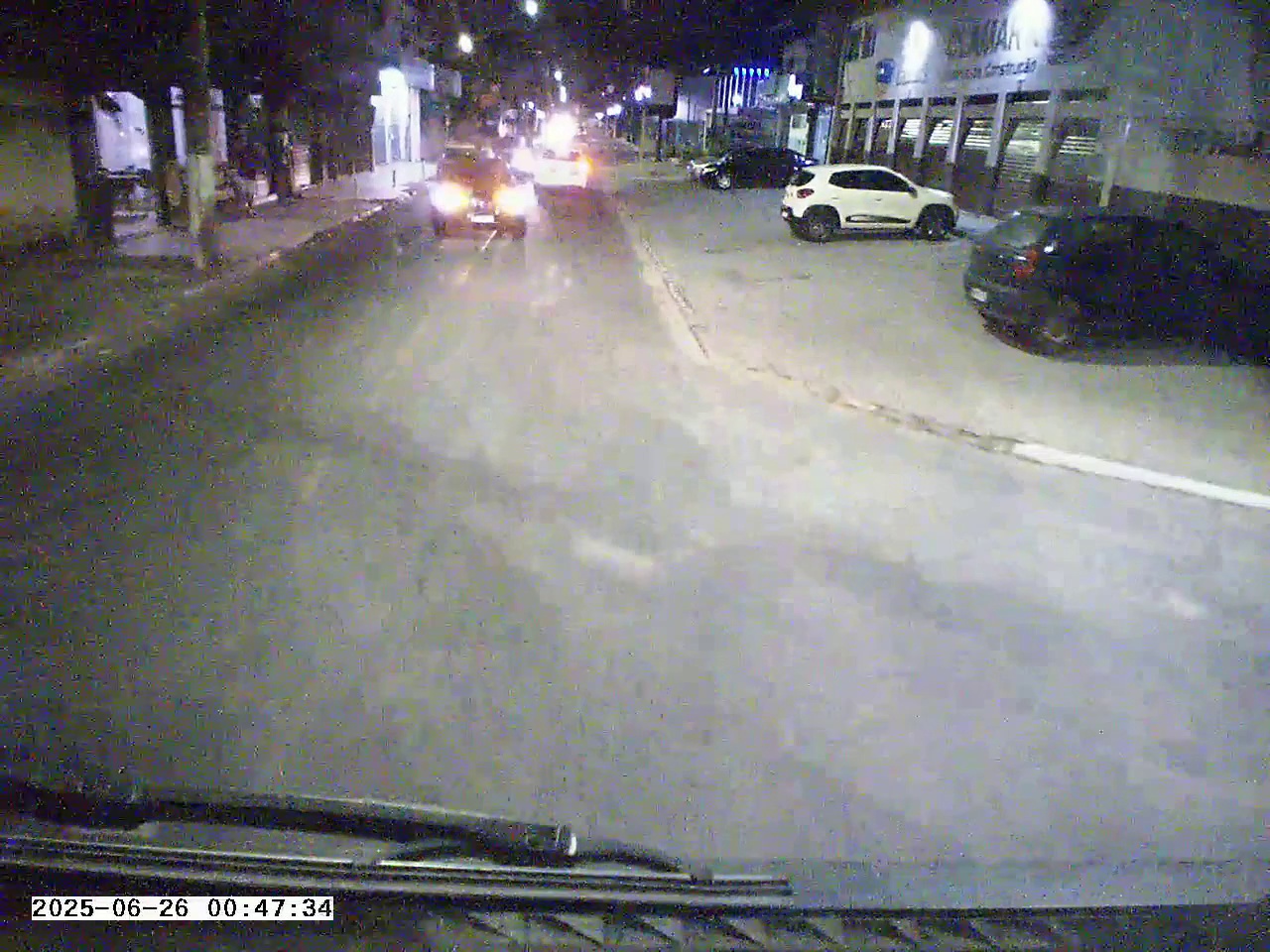}
    \end{subfigure}
    \hfill
    \begin{subfigure}{0.24\textwidth}
        \centering
        \includegraphics[width=\linewidth]{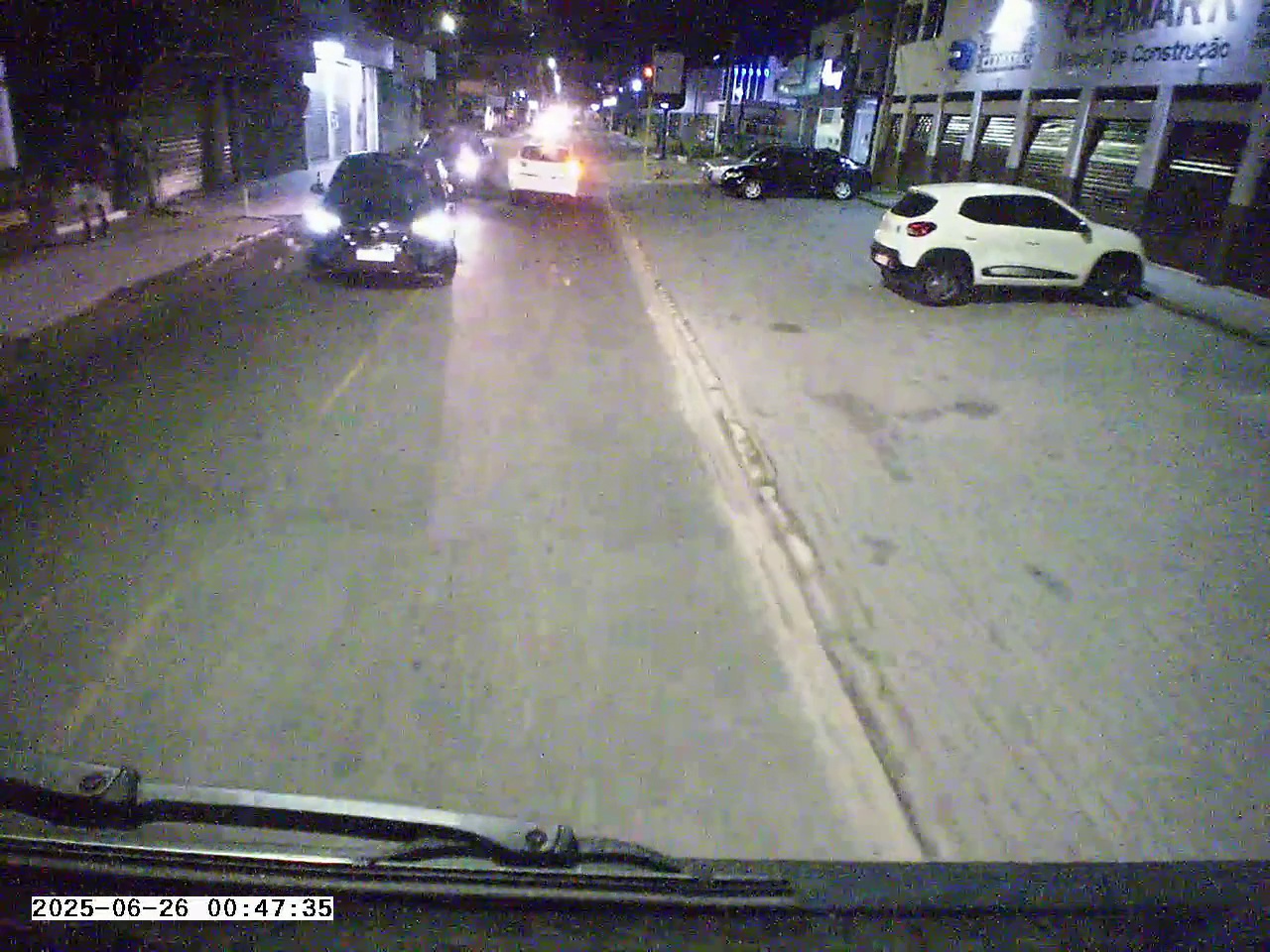}
    \end{subfigure}
    \hfill
    \begin{subfigure}{0.24\textwidth}
        \centering
        \includegraphics[width=\linewidth]{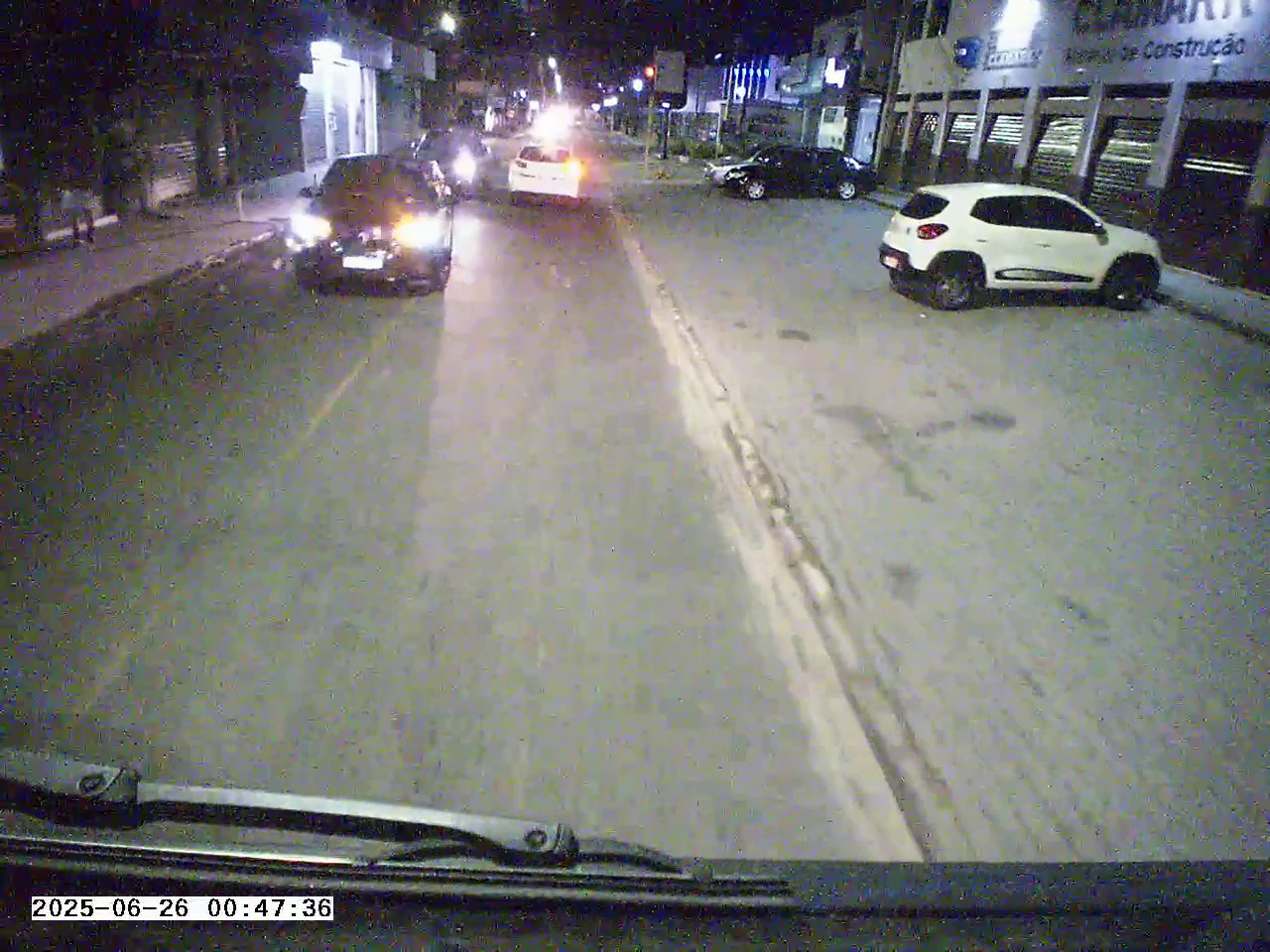}
    \end{subfigure}

    \centering
    \begin{subfigure}{0.24\textwidth}
        \centering
        \includegraphics[width=\linewidth]{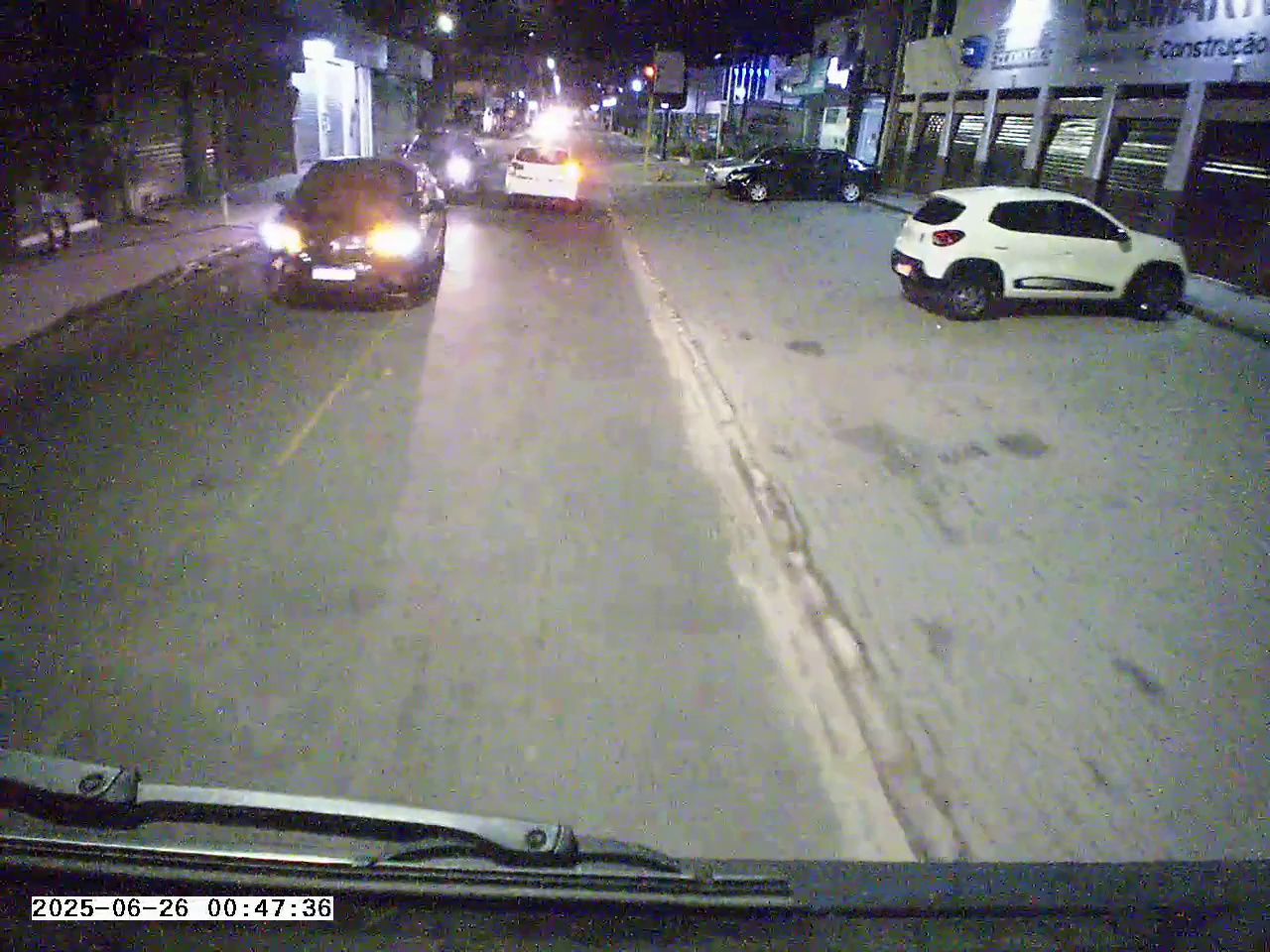}
    \end{subfigure}
    \hfill
    \begin{subfigure}{0.24\textwidth}
        \centering
        \includegraphics[width=\linewidth]{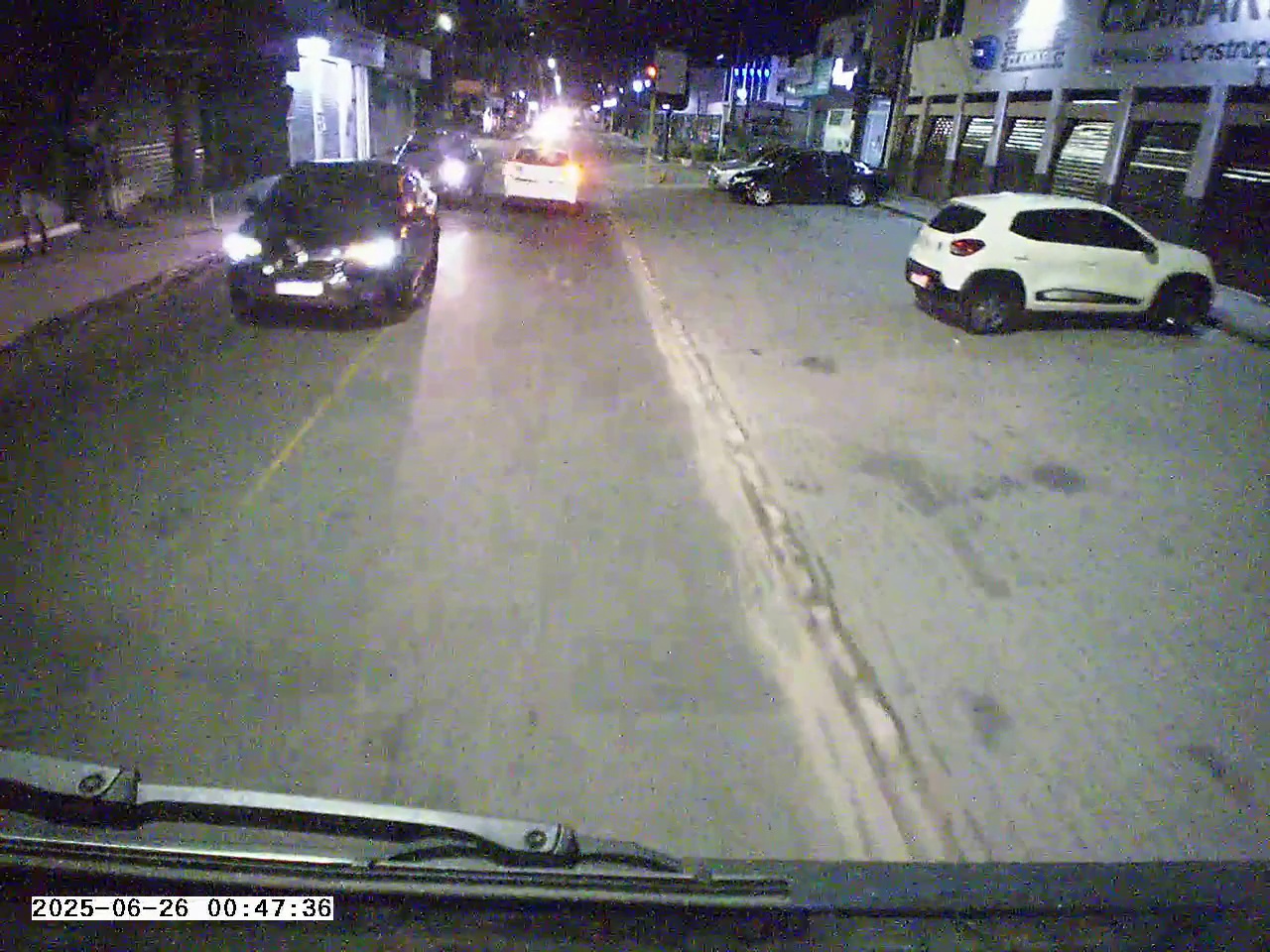}
    \end{subfigure}
    \hfill
    \begin{subfigure}{0.24\textwidth}
        \centering
        \includegraphics[width=\linewidth]{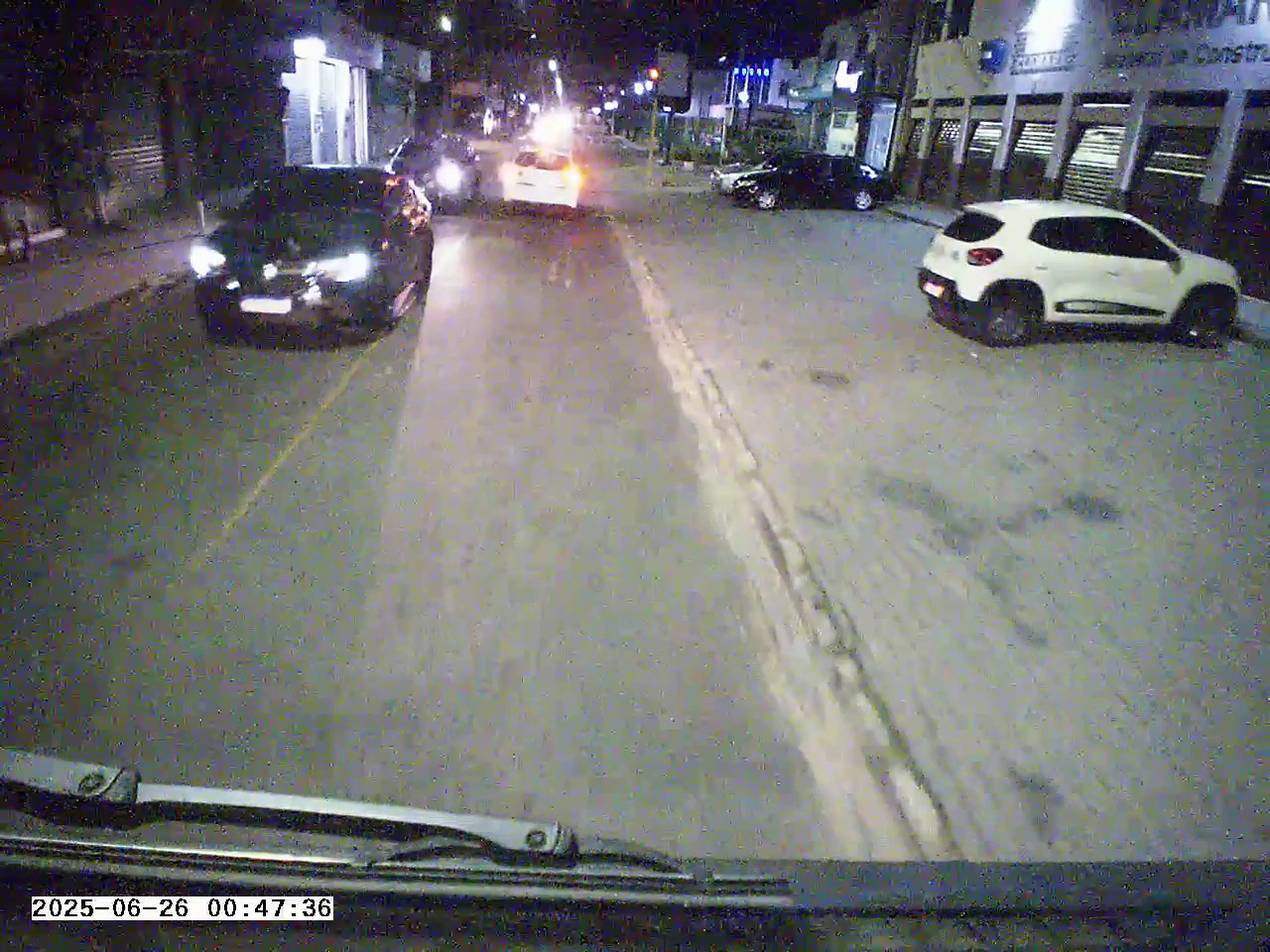}
    \end{subfigure}
    \hfill
    \begin{subfigure}{0.24\textwidth}
        \centering
        \includegraphics[width=\linewidth]{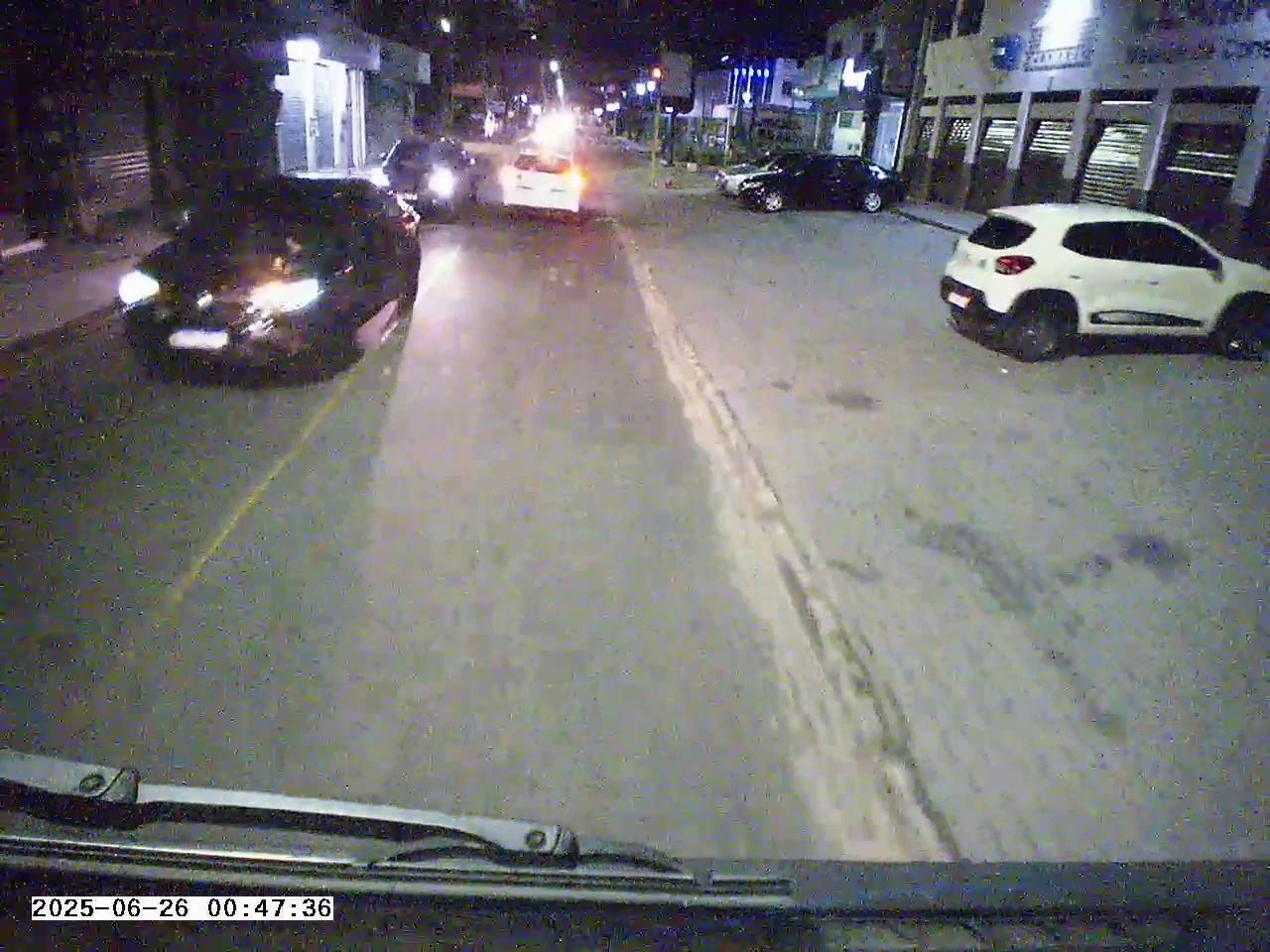}
    \end{subfigure}

    \centering
    \begin{subfigure}{0.24\textwidth}
        \centering
        \includegraphics[width=\linewidth]{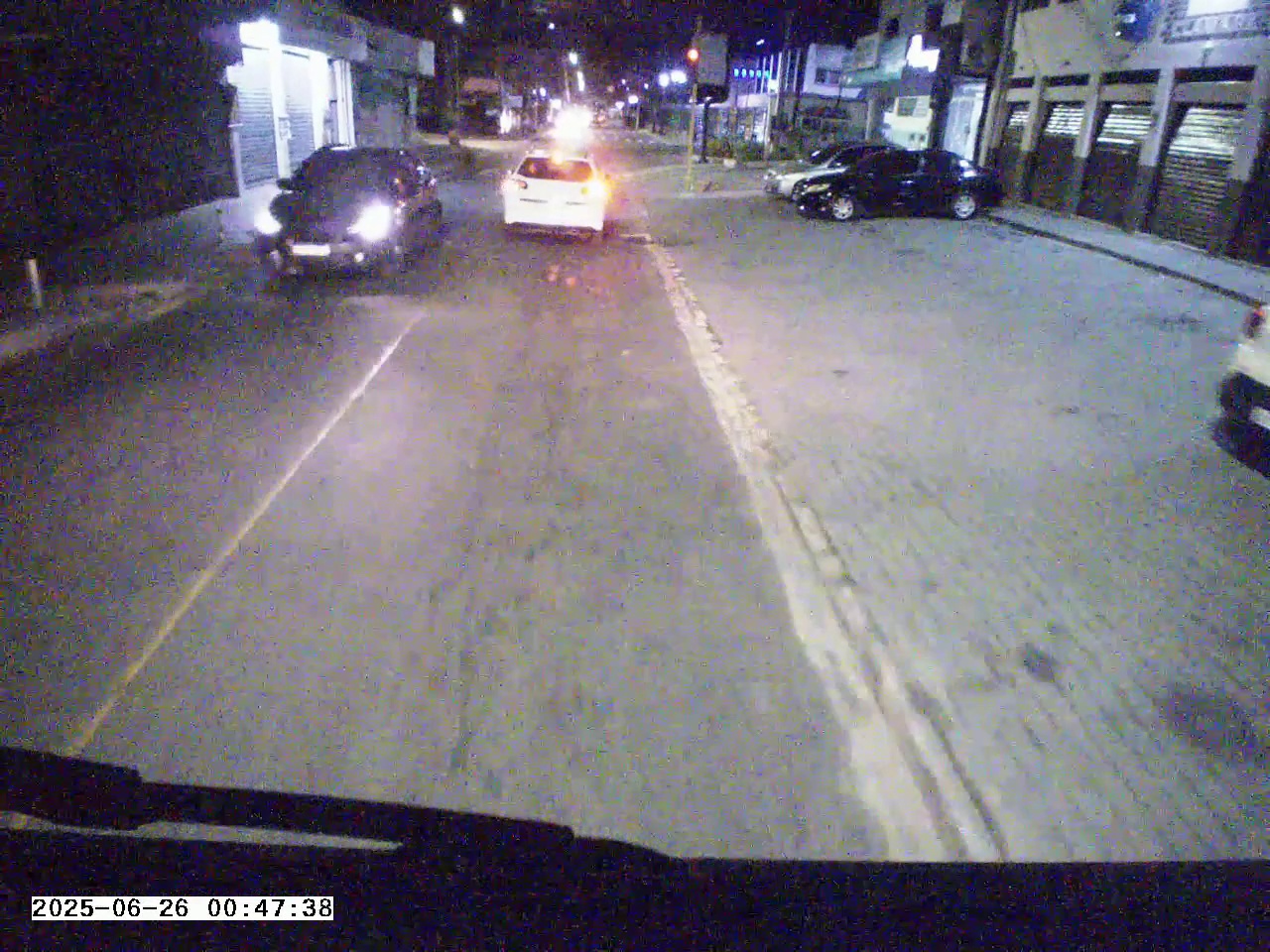}
    \end{subfigure}
    \hfill
    \begin{subfigure}{0.24\textwidth}
        \centering
        \includegraphics[width=\linewidth]{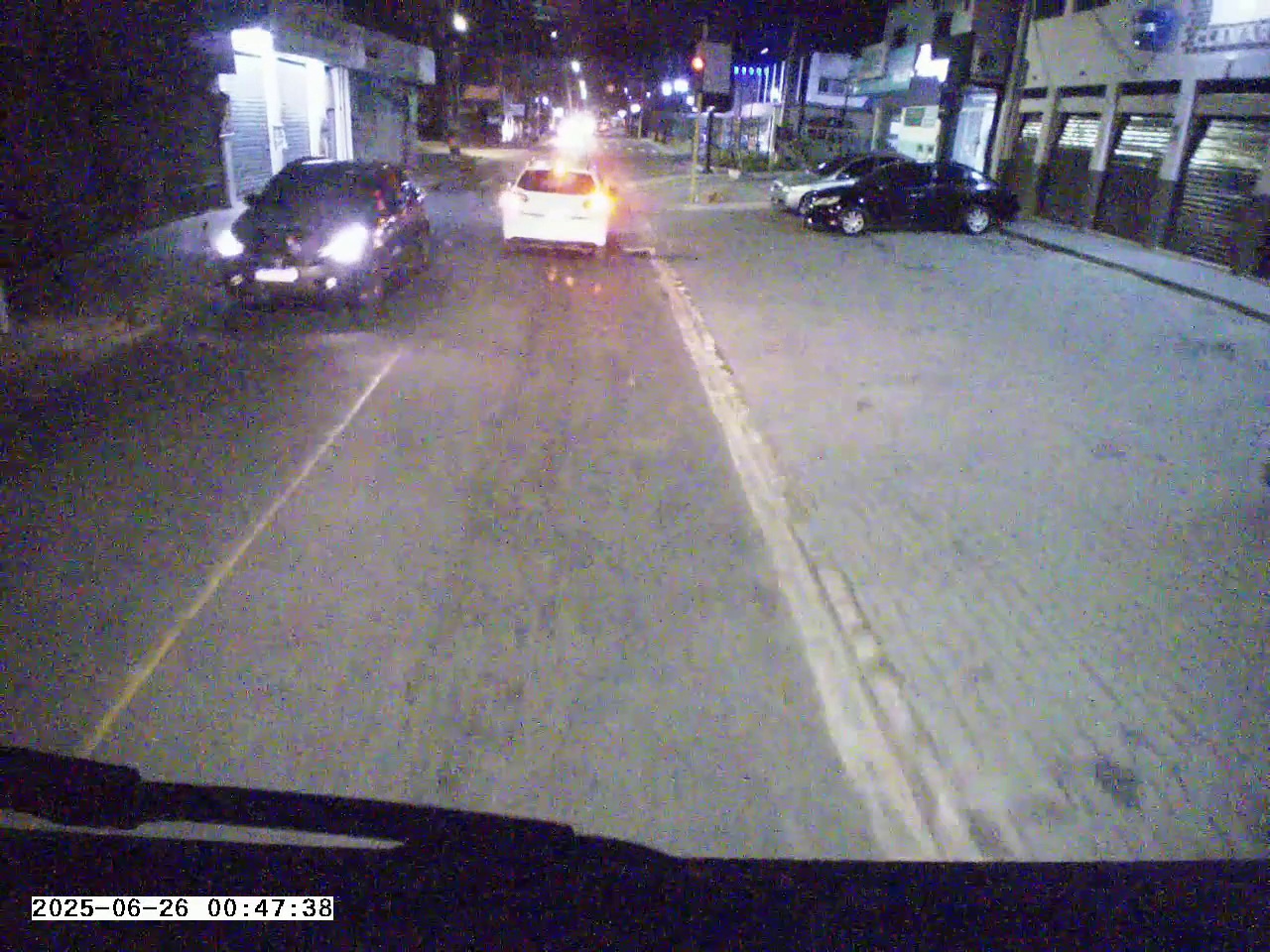}
    \end{subfigure}
    \hfill
    \begin{subfigure}{0.24\textwidth}
        \centering
        \includegraphics[width=\linewidth]{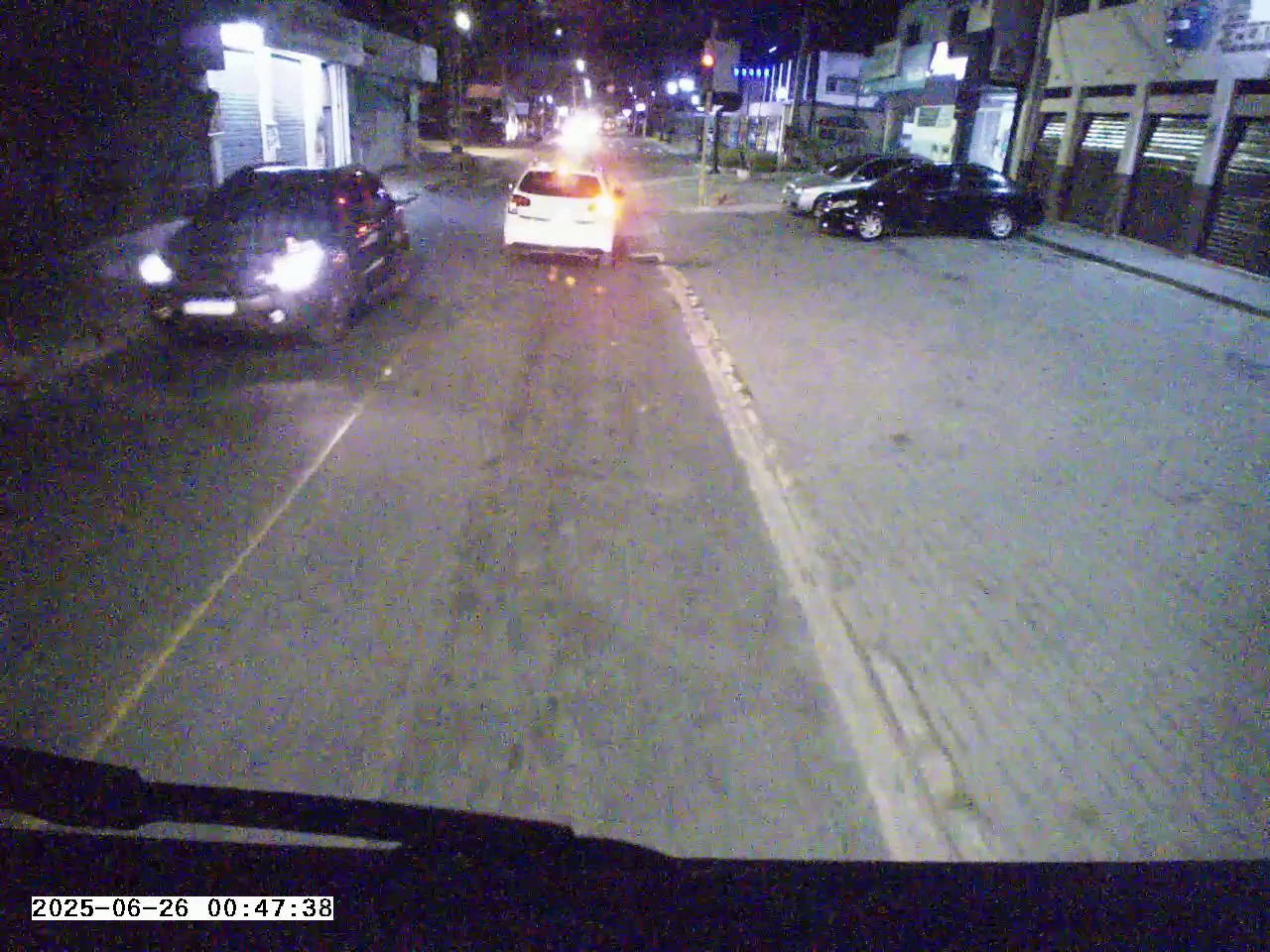}
    \end{subfigure}
    \hfill
    \begin{subfigure}{0.24\textwidth}
        \centering
        \includegraphics[width=\linewidth]{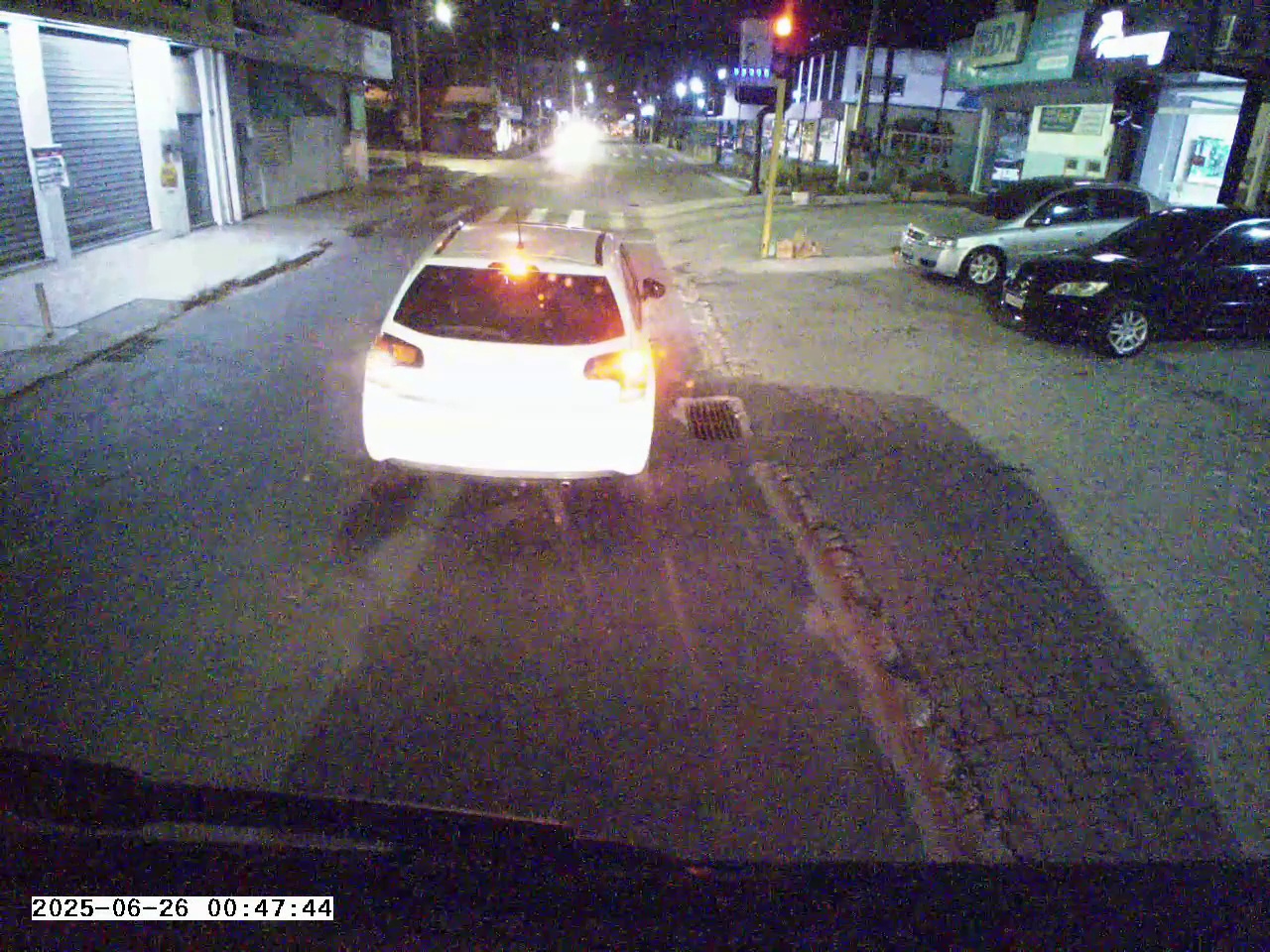}
    \end{subfigure}
        \begin{subfigure}{0.24\textwidth}
        \centering
        \includegraphics[width=\linewidth]{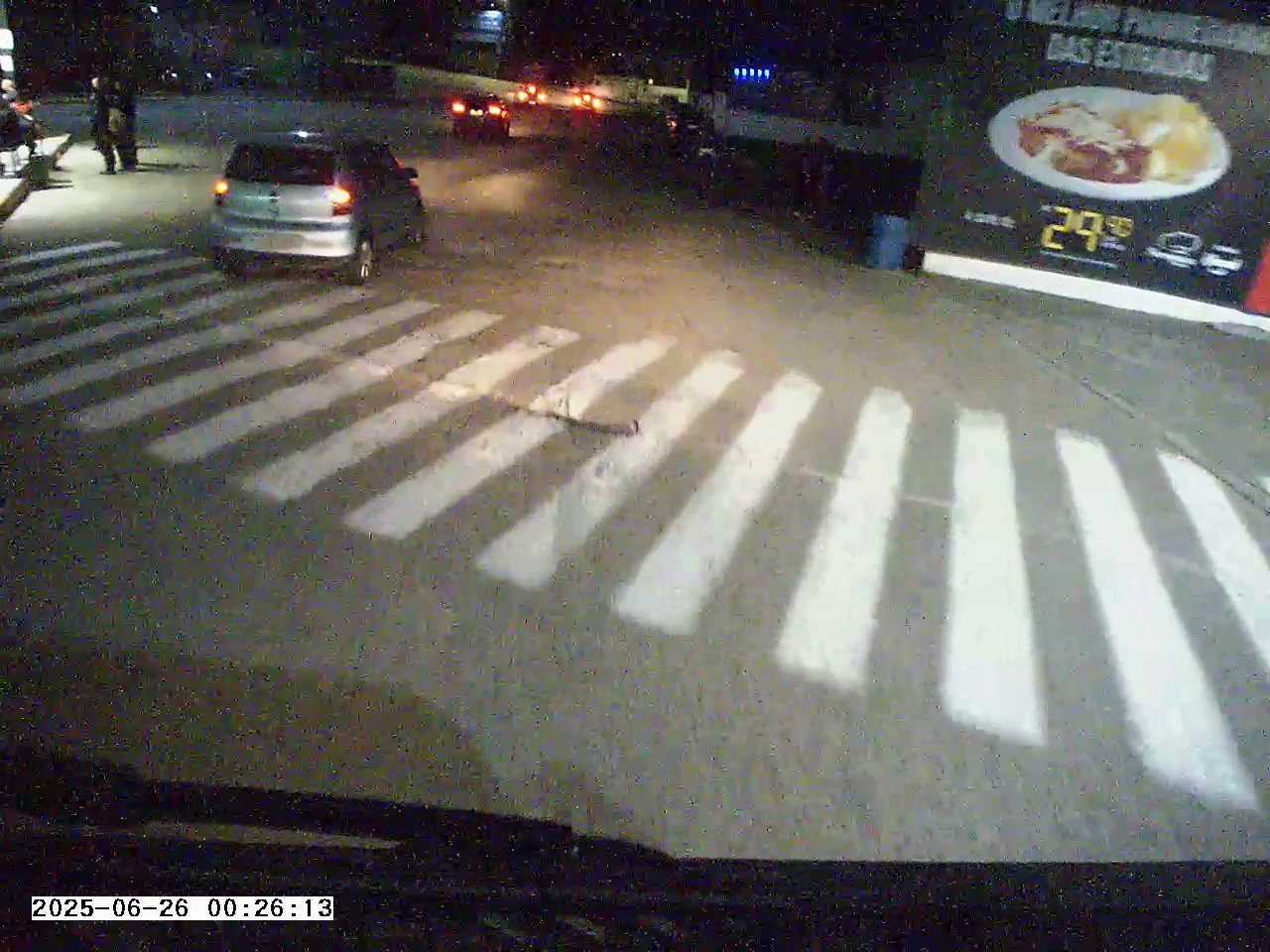}
    \end{subfigure}
    \hfill
    \begin{subfigure}{0.24\textwidth}
        \centering
        \includegraphics[width=\linewidth]{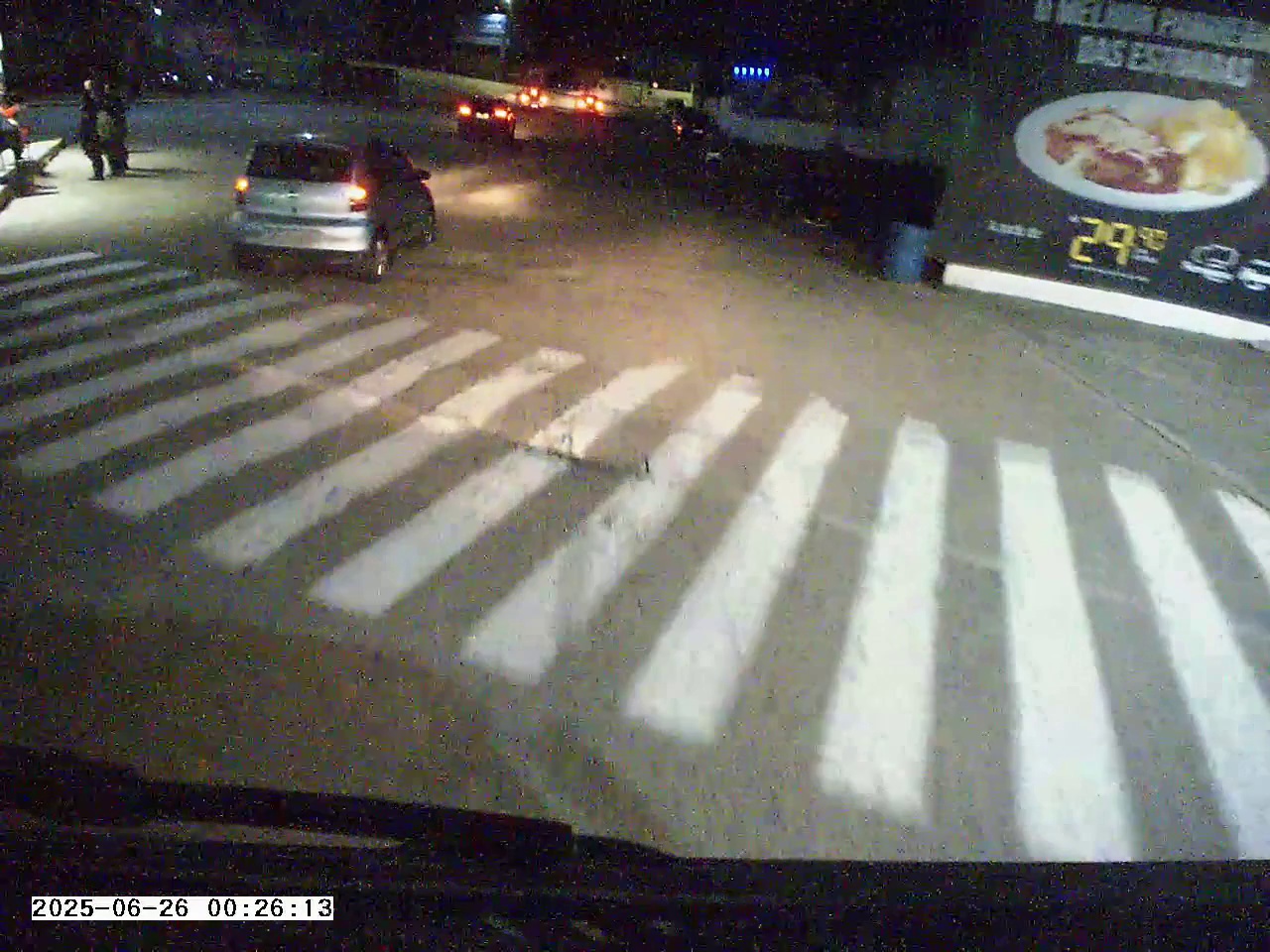}
    \end{subfigure}
    \hfill
    \begin{subfigure}{0.24\textwidth}
        \centering
        \includegraphics[width=\linewidth]{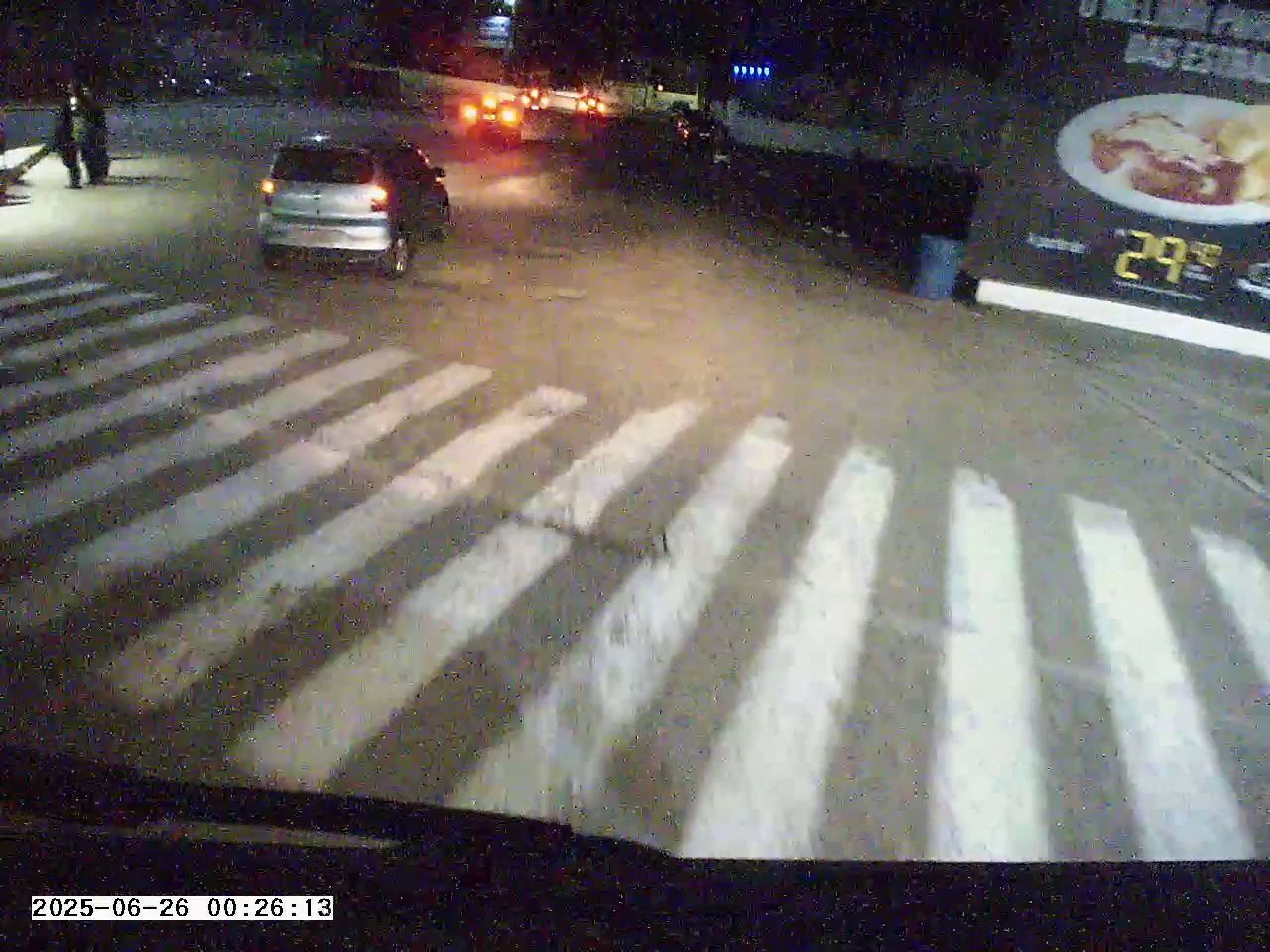}
    \end{subfigure}
    \hfill
    \begin{subfigure}{0.24\textwidth}
        \centering
        \includegraphics[width=\linewidth]{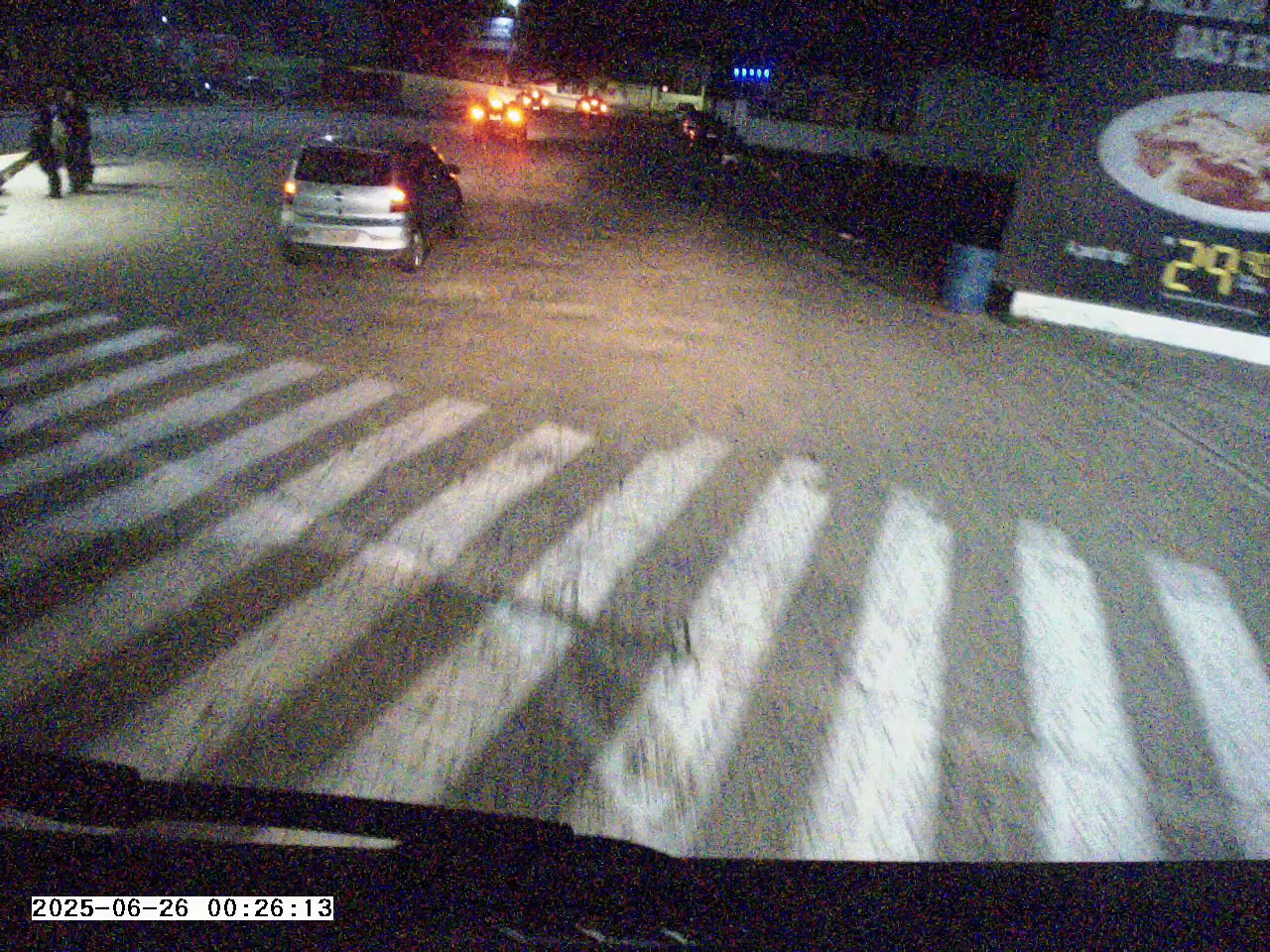}
    \end{subfigure}

\caption{Temporal blocks of misclassifications from our model.}
\label{fig:appendix_a_figure}
\end{figure*}

We present additional qualitative samples illustrating these temporally clustered failure cases. As we show in the sequences below, in \autoref{fig:appendix_a_figure}, misclassifications rarely occur as isolated frames. Instead, they will commonly appear as compact temporal blocks where the model briefly drifts to an incorrect class before re-stabilizing. By providing these extended examples, we aim to offer greater insight into the dynamics of our multimodal model and highlight opportunities for future improvements in temporal fusion and synchronization.

\FloatBarrier

\bibliographystyle{cas-model2-names}

\bibliography{refs}

\end{document}